\pdfoutput=1

\documentclass[10pt,twocolumn,letterpaper]{article}

\usepackage[pagenumbers]{cvpr} %

\usepackage{cuted} %

\usepackage{currfile} %

\usepackage{caption} %

\usepackage{microtype}  %

\usepackage{xspace}
\newcommand{\method}{Flow4RT\xspace}

\usepackage{amsmath,amsfonts,bm}

\def\eqref#1{equation~\ref{#1}}

\def\1{\bm{1}}

\def\mC{{\mathbf{C}}}

\def\mF{{\mathbf{F}}}

\def\mI{{\mathbf{I}}}

\def\mM{{\mathbf{M}}}

\def\mP{{\mathbf{P}}}

\def\mT{{\mathbf{T}}}

\def\mW{{\mathbf{W}}}

\DeclareMathAlphabet{\mathsfit}{\encodingdefault}{\sfdefault}{m}{sl}
\SetMathAlphabet{\mathsfit}{bold}{\encodingdefault}{\sfdefault}{bx}{n}

\newcommand{\R}{\mathbb{R}}

\DeclareMathOperator*{\argmin}{arg\,min}

\usepackage{multirow}
\usepackage{tabularx}
\usepackage{colortbl}
\definecolor{flowgreen}{HTML}{1E8E3E}
\colorlet{rank1}{flowgreen!30}   %
\colorlet{rank2}{flowgreen!17}   %
\colorlet{rank3}{flowgreen!8}    %

\definecolor{cvprblue}{rgb}{0.21,0.49,0.74}
\usepackage[breaklinks,colorlinks,allcolors=cvprblue]{hyperref}
\hypersetup{pdftitle={Flow4RT: Seeing Motion through Geometry},pdfauthor={Shenhan Qian, Shangzhe Wu, Daniel Cremers}}

\def\confName{CVPR}
\def\confYear{2027}

\title{\method: Seeing Motion through Geometry}

\author{Shenhan Qian\textsuperscript{1,2}
\qquad
Shangzhe Wu\textsuperscript{3}
\qquad
Daniel Cremers\textsuperscript{1,2}
\\[4pt]
\textsuperscript{1}Technical University of Munich
\qquad
\textsuperscript{2}MCML
\qquad
\textsuperscript{3}University of Cambridge
\\[18pt]
{\normalsize\href{https://shenhanqian.github.io/flow4rt}{\texttt{shenhanqian.github.io/flow4rt}}}
}

\makeatletter
\patchcmd{\@maketitle}{\vspace*{12pt}}{\vspace*{-6pt}}{}{\PackageWarning{arxiv}{title patch failed}}
\makeatother

\begin{document}
\maketitle
\begin{strip}
\centering
\includegraphics{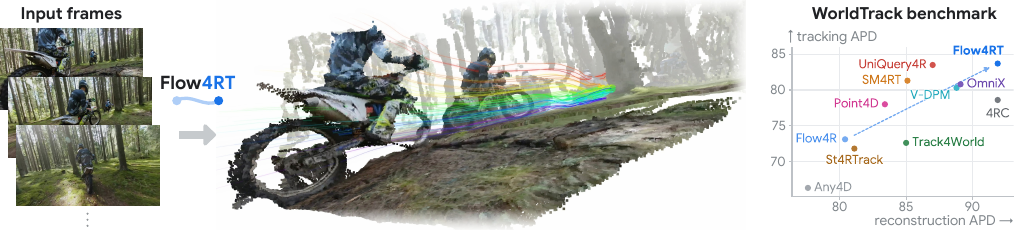}  %
\captionsetup{hypcap=false}
\captionof{figure}{%
    \textbf{World-space reconstruction and tracking from a single relative-motion formulation.}
    \method predicts local points for every view and displaced points with per-pixel weights for any requested pair, and a closed-form alignment turns these predictions into both reconstruction and tracking (\cref{fig:downstream-tasks}).
    Center: three reconstructed frames of a motocross clip stacked in one world frame, with trajectories of the front rider shown where unoccluded.
    Right: average APD on WorldTrack for reconstruction and tracking.
    \method improves substantially over its two-view predecessor Flow4R on both tasks and is the only evaluated method in the top two on both.
    }
\captionsetup{hypcap=true}
\label{fig:teaser}%
\end{strip}

\begin{abstract}
Most feedforward models for dynamic scenes follow VGGT's multi-head design: a dedicated head regresses camera poses, while separate decoders predict motion or tracks.
For image pairs, Flow4R offered a geometric alternative: predicting local and displaced 3D point maps with a rigidity map recovers camera pose, reconstruction, and tracking from one closed-form alignment.
We present \method, which extends this formulation to joint multi-view inference by seeing motion as displaced geometry.
Both outputs reduce to the same primitive---a 3D point map with a per-pixel weight, denoting confidence for local geometry and rigidity for displaced geometry.
A single shared dense head therefore decodes both: local geometry is predicted once per view with context from all input views, and relative motion for any requested pair.
Rigidity-weighted alignment recovers camera motion, reconstruction, and 3D tracks without a pose head, and lets the rigid reference be changed at inference by reweighting the alignment.
On WorldTrack, \method achieves the highest average tracking APD and ties for the highest reconstruction APD, ranking in the top two on both metrics for both tasks.
These results indicate that 4D reconstruction and tracking emerge from one ability: tracing 3D points across views.
\end{abstract}

\section{Introduction}
\label{sec:intro}

Feedforward models such as DUSt3R~\cite{dust3r} and VGGT~\cite{vggt} have made dense 3D reconstruction possible directly from uncalibrated images.
Extending this capability to dynamic scenes requires recovering 3D geometry and point trajectories while the camera and scene objects move independently.

Most recent multi-view dynamic models follow VGGT's multi-head blueprint~\cite{any4d,fourrc}: a dedicated head regresses camera poses, while separate decoders predict scene flow or time-varying geometry.
This design splits reconstruction and tracking across disjoint heads, and the regressed camera pose does not reveal which scene points support it.

Flow4R~\cite{flow4r} introduced a geometric alternative for image pairs.
It predicts a local 3D point map for one view and a displaced point map placing the same pixels at the other view's viewpoint and time, \ie, 3D scene flow.
Stationary points move between the two maps by a single rigid transformation, recovered in closed form via alignment weighted by a learned \emph{rigidity map}.
The same alignment brings geometry into a shared frame for reconstruction and locates points at the target time for tracking, uniting both tasks without a pose head.
However, Flow4R processes only two views at a time, lacking multi-view context and re-predicting local geometry for every pair.
This raises a central question: \emph{can this pairwise geometric formulation scale to joint multi-view inference without redundant predictions?}

\method shows that it can by decoupling multi-view context encoding from pairwise motion prediction.
All input views are encoded jointly so every view aggregates global context, and relative motion is decoded for requested pairs from these shared features.
Pairing every view with one anchor view in both directions suffices for both tasks: motion \emph{to} the anchor aligns local geometry in a common world frame for reconstruction, while motion \emph{from} the anchor tracks its points across time.

Moving to multiple views also simplifies the output representation.
Because \emph{motion can be seen as displaced geometry}, local geometry $\mP$ (with confidence $\mC$) and displaced geometry $\mP_{vt}$ (with rigidity $\mW$) share one format: a 3D point map with a per-pixel scalar weight.
A single shared dense head decodes both streams with separate scalar activations.
Geometry blocks infer local geometry once per view with context from all views, and motion blocks condition those features on a target view for each requested pair, matching the accuracy of separate heads (\cref{tab:ablation-head}).

On WorldTrack (\cref{tab:worldtrack}), \method achieves the highest average tracking APD, ties with 4RC for the highest reconstruction APD, and ranks second in average EPE on both tasks---the only method in the top two on both metrics for both tasks.
With weights fixed, joint multi-view context adds 3.7 tracking APD and 2.9 reconstruction APD over pairwise inference, and rigidity weights learned through alignment outperform even an oracle static mask.
Because camera pose is an explicit alignment rather than a regressed vector, reweighting the alignment to a moving rigid object expresses the same predictions relative to that object without re-running the network (\cref{fig:reference}).
Together, these results suggest that 4D reconstruction and tracking rest on the same geometric primitive: tracing 3D points between views.

Our contributions are:
\begin{itemize}
  \item We extend pairwise local-geometry and relative-motion prediction to joint multi-view inference, recovering world-space reconstruction and tracking via bidirectional anchor pairing and closed-form alignment without a pose head.
  \item We unify local and displaced geometry into a single point-map-and-weight representation decoded by one shared dense head, predicting local geometry once per view and motion for any requested pair.
  \item We achieve top balanced performance on WorldTrack (first in tracking APD, tied first in reconstruction APD, first overall) and isolate the gains of multi-view context and alignment-supervised rigidity.
\end{itemize}

\section{Related Work}
\label{sec:related_work}

\begin{figure*}[t]
    \centering
    \includegraphics[width=\linewidth]{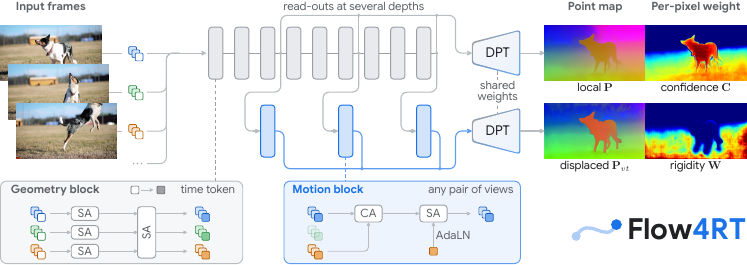}
    \caption{
    \textbf{\method predicts pixel-aligned local geometry and relative motion for every view.}
    Geometry blocks process all $N$ views, applying self-attention (SA) within each view and, from a chosen depth on, across views, so every view keeps its own tokens while sharing context.
    Each view also carries a time token that starts as the same learned embedding in every view (white) and is made view-specific by the geometry blocks (filled).
    At several read-out depths, motion blocks branch off and map a source view toward a target view: the source tokens cross-attend (CA) to the target tokens and are conditioned on the target's time token through AdaLN (insets, bottom).
    A DPT head with shared weights decodes each stream separately from its read-outs into the same format, a point map with a per-pixel weight: local points $\mP$ with confidence $\mC$, and displaced points $\mP_{vt}$ with rigidity $\mW$.
    \Cref{fig:downstream-tasks} turns these into reconstruction and tracking, and block and read-out counts are schematic.
    }
    \label{fig:pipeline}
\end{figure*}

\paragraph{Feedforward 3D reconstruction.}
Classical structure-from-motion systems such as COLMAP~\cite{colmap} reconstruct scenes through feature matching, triangulation, and bundle adjustment.
DUSt3R~\cite{dust3r} instead predicts dense point maps from image pairs, and MASt3R~\cite{mast3r} augments this representation with features for matching.
Subsequent methods support image streams, including Spann3R~\cite{spann3r}, CUT3R~\cite{cut3r}, and StreamVGGT~\cite{streamvggt}, or joint multi-view inference, including Fast3R~\cite{fast3r}, MUSt3R~\cite{must3r}, VGGT~\cite{vggt}, and VGGT-$\Omega$~\cite{vggtomega}.
Other developments include metric reconstruction with optional geometric inputs in MapAnything~\cite{mapanything}, depth-ray prediction in Depth Anything 3~\cite{da3}, and permutation-equivariant geometry prediction in $\pi^3$~\cite{pi3}.

\paragraph{Point tracking.}
PIPs~\cite{pips} recasts pixel tracking as long-range trajectory estimation through occlusions, and TAPIR~\cite{tapir}, LocoTrack~\cite{locotrack}, CoTracker~\cite{cotracker}, and CoTracker3~\cite{cotracker3} improve its accuracy and speed in the image plane.
TAPVid-3D~\cite{tapvid3d} benchmarks the task in 3D, where SpatialTracker~\cite{spatialtracker}, DELTA~\cite{delta}, and TAPIP3D~\cite{tapip3d} lift pixels into 3D using given depth (and optionally camera poses in TAPIP3D) to track them there.
While these methods either remain in 2D or rely on external geometric inputs, \method jointly infers dense 3D geometry and motion from raw images alone.

\paragraph{Dynamic reconstruction.}
MonST3R~\cite{monst3r} fine-tunes DUSt3R on dynamic scenes, Align3R~\cite{align3r} conditions it on monocular depth for temporally consistent video depth, and MegaSaM~\cite{megasam} folds motion cues into bundle adjustment.
Easi3R~\cite{easi3r} and VGGT4D~\cite{vggt4d} mine motion cues from the attention of DUSt3R and VGGT without training, while PAGE-4D~\cite{page4d} fine-tunes VGGT with a dynamics-aware mask that suppresses moving regions for pose estimation.
Geo4D~\cite{geo4d} instead repurposes a video diffusion model to predict point, disparity, and ray maps.
While these systems recover per-frame geometry and camera poses, dense trajectory estimation requires explicitly modeling point displacements across time.

\paragraph{4D reconstruction and tracking.}
Dynamic Point Maps~\cite{dynamicpointmaps}, D$^2$USt3R~\cite{d2ust3r}, and POMATO~\cite{pomato} extend the point map to carry time, regressing geometry at more than one timestamp.
Multi-view successors encode a video once and decode geometry at a requested time, either densely per target time, as in V-DPM~\cite{vdpm} and 4RC~\cite{fourrc}, or per point query, as in D4RT~\cite{d4rt}, UniQuery4R~\cite{uniquery4r}, and Point4D~\cite{point4d}.
Any4D~\cite{any4d} instead factors multi-view predictions into per-view depth, camera parameters, and scene flow.
Other approaches use dedicated motion representations: OmniX~\cite{omnix} predicts trajectory fields from dynamic tokens, while SM4RT~\cite{sm4rt} combines rigid $SE(3)$ motion bases with per-pixel assignment weights.
Track4World~\cite{track4world} estimates dense flow between frame pairs using 3D correlation, and SpatialTrackerV2~\cite{spatialtrackerv2} jointly estimates geometry, camera motion, and point trajectories.
Flow4R~\cite{flow4r} introduced a relative pairwise formulation, predicting local geometry and 3D displacements to recover camera pose via rigidity-weighted alignment.
\method scales this formulation to joint multi-view inference and unifies local and displaced geometry into a single format---a point map with a per-pixel weight---so that one shared dense head decodes local geometry once per view and motion for any requested pair without pose regression.

\section{Method}
\label{sec:methodology}

\begin{figure*}[t]
    \centering
    \includegraphics{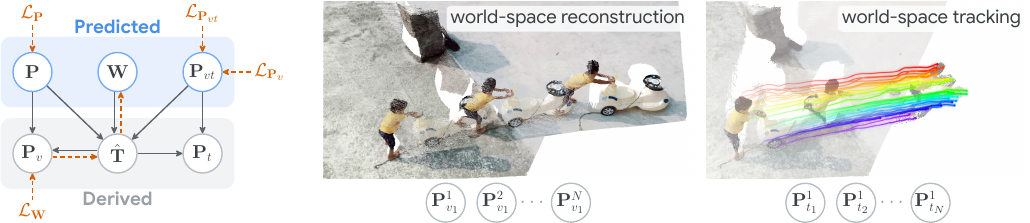}  %
    \caption{
    \textbf{Reconstruction and tracking from rigid alignment.}
    Left: predicted quantities (blue band), derived quantities (gray band), and supervision paths (dashed orange).
    Rigidity $\mW$ is supervised through the alignment solve (\cref{eq:solve}).
    Right: with view $1$ as the anchor, reconstruction aligns the static ground while preserving the child's and the toy car's positions across time, whereas tracking follows the anchor's points through the sequence (\cref{eq:recon_track}), drawn over the anchor's points at the times of the same three frames (dimmed) and colored per trajectory.
    Clip from SA-V~\cite{ravi2024sam2}.
    }
    \label{fig:downstream-tasks}
\end{figure*}

\method models dynamic scenes by combining joint multi-view feature processing with flexible, pairwise geometric motion estimation.

\subsection{Local Geometry and Relative Motion}
\label{sec:method_formulation}

\paragraph{Input.}
\method takes a set of $N$ images $\{\mI^n\}_{n=1}^{N}$ of a dynamic scene, with $\mI^n \in \R^{H \times W \times 3}$.
We denote the viewpoint of view $n$ by $v_n$ and its capture time by $t_n$.

\paragraph{Two pairs of predictions.}
For every view, the geometry branch predicts a local point map $\mP$, which places each pixel at that view's own viewpoint and time, and a confidence map $\mC$.
Toward a target view, the motion branch predicts a displaced point map $\mP_{vt}$, which places the same pixels at the target's viewpoint and time, and a rigidity map $\mW$:
\begin{equation*}
    \mP, \mP_{vt} \in \R^{H \times W \times 3} ,
    \quad
    \mC \in (1, \infty)^{H \times W} ,
    \quad
    \mW \in (0, 1)^{H \times W} .
\end{equation*}
Both point maps are indexed by source-view pixels, establishing dense correspondence by construction even when a point is occluded in the target view.
The rigidity is normalized over source pixels, $\sum_i \mW^{(i)} = 1$.
For clarity, we omit the source and target indices until \cref{sec:method_downstream}.

\paragraph{Shared output structure.}
Our key insight is that \emph{motion can be seen as displaced geometry}: $\mP$ and $\mP_{vt}$ describe the same physical points observed at different viewpoints and timestamps.
Because both are dense 3D point maps, both can be decoded by a single shared dense prediction head from differently conditioned features.
Their accompanying weights serve distinct roles: confidence $\mC$ guides regression during training, whereas rigidity $\mW$ weights the rigid alignment used to recover camera motion.
Unifying both outputs into a 3D point and a per-pixel scalar allows a single shared head with separate scalar activations (\cref{sec:method_arch}).

\subsection{Recovering Rigid Alignment and Scene Motion}
\label{sec:method_solve}

\paragraph{Rigidity-weighted alignment.}
For points on a stationary scene region, the change from $\mP$ to $\mP_{vt}$ is governed by a single rigid transformation.
We estimate this transformation by aligning corresponding entries of the two point maps, weighted by $\mW$:
\begin{equation}
    \hat{\mT} = \argmin_{\mT \in SE(3)} \sum_{i=1}^{HW} \mW^{(i)} \left\| \mP_{vt}^{(i)} - \mT \mP^{(i)} \right\|_2^2 .
    \label{eq:solve}
\end{equation}
This is a weighted Procrustes problem, solved in closed form by the singular value decomposition of the $\mW$-weighted cross-covariance of the two point sets.
We train $\mW$ end-to-end by differentiating through this solve (\cref{sec:method_training}).
When $\mW$ selects stationary scene regions, $\hat{\mT}$ recovers the relative camera pose without requiring a dedicated pose regression head.
This keeps the model simple and physically grounded: camera motion reuses the point maps essential for reconstruction and tracking anyway, with its only additional output being a per-pixel weight from the shared head.
Furthermore, changing the weights to select a moving rigid object directly yields an alignment relative to that object.

\paragraph{Reconstruction and tracking from the same alignment.}
The forward transformation places source-time geometry in the target camera frame, while its inverse expresses target-time geometry in the source camera frame:
\begin{equation}
    \mP_{v} = \hat{\mT} \mP ,
    \qquad
    \mP_{t} = \hat{\mT}^{-1} \mP_{vt} .
    \label{eq:pv_pt}
\end{equation}
Thus, $\mP_v$ supports reconstruction by aligning the source point cloud with the target viewpoint, and $\mP_t$ supports tracking by locating source points at the target time in the source coordinate frame.
The total displacement $\mF = \mP_{vt} - \mP$ can also be decomposed into the displacement induced by camera motion and the residual due to scene motion:
\begin{equation}
    \mF_{v} = \mP_{v} - \mP ,
    \qquad
    \mF_{t} = \mF - \mF_{v} .
    \label{eq:decomp}
\end{equation}

\subsection{World-Space Reconstruction and Tracking}
\label{sec:method_downstream}

\begin{figure*}[t]
    \centering
    \includegraphics[width=\linewidth]{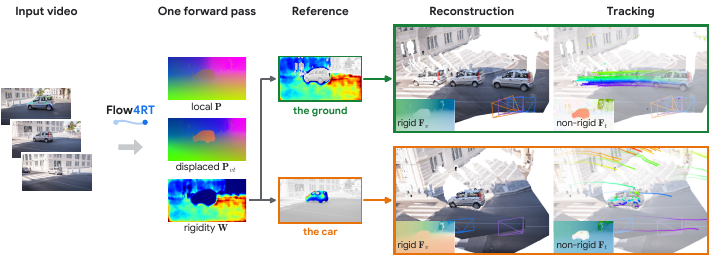}
    \caption{%
    \textbf{One forward pass, a selectable reference.}
    \method predicts, for all views jointly, local points $\mP$, displaced points $\mP_{vt}$ ($x,y,z \to$ R,G,B, histogram-equalized), and a rigidity map $\mW$ (red: high rigidity).
    Restricting $\mW$ to a chosen region (voting pixels: confident pixels of that region, each map normalized within its voters) and re-solving the closed-form alignment expresses reconstruction and tracking relative to that region, without running the network again.
    The alignment splits the scene flow $\mF = \mP_{vt} - \mP$ into the rigid flow $\mF_v$ it explains (inset, reconstruction) and the remaining non-rigid flow $\mF_t$ (inset, tracking, \cref{eq:decomp}).
    Flow maps color the 3D direction ($x,y,z \to$ R,G,B) and fade to white with decreasing magnitude on one scale.
    With the ground as reference, only the car moves and appears at three positions along its path.
    With the car as reference, the car is static, its observations fuse, and the street moves around it.
    }
    \label{fig:reference}
\end{figure*}

We now apply these pairwise operations across all input views to reconstruct a scene and track its points in a shared coordinate frame.
The geometry branch runs once over all $N$ views, while the motion branch runs on whichever ordered pairs are requested at inference.
We choose one input view as the anchor $a$ and express all reconstructed points and trajectories in its camera coordinate system.
Using stationary scene regions as the alignment reference, world-space reconstruction and tracking from $a$ are the collections
\begin{equation}
    \text{Reconstruction: } \left\{ \mP^{n}_{v_a} \right\}_{n=1}^{N} ,
    \qquad
    \text{Tracking: } \left\{ \mP^{a}_{t_n} \right\}_{n=1}^{N} ,
    \label{eq:recon_track}
\end{equation}
which require the pairs $(n, a)$ and $(a, n)$, respectively (\cref{fig:downstream-tasks}).
Here, $\mP^n_{v_a}$ is view $n$'s geometry transformed to the anchor viewpoint at its original time, while $\mP^a_{t_n}$ locates the anchor's points at time $t_n$ in the anchor coordinate frame.
Reconstruction uses $N$ predictions toward the anchor, while tracking uses $N$ predictions from the anchor, both sharing the self-pair $(a,a)$.
Local geometry is computed once per view and reused across pairs, which can be evaluated concurrently in a single batch.
The anchor and requested pairs are selected at inference, so the same trained model supports either task or both.

\paragraph{Choosing the reference region.}
The anchor view specifies the output coordinates at one instant, while the rigidity weights specify which scene region remains stationary across time.
By adjusting $\mW$ to select another rigid region and recomputing the alignment, we can express reconstruction and tracking relative to a moving object using the same dense predictions (\cref{fig:reference}).
In practice, we retain the learned $\mW$ on confident pixels inside an eroded object mask (from any segmentation model or user click) and re-solve \cref{eq:solve} without re-running the network.

\subsection{Architecture}
\label{sec:method_arch}

\paragraph{Geometry blocks.}
We convert the images into patch tokens and process them jointly with a vision transformer composed of geometry blocks (\cref{fig:pipeline}).
Early blocks attend within each view, and from a chosen depth onward, blocks alternate within-view and cross-view self-attention.
Cross-view attention provides shared context while each output point map remains in its own camera frame.
Tokens are read out at four depths to predict local geometry.

\paragraph{Motion blocks.}
At each of the four read-out depths, motion blocks map the source view's geometric tokens toward the target view (\cref{fig:pipeline}).
Each block has a cross-attention layer, with source tokens as queries and target tokens as keys and values, followed by a self-attention layer with AdaLN~\cite{dit}.
AdaLN is modulated by a token representing the target view's time.
This token starts from the same learned embedding in every view at the first cross-view block, then becomes view-specific as it is updated alongside the image tokens.
Because motion blocks branch off the geometry blocks, each pairwise prediction draws on context from all input views, including those outside the source--target pair.

\paragraph{Shared head.}
A single DPT head fuses tokens from the four depths coarse-to-fine into pixel-aligned maps.
It decodes the geometry branch into $(\mP, \mC)$ and the motion branch into $(\mP_{vt}, \mW)$, using the same head parameters for both.
Separate scalar activations produce confidence and rigidity in their respective ranges, and rigidity is normalized over pixels.
Local geometry is decoded once per view, while displaced geometry is decoded for each requested ordered pair.
Each motion block attends only within its pair, so the motion branch scales linearly with the number of requested pairs ($2N$ for the anchor configuration).

\begin{table*}[t]
\centering
\caption{\textbf{World-space 3D tracking and reconstruction on WorldTrack.}
\method is the only method in the top two on both metrics for both tasks and leads overall, showing that one shared head with closed-form alignment achieves balanced 4D accuracy.
Overall averages the tracking and reconstruction averages, and shading marks the top three per column.}
\label{tab:worldtrack}
\footnotesize
\setlength{\tabcolsep}{1.6pt}
\begin{tabular}{@{}l@{\hspace{3pt}}cc@{\hspace{3pt}}cc@{\hspace{3pt}}cc@{\hspace{3pt}}cc@{\hspace{3pt}}cc@{\hspace{6pt}}cc@{\hspace{3pt}}cc@{\hspace{3pt}}cc@{\hspace{6pt}}cc@{}}
\toprule
 & \multicolumn{10}{c}{Tracking} & \multicolumn{6}{c}{Reconstruction} & \multicolumn{2}{c}{\textbf{Overall}} \\
\cmidrule(lr){2-11}\cmidrule(lr){12-17}\cmidrule(lr){18-19}
 & \multicolumn{2}{c}{ADT} & \multicolumn{2}{c}{DR} & \multicolumn{2}{c}{PO} & \multicolumn{2}{c}{PS} & \multicolumn{2}{c}{\textbf{Average}} & \multicolumn{2}{c}{PO} & \multicolumn{2}{c}{TUM} & \multicolumn{2}{c}{\textbf{Average}} & \multicolumn{2}{c}{\textbf{Average}} \\
\cmidrule(lr){2-3}\cmidrule(lr){4-5}\cmidrule(lr){6-7}\cmidrule(lr){8-9}\cmidrule(lr){10-11}\cmidrule(lr){12-13}\cmidrule(lr){14-15}\cmidrule(lr){16-17}\cmidrule(lr){18-19}
 & APD\,$\uparrow$ & EPE\,$\downarrow$ & APD\,$\uparrow$ & EPE\,$\downarrow$ & APD\,$\uparrow$ & EPE\,$\downarrow$ & APD\,$\uparrow$ & EPE\,$\downarrow$ & APD\,$\uparrow$ & EPE\,$\downarrow$ & APD\,$\uparrow$ & EPE\,$\downarrow$ & APD\,$\uparrow$ & EPE\,$\downarrow$ & APD\,$\uparrow$ & EPE\,$\downarrow$ & APD\,$\uparrow$ & EPE\,$\downarrow$ \\
\midrule
St4RTrack~\cite{st4rtrack} & 76.0 & 0.268 & 73.7 & 0.268 & 68.0 & 0.314 & 69.7 & 0.264 & 71.8 & 0.278 & 78.7 & 0.205 & 83.4 & 0.185 & 81.1 & 0.195 & 76.5 & 0.237 \\
Flow4R~\cite{flow4r} & 78.6 & 0.237 & 78.5 & 0.223 & 71.1 & 0.279 & 64.3 & 0.313 & 73.1 & 0.263 & 81.0 & 0.182 & 79.9 & 0.202 & 80.4 & 0.192 & 76.8 & 0.228 \\
Any4D~\cite{any4d} & 58.5 & 0.408 & 77.8 & 0.231 & 65.2 & 0.413 & 63.8 & 0.300 & 66.3 & 0.338 & 75.1 & 0.221 & 80.2 & 0.240 & 77.6 & 0.230 & 72.0 & 0.284 \\
V-DPM~\cite{vdpm} & 85.7 & 0.170 & 75.3 & 0.249 & \cellcolor{rank3}81.3 & \cellcolor{rank3}0.189 & 78.8 & \cellcolor{rank3}0.180 & 80.3 & \cellcolor{rank3}0.197 & 87.5 & 0.130 & \cellcolor{rank2}90.2 & \cellcolor{rank2}0.136 & \cellcolor{rank3}88.8 & \cellcolor{rank3}0.133 & 84.6 & \cellcolor{rank3}0.165 \\
4RC~\cite{fourrc} & 83.2 & 0.190 & \cellcolor{rank2}82.3 & \cellcolor{rank2}0.195 & 79.3 & 0.282 & 69.4 & 0.256 & 78.6 & 0.231 & \cellcolor{rank1}97.0 & \cellcolor{rank1}0.070 & 86.7 & \cellcolor{rank3}0.154 & \cellcolor{rank1}91.9 & \cellcolor{rank1}0.112 & \cellcolor{rank2}85.3 & 0.172 \\
Track4World~\cite{track4world} & 79.9 & 0.217 & 71.2 & 0.274 & 66.3 & 0.363 & 73.0 & 0.232 & 72.6 & 0.272 & 81.0 & 0.185 & \cellcolor{rank3}89.0 & \cellcolor{rank1}0.125 & 85.0 & 0.155 & 78.8 & 0.214 \\
OmniX~\cite{omnix} & \cellcolor{rank1}91.7 & \cellcolor{rank1}0.103 & 78.3 & 0.230 & 75.7 & 0.303 & 77.3 & 0.199 & 80.8 & 0.209 & \cellcolor{rank3}91.5 & \cellcolor{rank3}0.108 & 86.8 & 0.163 & \cellcolor{rank2}89.1 & 0.135 & \cellcolor{rank3}85.0 & 0.172 \\
SM4RT~\cite{sm4rt} & 87.5 & 0.146 & \cellcolor{rank3}80.9 & 0.205 & 77.5 & 0.297 & \cellcolor{rank3}79.3 & 0.182 & \cellcolor{rank3}81.3 & 0.207 & 90.8 & 0.117 & 79.4 & 0.196 & 85.1 & 0.156 & 83.2 & 0.182 \\
Point4D~\cite{point4d} & \cellcolor{rank2}90.6 & \cellcolor{rank3}0.134 & 77.8 & 0.244 & 69.5 & 0.370 & 74.0 & 0.228 & 78.0 & 0.244 & 81.5 & 0.184 & 85.3 & 0.218 & 83.4 & 0.201 & 80.7 & 0.223 \\
UniQuery4R~\cite{uniquery4r} & 87.1 & 0.138 & 80.8 & \cellcolor{rank3}0.201 & \cellcolor{rank1}82.1 & \cellcolor{rank1}0.175 & \cellcolor{rank1}84.1 & \cellcolor{rank1}0.137 & \cellcolor{rank2}83.5 & \cellcolor{rank1}0.163 & 89.3 & 0.111 & 84.8 & 0.172 & 87.0 & 0.141 & \cellcolor{rank2}85.3 & \cellcolor{rank2}0.152 \\
\midrule
\textbf{\method} & \cellcolor{rank3}88.7 & \cellcolor{rank2}0.133 & \cellcolor{rank1}84.9 & \cellcolor{rank1}0.170 & \cellcolor{rank2}81.8 & \cellcolor{rank2}0.185 & \cellcolor{rank2}79.5 & \cellcolor{rank2}0.172 & \cellcolor{rank1}83.7 & \cellcolor{rank2}0.165 & \cellcolor{rank2}93.4 & \cellcolor{rank2}0.099 & \cellcolor{rank1}90.4 & \cellcolor{rank3}0.154 & \cellcolor{rank1}91.9 & \cellcolor{rank2}0.126 & \cellcolor{rank1}87.8 & \cellcolor{rank1}0.146 \\
\bottomrule
\end{tabular}
\end{table*}

\subsection{Training}
\label{sec:method_training}

We train on static and dynamic data with five objectives:
\begin{equation}
    \mathcal{L} = \lambda_1 \mathcal{L}_{\mP} + \lambda_2 \mathcal{L}_{\mP_{vt}} + \lambda_3 \mathcal{L}_{\mP_v} + \lambda_4 \mathcal{L}_{\mW} + \lambda_5 \mathcal{L}_{\mathcal{H}} ,
    \label{eq:loss_total}
\end{equation}
where $\mathcal{L}_{\mP}$ supervises local geometry and $\mathcal{L}_{\mP_{vt}}$ supervises displaced geometry from scene flow and 3D tracks, both weighted by confidence $\mC$.
The auxiliary term $\mathcal{L}_{\mP_v}$ supervises camera-induced displacement from ground-truth poses, with detached rigidity weights gating moving areas on dynamic sequences.
Finally, $\mathcal{L}_{\mW}$ supervises rigidity through the differentiable alignment in \cref{eq:solve} with gradients to the point maps blocked, learning reliable alignment weights without rigidity annotations, and $\mathcal{L}_{\mathcal{H}}$ maximizes the entropy of $\mW$ to keep it from collapsing onto a few pixels.
We set $\lambda_1=1$, $\lambda_2=\lambda_3=\lambda_4=0.5$, and $\lambda_5=10^{-3}$, with scale-normalized point maps and confidence regularization weight $\alpha=0.2$.
For each training sample, a random derangement assigns every view a target, predicting both directions of each pair.
\Cref{app:training_objectives} details the losses, and \cref{fig:downstream-tasks} illustrates their supervision paths.

\section{Experiments}
\label{sec:experiments}

We evaluate \method on challenging dynamic reconstruction and tracking benchmarks, assessing whether our unified geometry-and-motion representation achieves robust 4D perception.
We further analyze the core drivers of performance: joint multi-view context and learned rigidity.

\subsection{Setup}
\label{sec:exp_setup}

\begin{figure*}[t]
\centering
\ifdefined\qualbox\else\newsavebox\qualbox\fi
\setlength{\tabcolsep}{1pt}
\begin{tabular}{@{}cc@{}}
\footnotesize Input & \begin{tabular}{@{}cccc@{}}\makebox[0.185\linewidth]{\footnotesize 4RC} & \makebox[0.185\linewidth]{\footnotesize OmniX} & \makebox[0.185\linewidth]{\footnotesize UniQuery4R} & \makebox[0.185\linewidth]{\footnotesize \textbf{\method}}\end{tabular} \\
\global\setbox\qualbox\hbox{\begin{tabular}[b]{@{}cccc@{}}\includegraphics[width=0.185\linewidth]{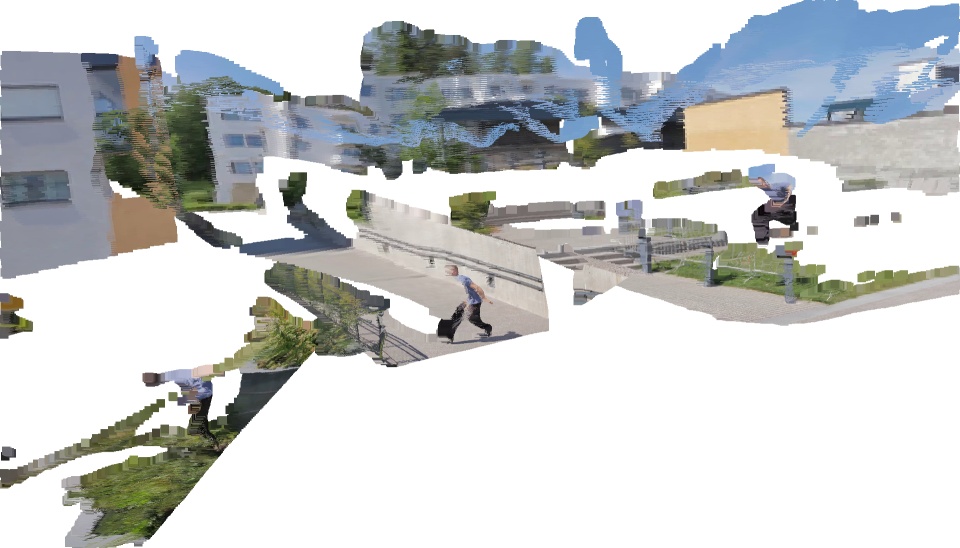} & \includegraphics[width=0.185\linewidth]{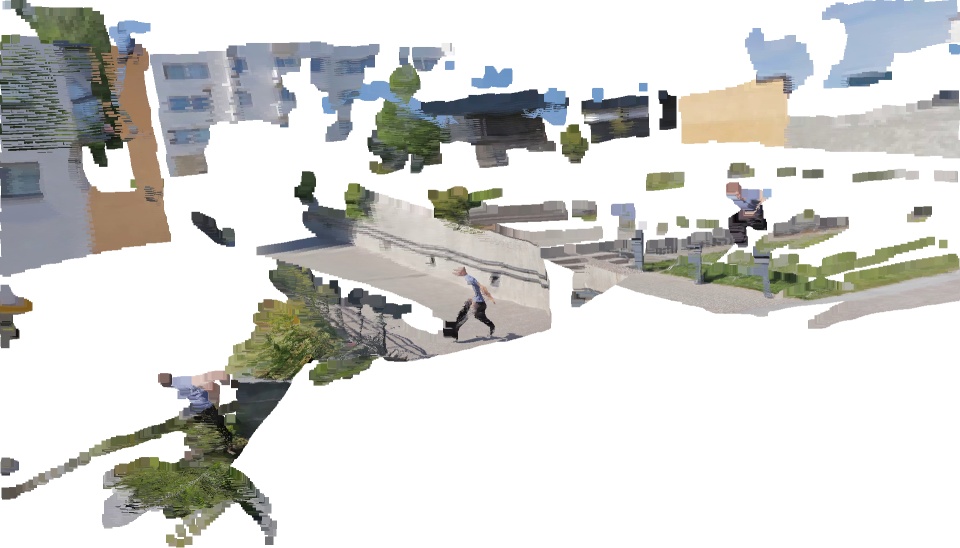} & \includegraphics[width=0.185\linewidth]{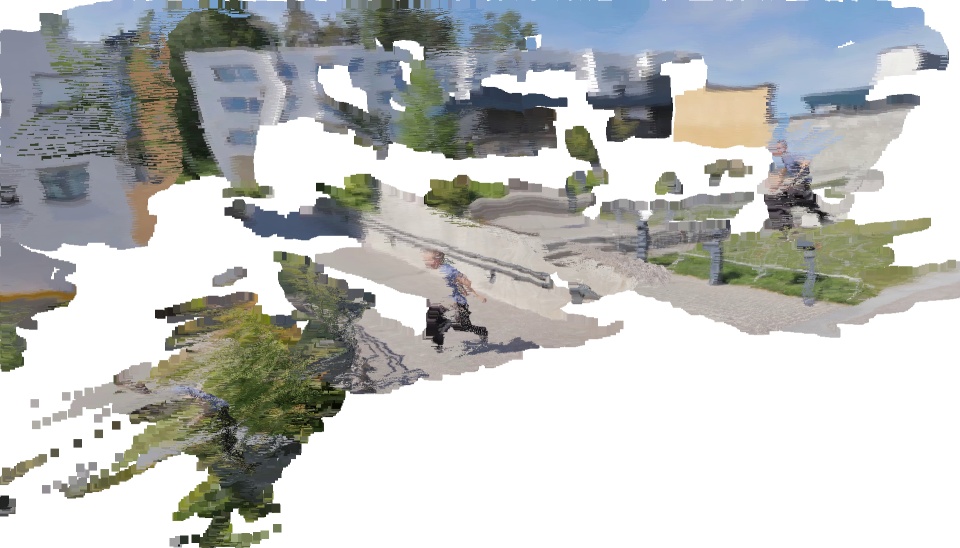} & \includegraphics[width=0.185\linewidth]{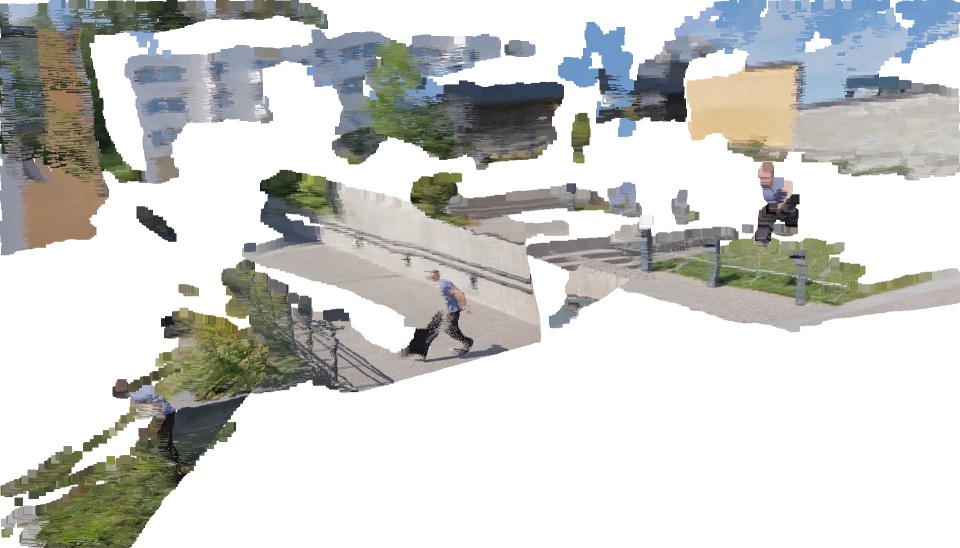} \\[-2pt] \includegraphics[width=0.185\linewidth]{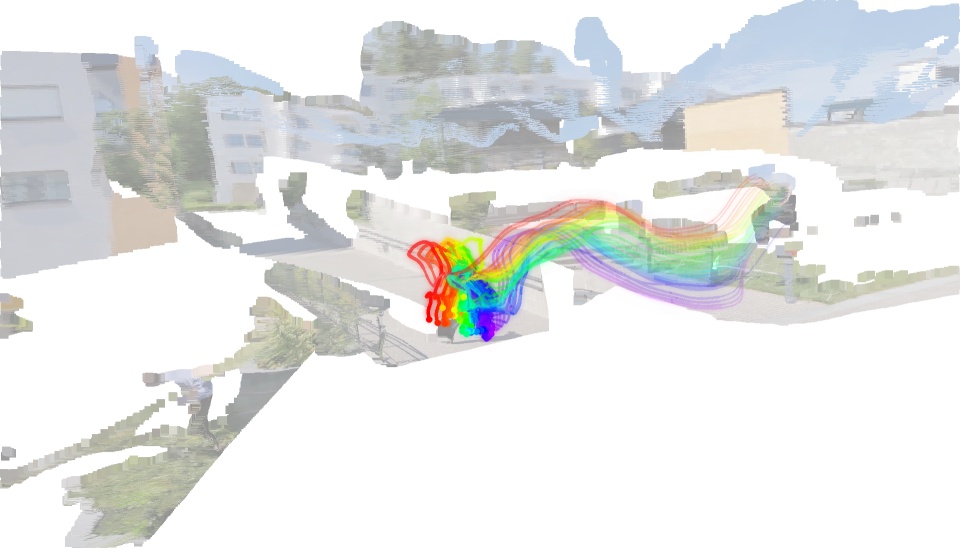} & \includegraphics[width=0.185\linewidth]{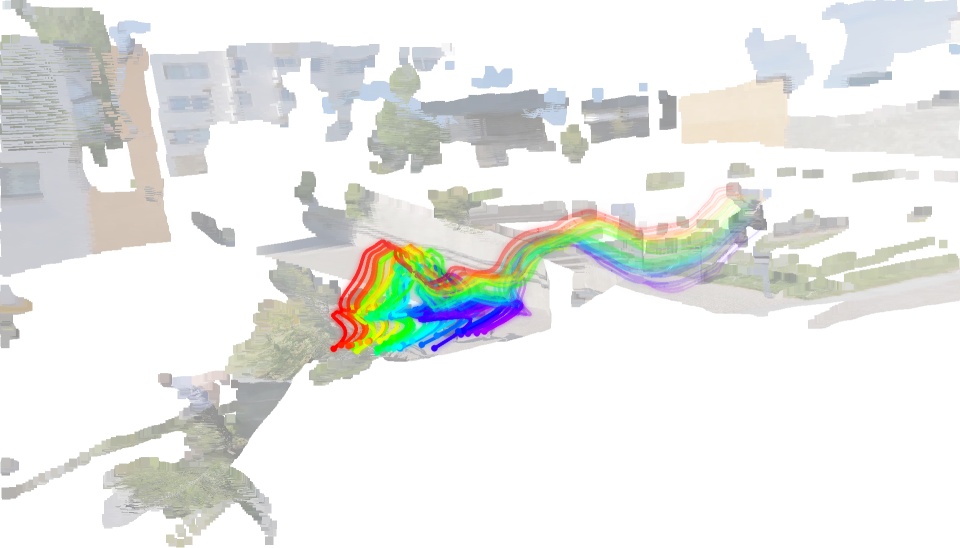} & \includegraphics[width=0.185\linewidth]{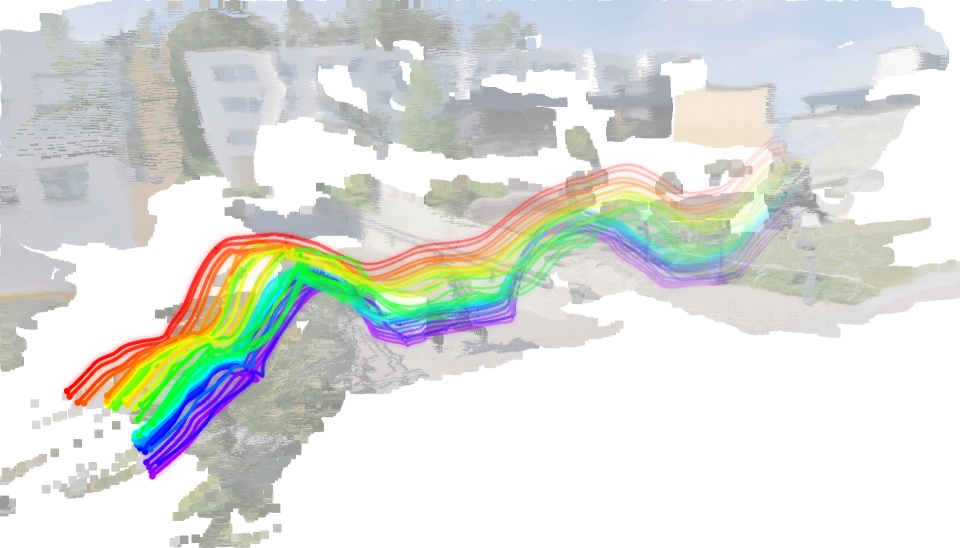} & \includegraphics[width=0.185\linewidth]{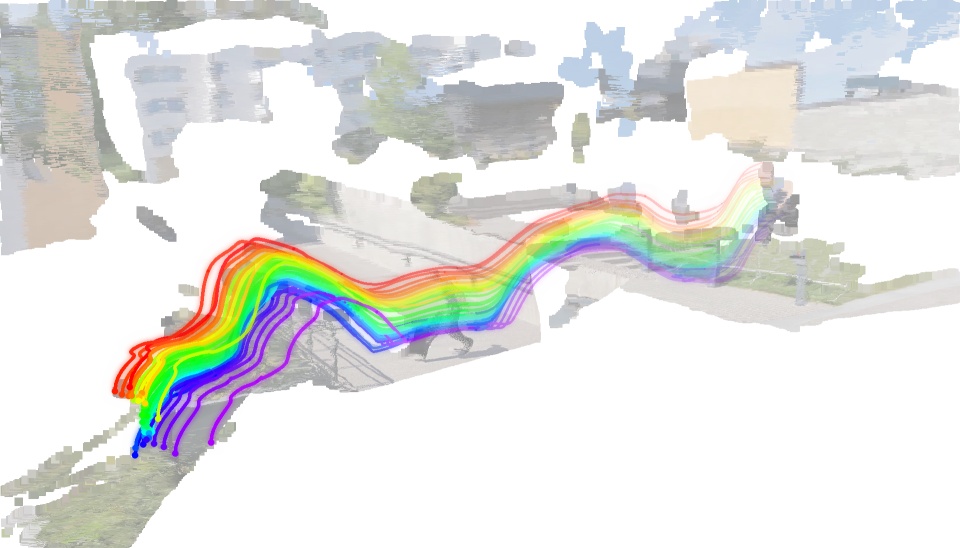} \end{tabular}}%
\raisebox{-\dp\qualbox}{\includegraphics[height=\dimexpr\ht\qualbox+\dp\qualbox\relax]{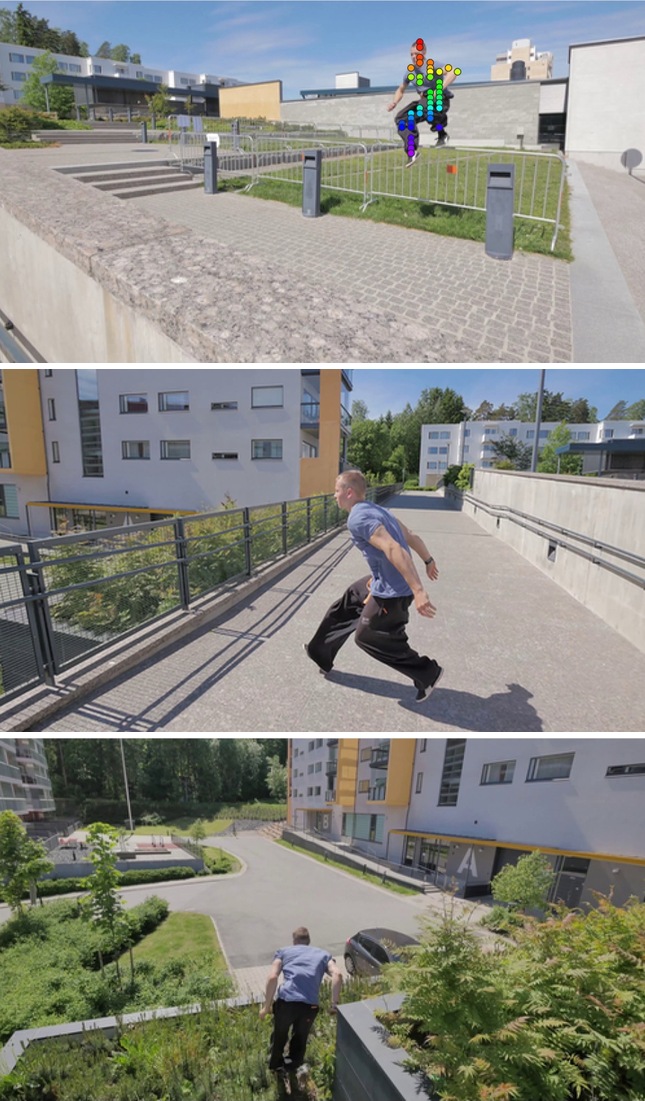}} & \usebox\qualbox \\[3pt]
\global\setbox\qualbox\hbox{\begin{tabular}[b]{@{}cccc@{}}\includegraphics[width=0.185\linewidth]{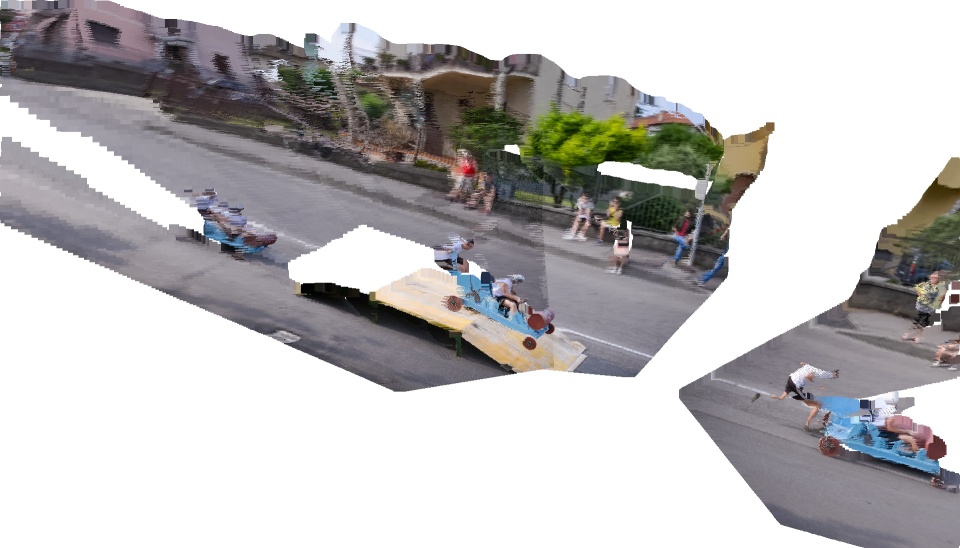} & \includegraphics[width=0.185\linewidth]{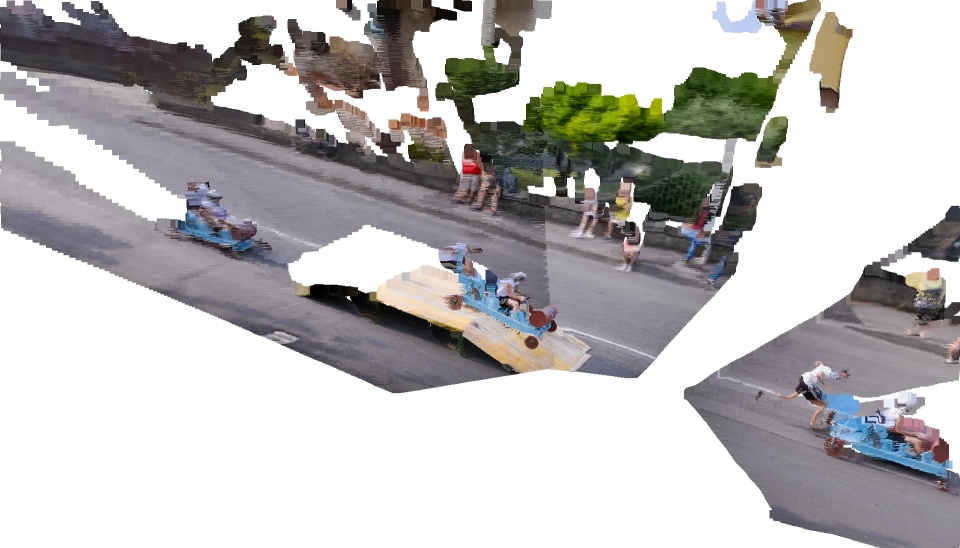} & \includegraphics[width=0.185\linewidth]{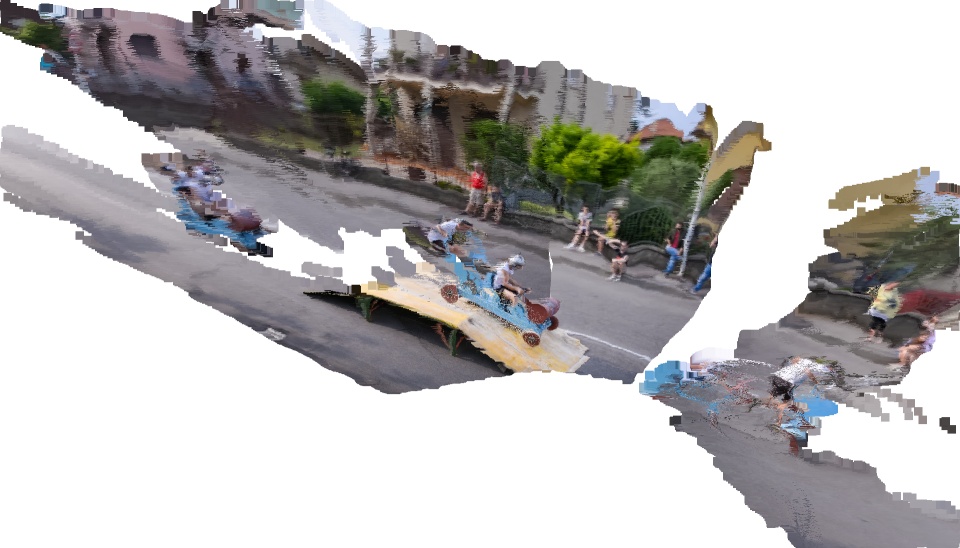} & \includegraphics[width=0.185\linewidth]{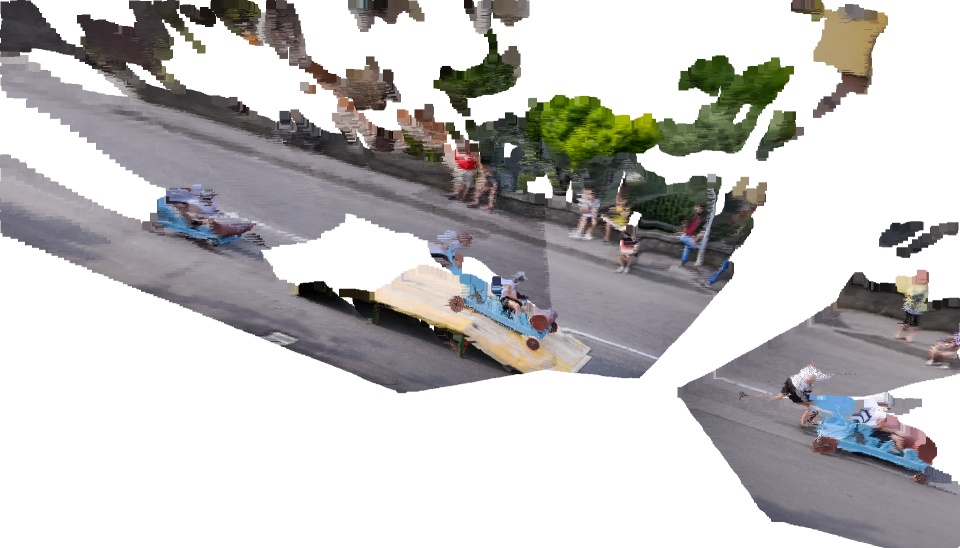} \\[-2pt] \includegraphics[width=0.185\linewidth]{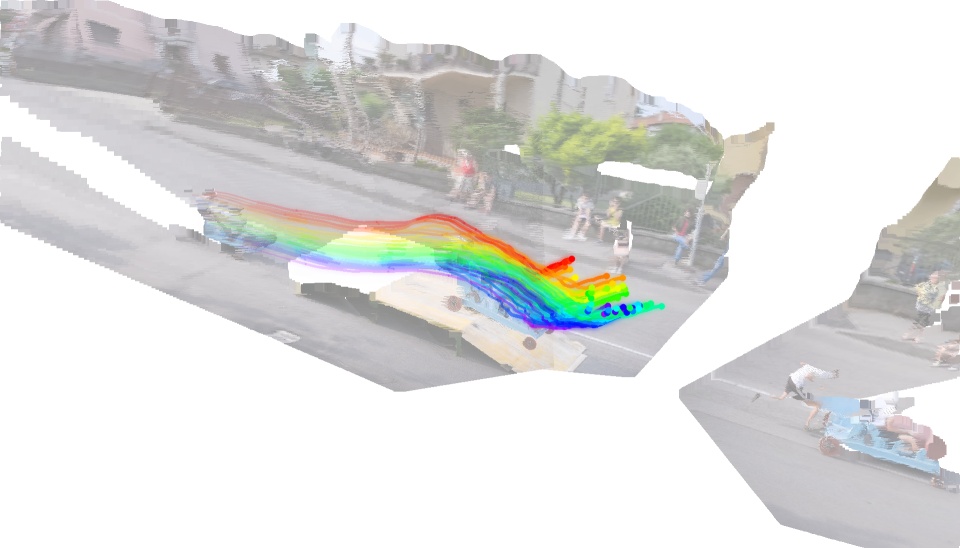} & \includegraphics[width=0.185\linewidth]{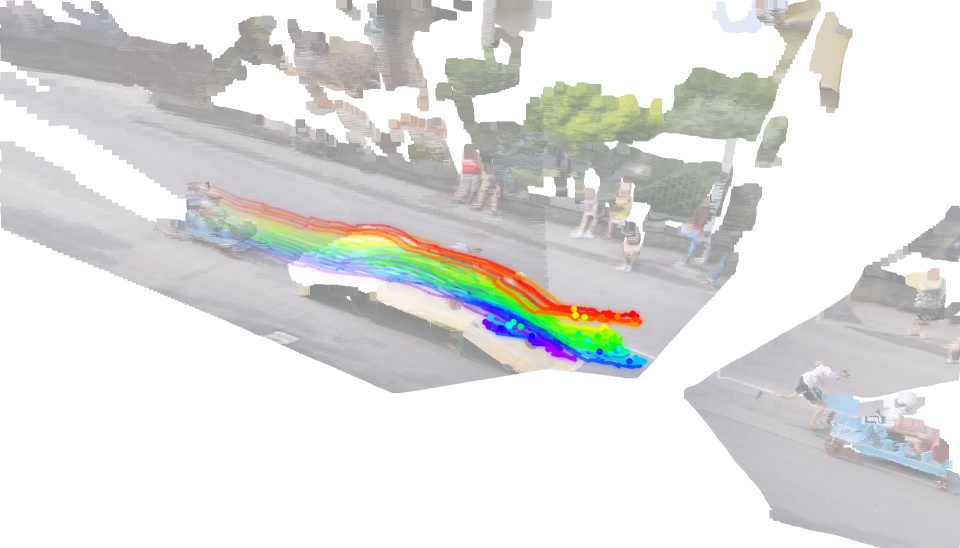} & \includegraphics[width=0.185\linewidth]{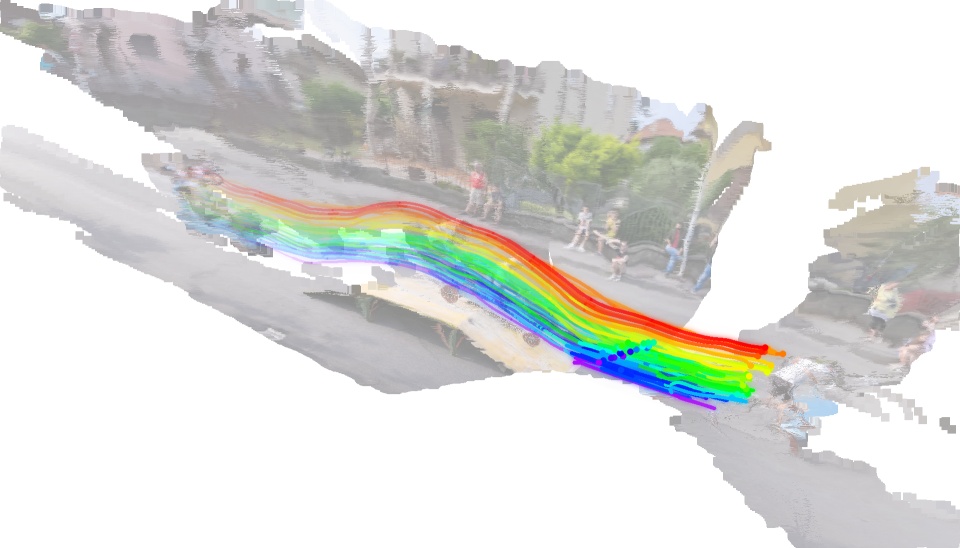} & \includegraphics[width=0.185\linewidth]{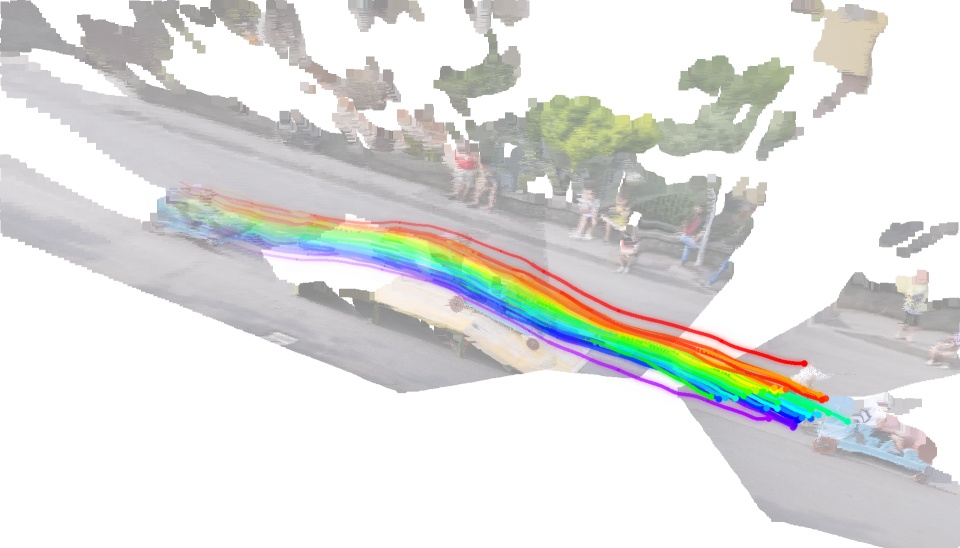} \end{tabular}}%
\raisebox{-\dp\qualbox}{\includegraphics[height=\dimexpr\ht\qualbox+\dp\qualbox\relax]{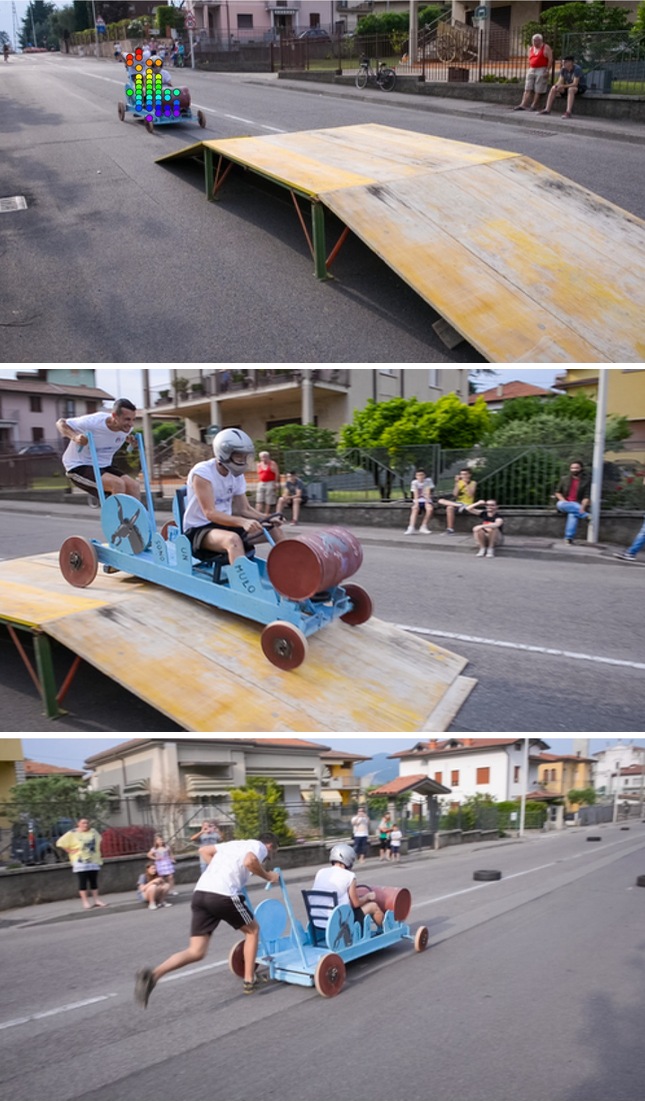}} & \usebox\qualbox \\[3pt]
\global\setbox\qualbox\hbox{\begin{tabular}[b]{@{}cccc@{}}\includegraphics[width=0.185\linewidth]{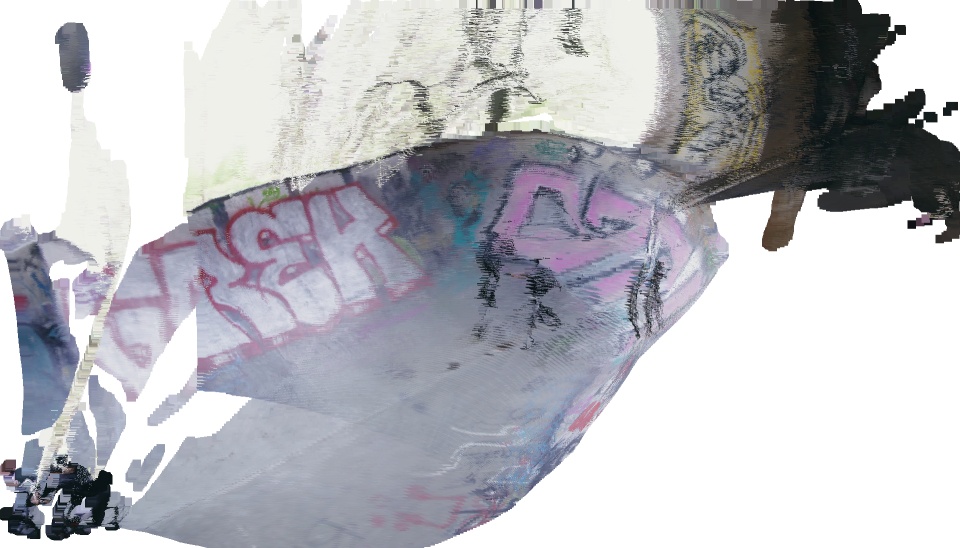} & \includegraphics[width=0.185\linewidth]{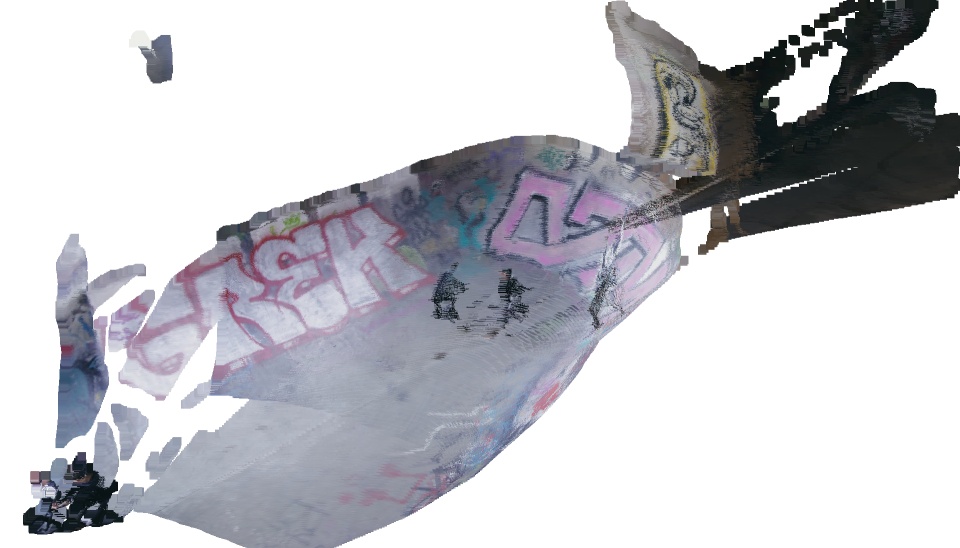} & \includegraphics[width=0.185\linewidth]{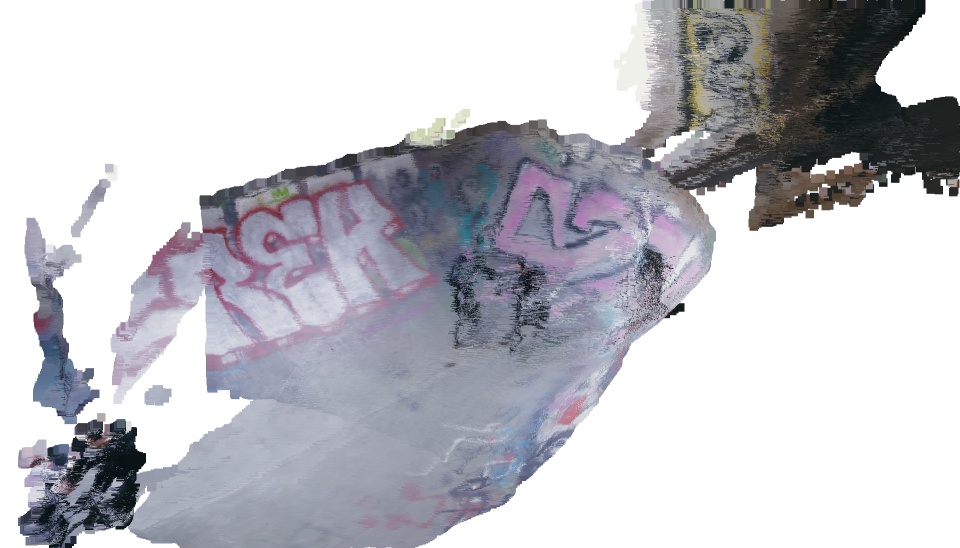} & \includegraphics[width=0.185\linewidth]{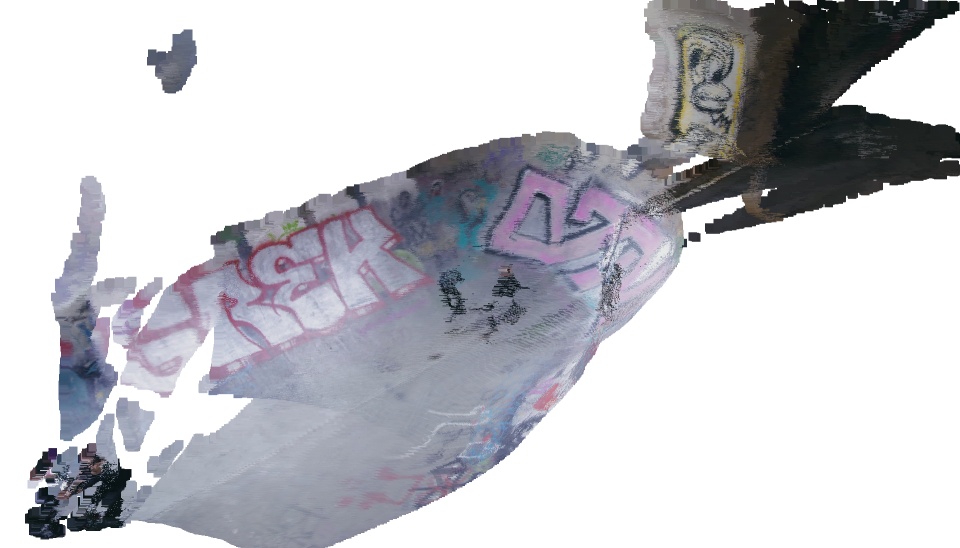} \\[-2pt] \includegraphics[width=0.185\linewidth]{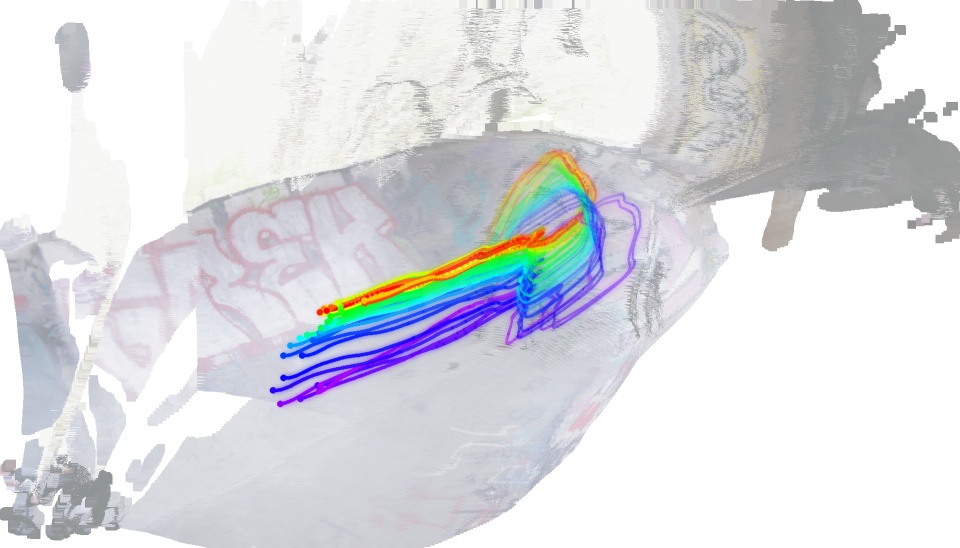} & \includegraphics[width=0.185\linewidth]{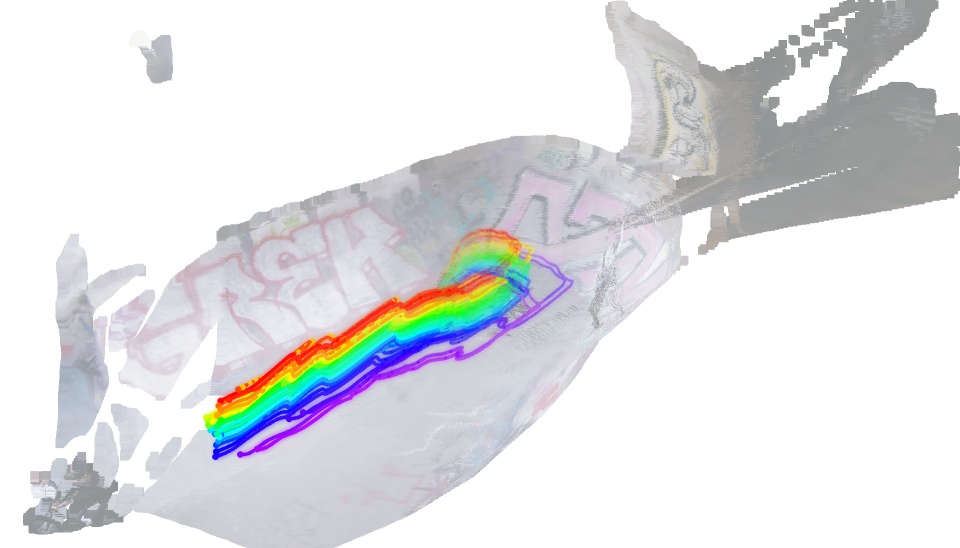} & \includegraphics[width=0.185\linewidth]{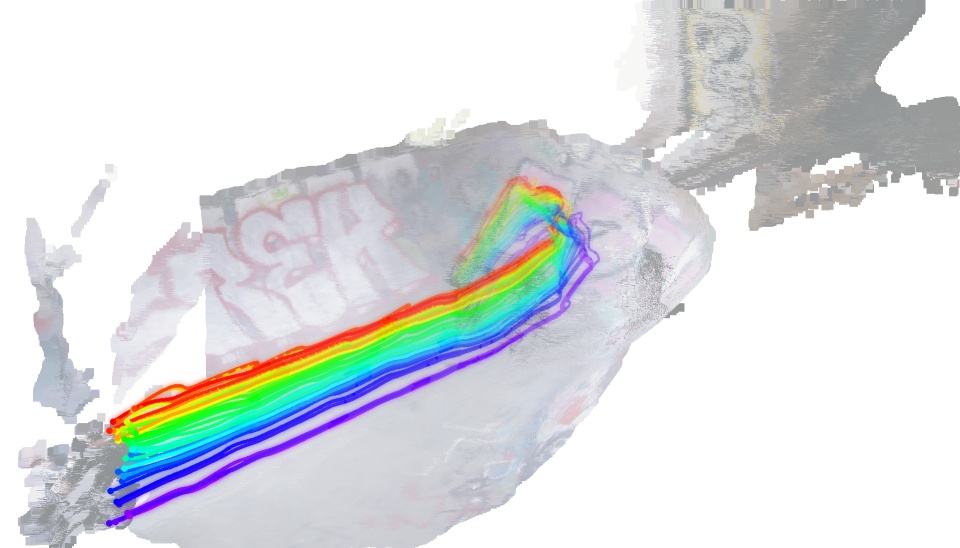} & \includegraphics[width=0.185\linewidth]{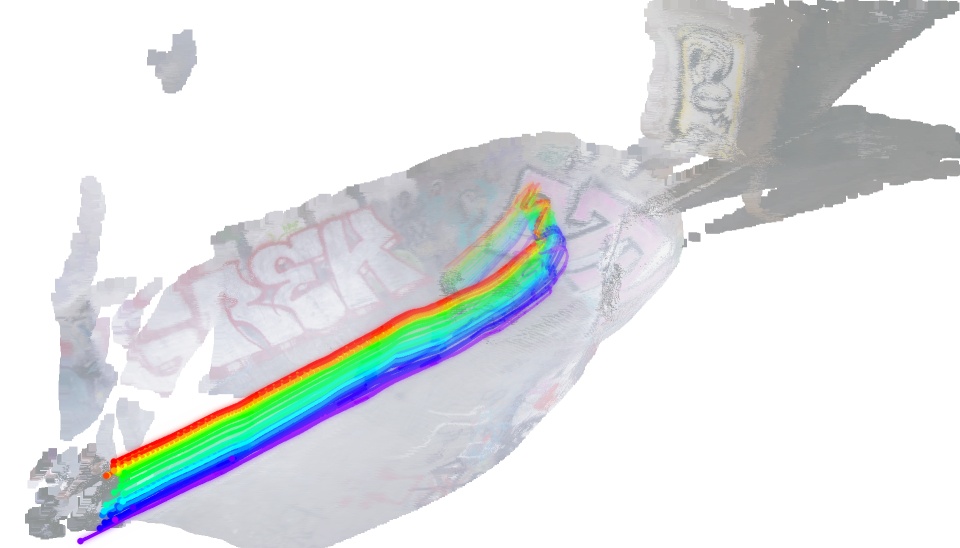} \end{tabular}}%
\raisebox{-\dp\qualbox}{\includegraphics[height=\dimexpr\ht\qualbox+\dp\qualbox\relax]{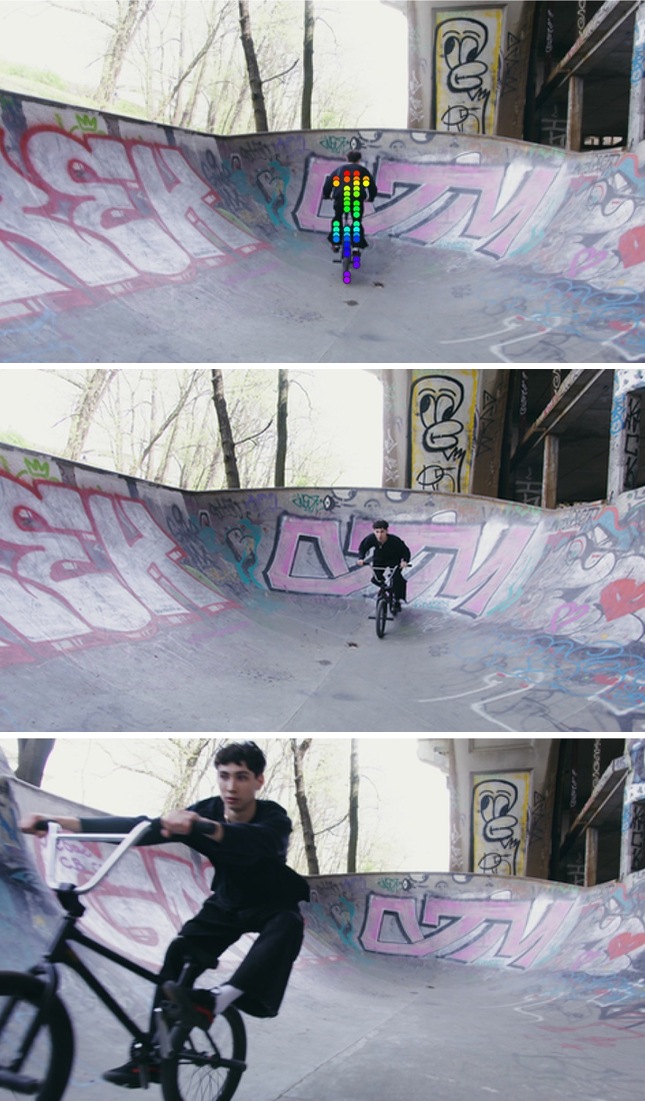}} & \usebox\qualbox \\[3pt]
\global\setbox\qualbox\hbox{\begin{tabular}[b]{@{}cccc@{}}\includegraphics[width=0.185\linewidth]{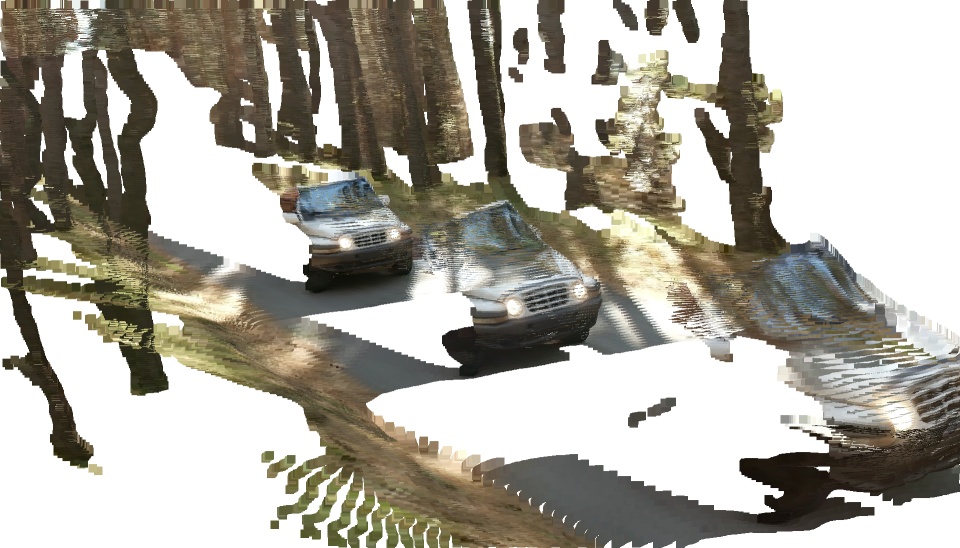} & \includegraphics[width=0.185\linewidth]{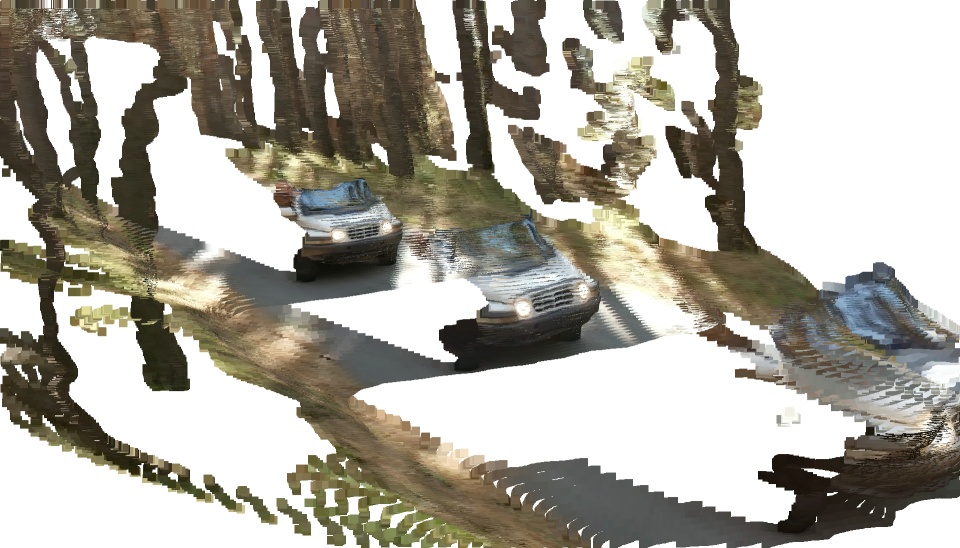} & \includegraphics[width=0.185\linewidth]{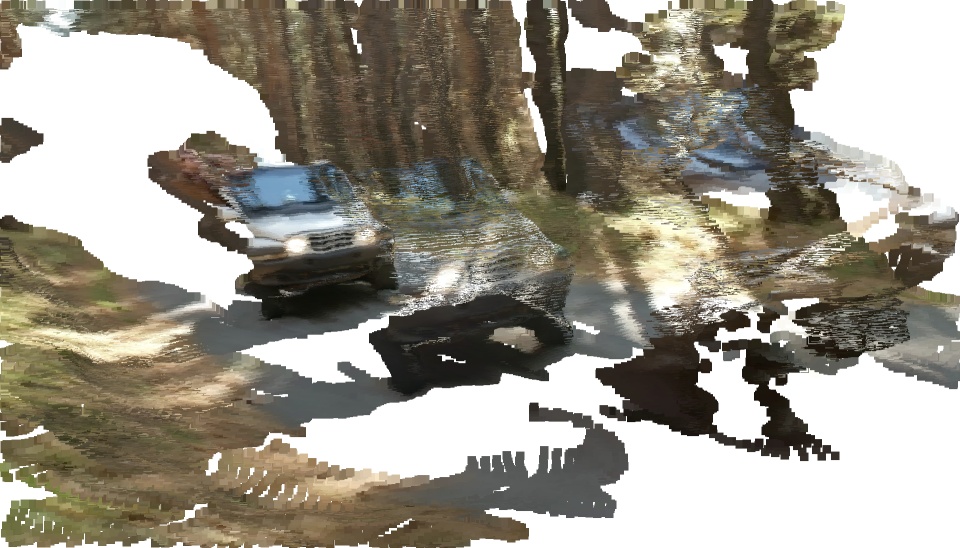} & \includegraphics[width=0.185\linewidth]{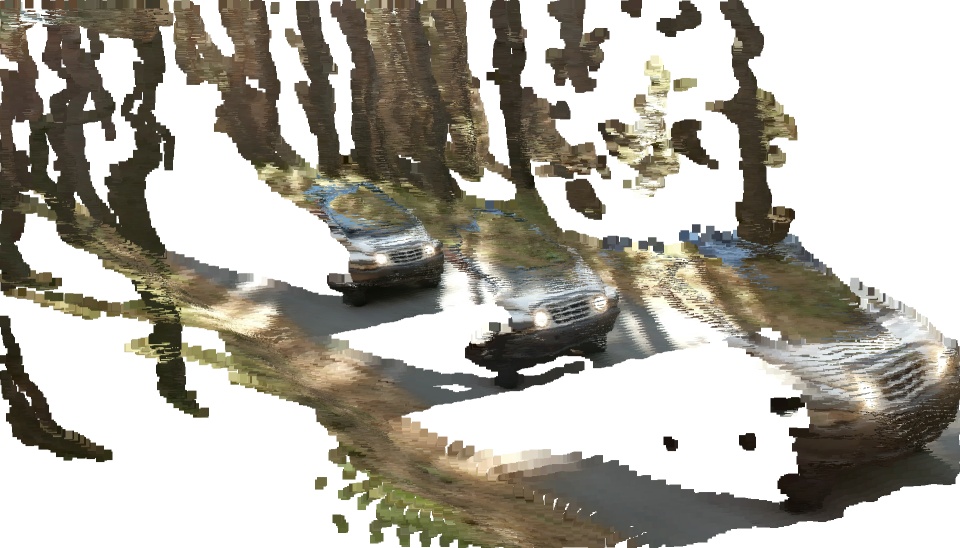} \\[-2pt] \includegraphics[width=0.185\linewidth]{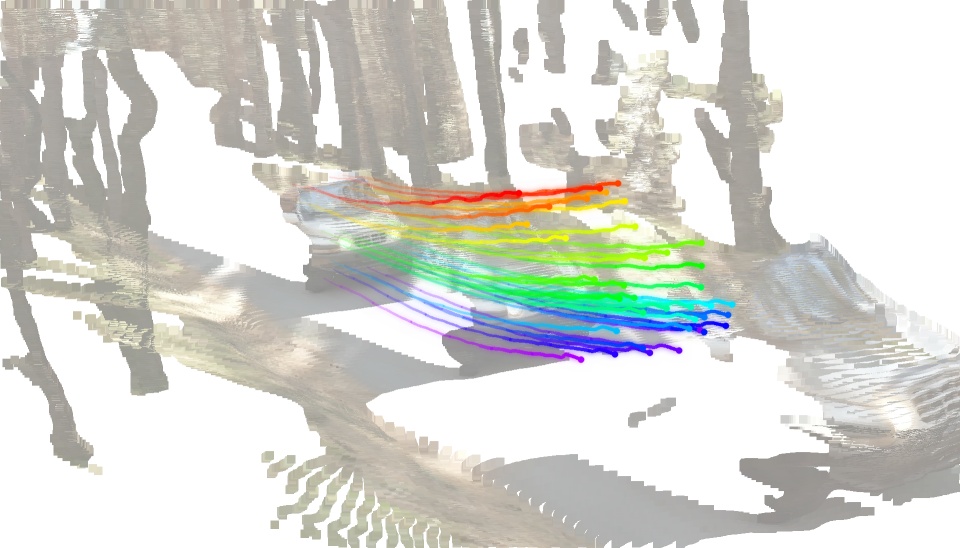} & \includegraphics[width=0.185\linewidth]{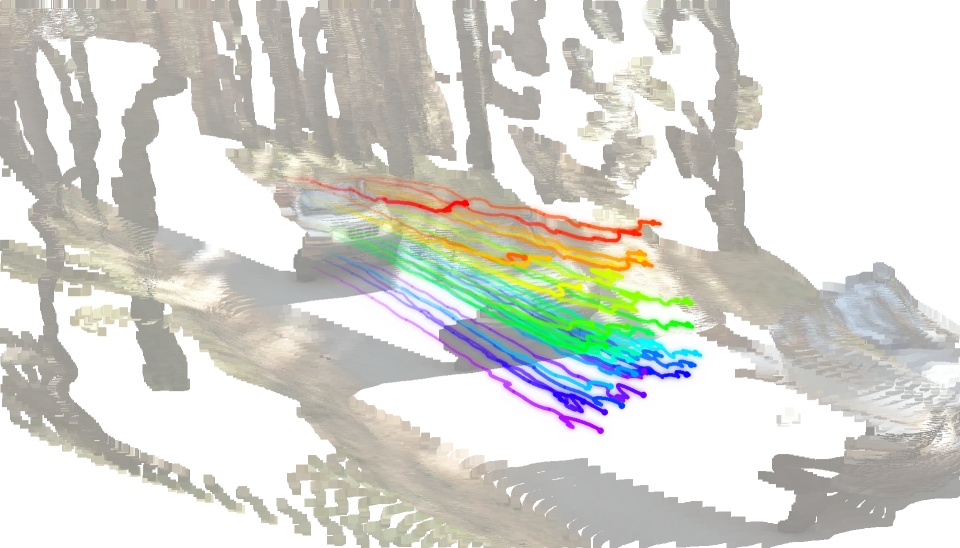} & \includegraphics[width=0.185\linewidth]{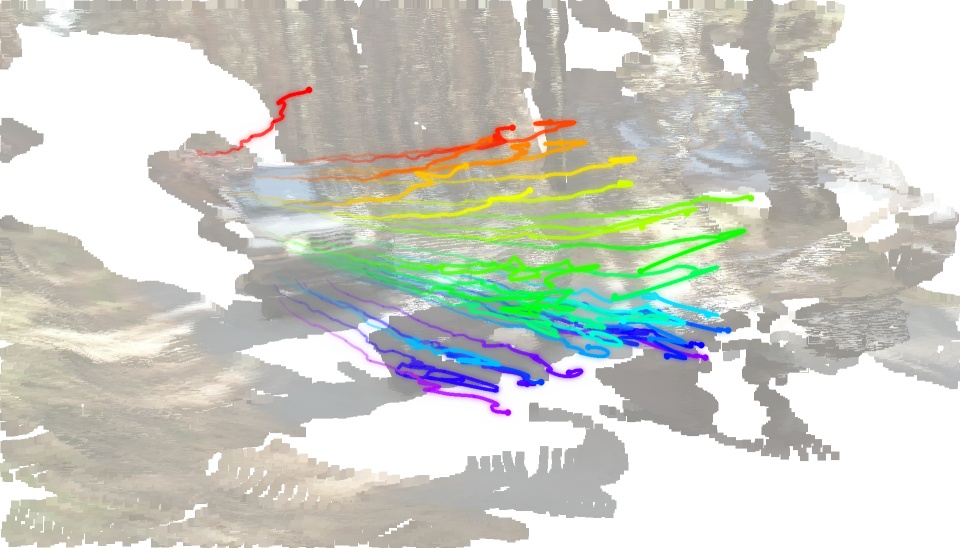} & \includegraphics[width=0.185\linewidth]{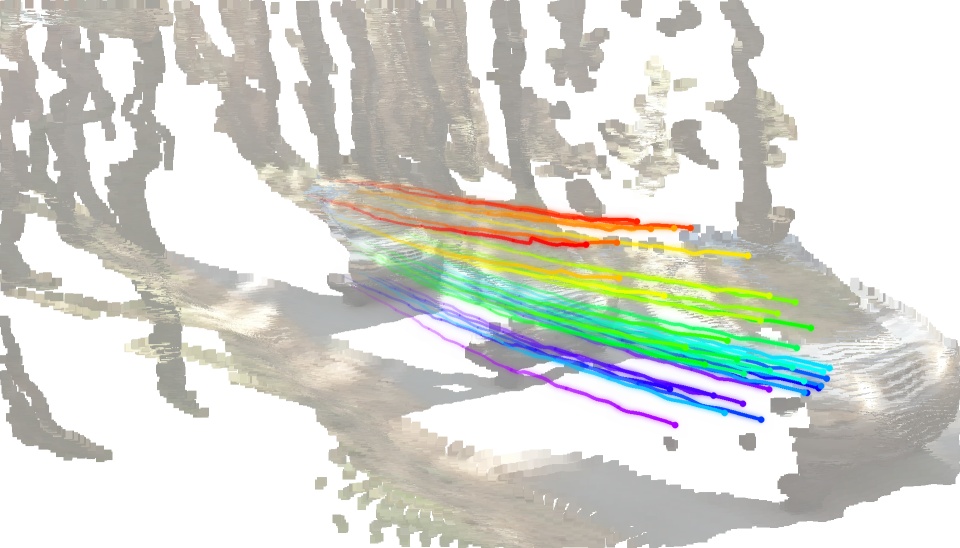} \end{tabular}}%
\raisebox{-\dp\qualbox}{\includegraphics[height=\dimexpr\ht\qualbox+\dp\qualbox\relax]{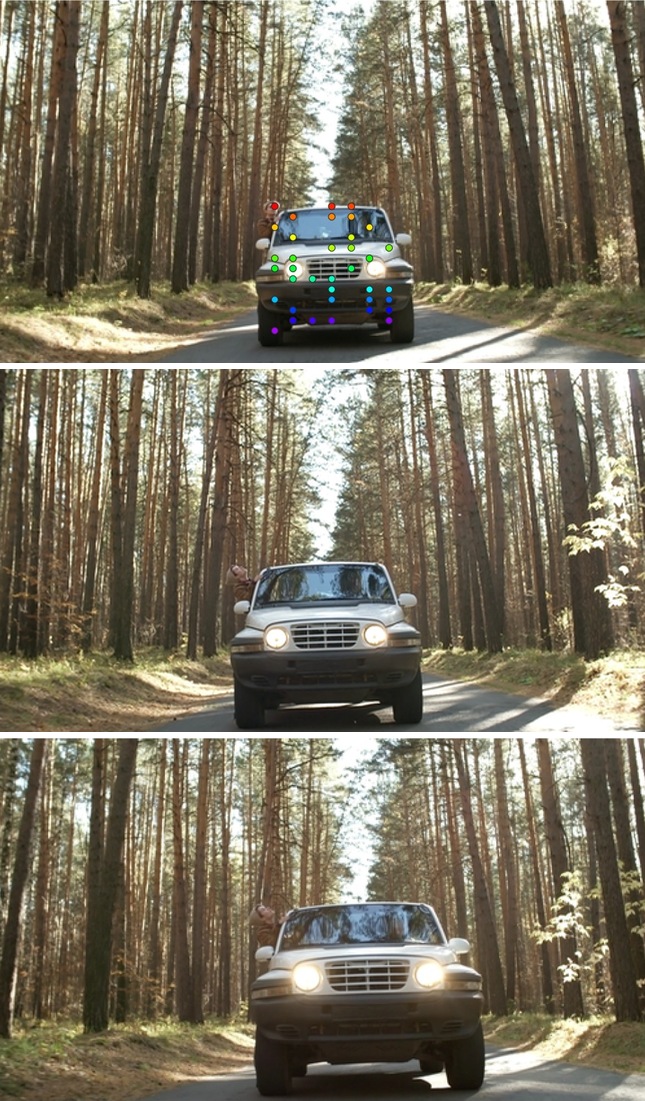}} & \usebox\qualbox
\end{tabular}
\vspace{-0.6em}
\caption{\textbf{4D reconstruction and world-space tracking.}
For each clip, the upper row stacks the predicted point maps of the first, middle, and last frames in world space, and the lower row draws the predicted world trajectories of first-frame query points.
The input column shows the first, middle, and last frames, with the query points (from the first-frame object mask) colored to match their trajectories.
All methods share one virtual camera, one set of query points, and one global scale.
Clips: DAVIS~\cite{davis} \emph{parkour} and \emph{soapbox}, and Pexels BMX bowl and forest car (videos courtesy of Pexels).}
\label{fig:qual}
\end{figure*}

\paragraph{Training data.}
We train on a mixture of static and dynamic, real and synthetic datasets, listed with their supervision and samples per epoch in \cref{tab:training_data}.

\paragraph{Implementation details.}
We initialize the backbone from 4RC~\cite{fourrc} and train the motion blocks and shared DPT head from scratch.
The model has 1.73\,B parameters and is trained for 100 epochs on 8 H100 GPUs, with 352 views per step at a long side of 518 pixels.
Architecture settings and the training schedule are detailed in \cref{app:implementation}.

\paragraph{Benchmark.}
We evaluate in world coordinates on the WorldTrack benchmark~\cite{st4rtrack}.
Tracking is evaluated on two real datasets, Aria Digital Twin (ADT)~\cite{adt} and Panoptic Studio (PS)~\cite{panoptic}, and two synthetic ones, Point Odyssey (PO)~\cite{pointodyssey} and Dynamic Replica (DR)~\cite{dynamicstereo}, each with 50 sequences over their first 64 frames.
Reconstruction is evaluated on the synthetic Point Odyssey and the real TUM-Dynamics (TUM)~\cite{tumrgbd}.

\paragraph{Metrics.}
Following St4RTrack~\cite{st4rtrack}, we report the Average Percentage of 3D Points within Delta (APD): the fraction of points within $\delta_\text{3D} \in \{0.1, 0.3, 0.5, 1.0\}$ of the ground truth, averaged over thresholds and frames.
We also report the Euclidean End-Point Error (EPE), which measures how large the residual is rather than only whether it crosses a threshold.
Before scoring, predictions and ground truth are each normalized by their global median per sequence, so thresholds and errors are in normalized units.

\paragraph{Baselines.}
We compare against feedforward methods that output world-space geometry and motion without per-scene optimization, including St4RTrack~\cite{st4rtrack} and Flow4R~\cite{flow4r} as representative two-view methods, and Any4D~\cite{any4d}, V-DPM~\cite{vdpm}, 4RC~\cite{fourrc}, Track4World~\cite{track4world}, OmniX~\cite{omnix}, SM4RT~\cite{sm4rt}, Point4D~\cite{point4d}, and UniQuery4R~\cite{uniquery4r} as multi-view methods, which differ in how time and viewpoint enter.
For UniQuery4R, we score reconstruction on the world-space point maps of its global point head, the counterpart of the world-space point maps of the other methods.
All multi-view methods receive frames resized to a long side of 518 pixels, except Track4World, which keeps the canvases of its own evaluation, up to $360\times640$.
St4RTrack and Flow4R run at their native long side of 512 pixels.
We do not compare with D4RT~\cite{d4rt}, whose code and weights are not public, or with SpatialTrackerV2~\cite{spatialtrackerv2}, which trails all our baselines in the WorldTrack tracking evaluation of UniQuery4R~\cite{uniquery4r} (56.8 average APD, vs.\ 71.3 for St4RTrack).

\subsection{World-Space Tracking and Reconstruction}
\label{sec:exp_main}

\paragraph{Tracking.}
Evaluating tracks in world coordinates tests the full pipeline simultaneously, compounding errors from local geometry, camera motion, and object displacement (\cref{tab:worldtrack}).
\method achieves the highest average tracking APD and second-lowest EPE, showing that a dense head shared with geometry captures complex 3D motion without a dedicated tracking decoder.
While query-based UniQuery4R is competitive on PO and PS, \method leads clearly on ADT and DR, and outperforms the remaining multi-view baselines by wide margins.

\paragraph{Reconstruction.}
\cref{tab:worldtrack} also evaluates world-space point clouds aligned to the anchor frame, where prior methods exhibit strong task or domain trade-offs.
4RC's regressed camera poses excel on synthetic Point Odyssey but drop on real TUM sequences, while UniQuery4R trails on reconstruction despite strong tracking.
By grounding camera pose in rigid scene alignment, \method ties for the highest average reconstruction APD, leads on real TUM sequences, and is the only method in the top two on both metrics for both tasks---leading every baseline overall.

\paragraph{Qualitative comparison.}
\cref{fig:qual} compares \method with 4RC, OmniX, and UniQuery4R on in-the-wild videos from DAVIS~\cite{davis} and Pexels.
In \emph{parkour}, a person jumps across a terrace under a moving camera.
\method and UniQuery4R trace the full arc of the jumps, whereas 4RC and OmniX bunch up near the end.
In \emph{soapbox}, \method's trajectories follow the cart all the way to its final position down the street, while 4RC and OmniX stop short.
In the BMX clip, the rider drops into a bowl toward the camera, where \method and UniQuery4R preserve a coherent trajectory bundle while 4RC and OmniX bend into a hook.
In the forest clip, UniQuery4R's reconstructed road breaks into dark fragments and one trajectory drifts off the car, whereas \method reconstructs the road cleanly and tracks the car accurately.
Additional qualitative comparisons appear in \cref{app:qualitative}.

\subsection{What Makes It Work}
\label{sec:exp_analysis}

\paragraph{Joint inference.}
Because motion blocks attend only within each source--target pair, \method can run on any subset of views without retraining.
Running the same weights pairwise degrades both tracking and reconstruction substantially, while adding six intermediate frames recovers roughly half the gap (\cref{tab:inference}).
At equal two-view context, \method still exceeds Flow4R by 6.9 tracking APD from backbone, training, and architectural improvements, and joint multi-view context adds the remaining 3.7 APD.
The pairwise relative formulation thus benefits directly from multi-view context.

\paragraph{Learned rigidity.}
Varying the alignment weights in \cref{eq:solve} on the same predictions isolates the effect of learned rigidity (\cref{tab:weights}).
Learned $\mW$ is best in every column and outperforms even an oracle static mask derived from ground-truth depth and poses.
A binary static mask retains all stationary pixels regardless of depth uncertainty or boundary artifacts, whereas learned $\mW$ down-weights geometrically unreliable points to stabilize the alignment.
The gain over uniform weights is largest in reconstruction (+5.8 APD).

\paragraph{Shared head.}
One DPT head decodes both $(\mP,\mC)$ and $(\mP_{vt},\mW)$ and matches two separate heads in average APD and EPE (\cref{tab:ablation-head}).
Initializing the head from 4RC rather than from scratch degrades both tasks, so we initialize the shared head randomly but keep the pretrained backbone.

\paragraph{Changing the reference.}
Because camera motion is solved by closed-form alignment rather than regressed by a pose head, restricting $\mW$ to a moving object's mask re-expresses reconstruction and tracking relative to that object without re-running the network (\cref{fig:reference}).

\begin{table}[t]
\centering
\caption{\textbf{Joint versus pairwise inference on WorldTrack.}
Evaluating the same \method weights with increasing view context (averages over \cref{tab:worldtrack}, with Flow4R for reference) yields monotonic gains, confirming that pairwise prediction benefits directly from joint multi-view encoding.
Shading marks the top three per column.}
\label{tab:inference}
\footnotesize
\setlength{\tabcolsep}{3pt}
\begin{tabular}{@{}l@{\hspace{6pt}}cc@{\hspace{6pt}}cc@{}}
\toprule
 & \multicolumn{2}{c}{Tracking} & \multicolumn{2}{c}{Reconstruction} \\
\cmidrule(lr){2-3}\cmidrule(lr){4-5}
 & APD\,$\uparrow$ & EPE\,$\downarrow$ & APD\,$\uparrow$ & EPE\,$\downarrow$ \\
\midrule
Flow4R~\cite{flow4r} & 73.1 & 0.263 & 80.4 & 0.192 \\
\midrule
\method, pairs & \cellcolor{rank3}80.0 & \cellcolor{rank3}0.202 & \cellcolor{rank3}89.0 & \cellcolor{rank3}0.144 \\
\method, pair + 6 frames & \cellcolor{rank2}81.7 & \cellcolor{rank2}0.183 & \cellcolor{rank2}90.3 & \cellcolor{rank2}0.134 \\
\method, joint & \cellcolor{rank1}83.7 & \cellcolor{rank1}0.165 & \cellcolor{rank1}91.9 & \cellcolor{rank1}0.126 \\
\bottomrule
\end{tabular}

\vspace{8pt}
\caption{\textbf{Alignment weights on fixed predictions.}
Varying only the weights in \cref{eq:solve} shows that learned rigidity $\mW$ outperforms uniform, confidence, and even oracle static-mask weights by down-weighting unreliable pixels.
Oracle static masks require ground-truth depth (unavailable on ADT/PS), and shading marks the top three per column.}
\label{tab:weights}
\footnotesize
\setlength{\tabcolsep}{3pt}
\begin{tabular}{@{}l@{\hspace{6pt}}cc@{\hspace{6pt}}cc@{}}
\toprule
 & \multicolumn{2}{c}{Tracking} & \multicolumn{2}{c}{Reconstruction} \\
\cmidrule(lr){2-3}\cmidrule(lr){4-5}
Weights & APD\,$\uparrow$ & EPE\,$\downarrow$ & APD\,$\uparrow$ & EPE\,$\downarrow$ \\
\midrule
Learned $\mW$ & \cellcolor{rank1}83.7 & \cellcolor{rank1}0.165 & \cellcolor{rank1}91.9 & \cellcolor{rank1}0.126 \\
Uniform & \cellcolor{rank3}83.0 & \cellcolor{rank3}0.171 & 86.1 & 0.163 \\
Confidence $\mC$ & \cellcolor{rank2}83.3 & \cellcolor{rank2}0.169 & \cellcolor{rank3}90.1 & \cellcolor{rank3}0.138 \\
Oracle static mask & -- & -- & \cellcolor{rank2}91.2 & \cellcolor{rank2}0.131 \\
\bottomrule
\end{tabular}

\vspace{8pt}
\caption{\textbf{Ablation of the DPT head.}
\checkmark\ marks one head shared by $(\mP,\mC)$ and $(\mP_{vt},\mW)$, which matches two separate heads.
Training the head from scratch outperforms initializing it from 4RC.
All variants are trained for 25 epochs under the same protocol.}
\label{tab:ablation-head}
\footnotesize
\setlength{\tabcolsep}{2.5pt}
\begin{tabular}{@{}c@{\hspace{4pt}}l@{\hspace{4pt}}cc@{\hspace{4pt}}cc@{\hspace{4pt}}cc@{}}
\toprule
\multirow{2}{*}{Shared} & \multirow{2}{*}{Init.} & \multicolumn{2}{c}{Tracking} & \multicolumn{2}{c}{Recon.} & \multicolumn{2}{c}{\textbf{Average}} \\
\cmidrule(lr){3-4}\cmidrule(lr){5-6}\cmidrule(lr){7-8}
 & & APD\,$\uparrow$ & EPE\,$\downarrow$ & APD\,$\uparrow$ & EPE\,$\downarrow$ & APD\,$\uparrow$ & EPE\,$\downarrow$ \\
\midrule
 & 4RC & 82.9 & 0.174 & 88.4 & 0.159 & 85.6 & 0.166 \\
 & Scratch & \cellcolor{rank1}83.6 & \cellcolor{rank3}0.166 & \cellcolor{rank3}90.6 & \cellcolor{rank3}0.146 & \cellcolor{rank1}87.1 & \cellcolor{rank3}0.156 \\
\checkmark & Scratch & \cellcolor{rank3}83.5 & \cellcolor{rank1}0.164 & \cellcolor{rank1}90.7 & \cellcolor{rank1}0.145 & \cellcolor{rank1}87.1 & \cellcolor{rank1}0.155 \\
\bottomrule
\end{tabular}
\end{table}

\section{Conclusion}
\label{sec:conclusion}

We presented \method, showing that feedforward 4D reconstruction and tracking can be unified by seeing motion as displaced geometry.
Representing both local geometry and temporal motion as point maps with per-pixel weights lets one shared dense head decode both, predicting geometry once per view with multi-view context and motion for any requested pair.
Rigidity-weighted alignment recovers camera poses in closed form without a pose regression head and allows re-expressing the scene relative to a moving object without re-running the network.
On WorldTrack, \method achieves the highest tracking APD and ties for the highest reconstruction APD, ranking in the top two on both metrics for both tasks.
Without a pose head, it also achieves the lowest camera pose errors on ScanNet (\cref{app:video}).

\paragraph{Limitations and Outlook.}
\method predicts pairwise motion from shared multi-view features rather than propagating states along trajectories, making prolonged occlusions challenging.
Temporal priors could help bridge such intervals.
Predictions are also up to a sequence-level scale, which metric supervision or depth priors could resolve.
Small objects under heavy occlusion, such as a car in tire smoke, still yield noisy trajectories (\cref{app:qualitative}).
Finally, cross-view attention grows with the clip length, so very long videos would benefit from streaming or memory mechanisms.

\section*{Acknowledgments}
This work was supported by the ERC Advanced Grant ``SIMULACRON'' (agreement \#884679), the GNI Project ``AI4Twinning'', the DFG project CR 250/26-1 ``4DYoutube'', the Leibniz Supercomputing Centre (LRZ), and the UKRI AIRR programme.

{
    \small
    \bibliographystyle{ieeenat_fullname}
    \bibliography{main}
}

\appendix
\setlength{\textfloatsep}{10pt plus 2pt minus 2pt}
\setlength{\floatsep}{9pt plus 2pt minus 2pt}
\setlength{\dbltextfloatsep}{10pt plus 2pt minus 2pt}
\twocolumn[{%
\begin{minipage}[t]{0.6\textwidth}
\centering
\captionsetup{type=table,position=top}
\caption{\textbf{Summary of notation.}
Predictions and derived quantities are written for one view toward one target view.}
\label{tab:notation}
\footnotesize
\setlength{\tabcolsep}{2.5pt}
\begin{tabularx}{\linewidth}{@{}l>{\raggedright\arraybackslash}X@{}}
\toprule
Symbol & Meaning \\
\midrule
\multicolumn{2}{@{}l}{\emph{Input}} \\
\hspace{0.25em}$\{\mI^n\}_{n=1}^{N}$ & $N$ input images of a dynamic scene, $\mI^n \in \R^{H \times W \times 3}$ \\
\midrule
\multicolumn{2}{@{}l}{\emph{Predictions}} \\
\hspace{0.25em}$\mP$ & Local point map, placing each pixel at the view's own viewpoint and time \\
\hspace{0.25em}$\mC$ & Confidence map, $\mC \in (1, \infty)^{H \times W}$, weights geometry and motion regression \\
\hspace{0.25em}$\mP_{vt}$ & Displaced point map, placing the same pixels at the target's viewpoint and time \\
\hspace{0.25em}$\mW$ & Rigidity map, $\mW \in (0,1)^{H \times W}$ with $\sum_i \mW^{(i)} = 1$ \\
\midrule
\multicolumn{2}{@{}l}{\emph{Derived}} \\
\hspace{0.25em}$\hat{\mT}$ & Relative camera pose in $SE(3)$, solved from $\mP$, $\mP_{vt}$, and $\mW$ (\cref{eq:solve}) \\
\hspace{0.25em}$\mP_v$ & Point map at the target's viewpoint and the view's own time, $\mP_v = \hat{\mT}\mP$ \\
\hspace{0.25em}$\mP_t$ & Point map at the view's own viewpoint and the target's time, $\mP_t = \hat{\mT}^{-1}\mP_{vt}$ \\
\hspace{0.25em}$\mF$ & Scene flow, $\mF = \mP_{vt} - \mP$ \\
\hspace{0.25em}$\mF_v$, $\mF_t$ & Camera-induced displacement and scene-motion residual of $\mF$ (\cref{eq:decomp}) \\
\midrule
\multicolumn{2}{@{}l}{\emph{Training}} \\
\hspace{0.25em}$\bar{\mP}$, $\bar{\mP}_{vt}$, $\bar{\mT}$ & Ground-truth local point map, displaced point map, and relative pose \\
\hspace{0.25em}$\bar{\mP}_v$ & Ground-truth point map carried to the target viewpoint, $\bar{\mP}_v = \bar{\mT}\bar{\mP}$ \\
\hspace{0.25em}$\mM_{\mP}$, $\mM_{\mP_{vt}}$ & Masks of pixels with valid ground-truth depth and displacement \\
\bottomrule
\end{tabularx}
\end{minipage}%
\hfill
\begin{minipage}[t]{0.374\textwidth}
\centering
\captionsetup{type=table,position=top}
\caption{\textbf{Training data.}
Supervision provided by each dataset and samples drawn per epoch.}
\label{tab:training_data}
\footnotesize
\setlength{\tabcolsep}{1.0pt}
\renewcommand{\arraystretch}{1.055}  %
\begin{tabular}{@{}lccccr@{}}
\toprule
Dataset & Cam. & Depth & Flow & Tracks & Samples \\
\midrule
\multicolumn{6}{@{}l}{\emph{Static, real}} \\
\hspace{0.4em}WildRGB-D~\cite{wildrgbd} & \checkmark & \checkmark &  &  & 5{,}000 \\
\hspace{0.4em}CO3Dv2~\cite{co3d} & \checkmark & \checkmark &  &  & 5{,}000 \\
\hspace{0.4em}ARKitScenes~\cite{arkitscenes} & \checkmark & \checkmark &  &  & 5{,}000 \\
\hspace{0.4em}ScanNet~\cite{scannet} & \checkmark & \checkmark &  &  & 5{,}000 \\
\hspace{0.4em}Map-free~\cite{mapfree} & \checkmark & \checkmark &  &  & 5{,}000 \\
\hspace{0.4em}DL3DV~\cite{dl3dv} & \checkmark & \checkmark &  &  & 5{,}000 \\
\midrule
\multicolumn{6}{@{}l}{\emph{Static, synthetic}} \\
\hspace{0.4em}TartanAir~\cite{tartanair} & \checkmark & \checkmark &  &  & 5{,}000 \\
\hspace{0.4em}UnrealStereo4K~\cite{unreal4k} & \checkmark & \checkmark &  &  & 500 \\
\hspace{0.4em}Hypersim~\cite{hypersim} & \checkmark & \checkmark &  &  & 2{,}000 \\
\midrule
\multicolumn{6}{@{}l}{\emph{Dynamic, synthetic}} \\
\hspace{0.4em}Virtual KITTI 2~\cite{vkitti2} & \checkmark & \checkmark & \checkmark &  & 2{,}000 \\
\hspace{0.4em}Point Odyssey~\cite{pointodyssey} & \checkmark & \checkmark &  & \checkmark & 5{,}000 \\
\hspace{0.4em}Dynamic Replica~\cite{dynamicstereo} & \checkmark & \checkmark &  & \checkmark & 5{,}000 \\
\hspace{0.4em}Kubric~\cite{kubric} & \checkmark & \checkmark &  & \checkmark & 5{,}000 \\
\hspace{0.4em}Syn4D~\cite{syn4d} & \checkmark & \checkmark &  & \checkmark & 6{,}000 \\
\bottomrule
\end{tabular}
\end{minipage}

\vspace{2em}
}]

\section{Notation}
\label{app:notation}

\cref{tab:notation} summarizes the notation of \cref{sec:methodology} and \cref{app:training_objectives}.

\section{Implementation Details}
\label{app:implementation}

\subsection{Training Data}
\label{app:training_data}
\cref{tab:training_data} lists the training datasets, grouped by content and domain, with the supervision each provides and the number of samples drawn from it per epoch.
Depth and camera parameters supervise the local point map $\mP$ and, through the ground-truth relative pose, the displaced point map $\mP_{vt}$ (\cref{eq:loss_p_v}) and the rigidity $\mW$ (\cref{eq:loss_w}).
Scene flow and 3D point tracks supervise $\mP_{vt}$ directly (\cref{eq:loss_p_vt}).

\subsection{Architecture and Training Schedule}
\label{app:architecture}

The backbone is the 40-layer ViT-g/14 encoder of 4RC~\cite{fourrc}, initialized from its public weights, while the motion blocks and the DPT head are initialized randomly.
Within-view and cross-view attention alternate from layer 14 onward, and tokens are read out after layers 20, 28, 34, and 40.
Each read-out depth feeds two motion blocks of width 1536, for 1.73\,B parameters in total, 0.57\,B of them in the motion blocks.
We train in a single stage for 100 epochs at a long side of 518 pixels, with aspect ratios from $518 \times 392$ to $518 \times 168$, random crops, and color jitter.
Each step processes 352 views, 44 per GPU on 8 H100 GPUs.
We use AdamW~\cite{adamw} with $\beta = (0.9, 0.95)$ and weight decay 0.05, a linear warmup over 2 epochs to a peak learning rate of $5 \times 10^{-5}$, cosine decay to $10^{-6}$, gradient clipping at norm 10, and bfloat16 mixed precision.

\subsection{Training Objectives}
\label{app:training_objectives}

We supervise every ordered pair the model predicts, writing the terms below for one pair and averaging each over the valid pixels of all pairs in a batch.
Following DUSt3R~\cite{dust3r}, point maps are normalized to make every term scale-invariant.
The left panel of \cref{fig:downstream-tasks} shows which prediction each term reaches.

\paragraph{Geometry.}
The local point map is supervised against the back-projected ground-truth depth $\bar{\mP}$ with the standard confidence weighting:
\begin{equation}
    \mathcal{L}_{\mP} = \frac{1}{|\mM_{\mP}|} \sum_i \mM_{\mP}^{(i)} \bigl( \mC^{(i)} \bigl\| \mP^{(i)} - \bar{\mP}^{(i)} \bigr\|_2 - \alpha \log \mC^{(i)} \bigr) ,
    \label{eq:loss_p}
\end{equation}
where $\mM_{\mP}$ masks pixels with valid depth.

\begin{table*}[t]
\centering
\caption{\textbf{Video depth and camera pose evaluation.}
\method leads all three ScanNet pose metrics and ranks in the top three on Sintel/KITTI depth and TUM RPE$_t$ without a pose head.
$\delta$ denotes $\delta<1.25$, RPE$_t$/RPE$_r$ are one-frame relative pose errors, and split sizes are in parentheses.}
\label{tab:video}
\footnotesize
\setlength{\tabcolsep}{2pt}
\begin{tabularx}{\textwidth}{@{}l*{12}{>{\centering\arraybackslash}X}@{}}  %
\toprule
 & \multicolumn{6}{c}{Video depth} & \multicolumn{6}{c}{Camera pose} \\
\cmidrule(lr){2-7}\cmidrule(lr){8-13}
 & \multicolumn{2}{c}{Sintel (23)} & \multicolumn{2}{c}{Bonn (5)} & \multicolumn{2}{c}{KITTI (13)} & \multicolumn{3}{c}{ScanNet (94)} & \multicolumn{3}{c}{TUM (8)} \\
\cmidrule(lr){2-3}\cmidrule(lr){4-5}\cmidrule(lr){6-7}\cmidrule(lr){8-10}\cmidrule(lr){11-13}
 & Abs Rel\,$\downarrow$ & $\delta$\,$\uparrow$ & Abs Rel\,$\downarrow$ & $\delta$\,$\uparrow$ & Abs Rel\,$\downarrow$ & $\delta$\,$\uparrow$ & ATE\,$\downarrow$ & RPE$_t$\,$\downarrow$ & RPE$_r$\,$\downarrow$ & ATE\,$\downarrow$ & RPE$_t$\,$\downarrow$ & RPE$_r$\,$\downarrow$ \\
\midrule
V-DPM~\cite{vdpm} & 0.203 & 0.699 & 0.047 & \cellcolor{rank3}0.977 & 0.070 & 0.938 & \cellcolor{rank3}0.0342 & 0.0141 & 0.4041 & 0.0144 & 0.0103 & \cellcolor{rank3}0.3132 \\
4RC~\cite{fourrc} & 0.268 & 0.661 & 0.048 & 0.973 & 0.052 & 0.959 & \cellcolor{rank2}0.0315 & \cellcolor{rank2}0.0131 & \cellcolor{rank3}0.3848 & \cellcolor{rank1}0.0107 & \cellcolor{rank1}0.0091 & \cellcolor{rank2}0.3129 \\
Track4World~\cite{track4world} & 0.236 & 0.659 & \cellcolor{rank1}0.039 & \cellcolor{rank1}0.980 & \cellcolor{rank1}0.032 & \cellcolor{rank1}0.985 & 0.0347 & 0.0134 & \cellcolor{rank2}0.3826 & \cellcolor{rank2}0.0121 & \cellcolor{rank1}0.0091 & \cellcolor{rank1}0.3085 \\
OmniX~\cite{omnix} & 0.348 & 0.731 & 0.046 & 0.975 & 0.075 & 0.941 & 0.0345 & \cellcolor{rank3}0.0133 & 0.4895 & \cellcolor{rank3}0.0123 & 0.0101 & 0.3208 \\
SM4RT~\cite{sm4rt} & 0.269 & 0.661 & 0.048 & 0.973 & 0.052 & 0.959 & 0.0346 & 0.0137 & 0.5132 & 0.0129 & \cellcolor{rank3}0.0096 & 0.3245 \\
Point4D~\cite{point4d} & \cellcolor{rank2}0.169 & \cellcolor{rank2}0.748 & 0.045 & 0.975 & \cellcolor{rank3}0.043 & \cellcolor{rank2}0.979 & 0.0368 & 0.0135 & 0.4879 & 0.0268 & 0.0134 & 0.3940 \\
UniQuery4R~\cite{uniquery4r} & \cellcolor{rank1}0.109 & \cellcolor{rank1}0.870 & \cellcolor{rank2}0.041 & \cellcolor{rank2}0.978 & 0.118 & 0.878 & \cellcolor{rank3}0.0342 & 0.0140 & 0.3991 & 0.0138 & 0.0097 & 0.3236 \\
\midrule
\textbf{\method} & \cellcolor{rank3}0.192 & \cellcolor{rank3}0.743 & \cellcolor{rank3}0.043 & 0.975 & \cellcolor{rank2}0.041 & \cellcolor{rank3}0.978 & \cellcolor{rank1}0.0284 & \cellcolor{rank1}0.0115 & \cellcolor{rank1}0.3431 & 0.0125 & \cellcolor{rank2}0.0093 & 0.3197 \\
\bottomrule
\end{tabularx}
\end{table*}

\paragraph{Motion.}
Ground-truth 3D displacement is scarce, so we convert every available correspondence---scene flow and 3D point tracks---into the displaced point map it implies and supervise $\mP_{vt}$ directly:
\begin{equation}
    \mathcal{L}_{\mP_{vt}} = \frac{1}{|\mM_{\mP_{vt}}|} \sum_i \mM_{\mP_{vt}}^{(i)} \bigl( \mC^{(i)} \bigl\| \mP_{vt}^{(i)} - \bar{\mP}_{vt}^{(i)} \bigr\|_2 - \alpha \log \mC^{(i)} \bigr) .
    \label{eq:loss_p_vt}
\end{equation}
Both losses measure 3D point errors for the same source pixels, and the source confidence map $\mC$ weights both.

Static data supply displacement too.
For a static scene, the target displaced point map is $\bar{\mP}_v = \bar{\mT}\bar{\mP}$, obtained by applying the ground-truth relative camera pose $\bar{\mT}$ to the local geometry.
Every static pair thus supervises motion:
\begin{equation}
    \medmuskip=2mu plus 1mu minus 2mu \thickmuskip=3mu plus 2mu  %
    \begin{aligned}
    \mathcal{L}_{\mP_v} &= \frac{1}{|\mM_{\mP}|} \sum_i \mM_{\mP}^{(i)} \bigl( w^{(i)} \mC^{(i)} \bigl\| \mP_{vt}^{(i)} - \bar{\mP}_{v}^{(i)} \bigr\|_2 - \alpha \log \mC^{(i)} \bigr) , \\
    w^{(i)} &= \begin{cases}
        \mathrm{sg}(\mW^{(i)}) \cdot HW & \text{dynamic}, \\
        1 & \text{static}.
    \end{cases}
    \end{aligned}
    \label{eq:loss_p_v}
\end{equation}
On dynamic data, this term would contradict \cref{eq:loss_p_vt} wherever the scene moves, so the predicted rigidity gates it.
The gate is rescaled by $HW$ to undo the normalization of $\mW$ and detached by $\mathrm{sg}(\cdot)$ so that it does not interfere with the rigidity term below.

\paragraph{Rigidity.}
A binary motion mask does not fully specify which points provide a reliable rigid alignment.
Static foreground objects can support camera pose estimation, while uncertain geometry can make even stationary points unreliable.
We therefore supervise $\mW$ through the alignment it produces, backpropagating through \cref{eq:solve}:
\begin{equation}
    \mathcal{L}_{\mW} = \frac{1}{|\mM_{\mP}|} \sum_i \mM_{\mP}^{(i)} \bigl\| \mP_{v}^{(i)} - \bar{\mP}_{v}^{(i)} \bigr\|_2 ,
    \label{eq:loss_w}
\end{equation}
where $\mP_v = \hat{\mT}\mP$ is the predicted counterpart of $\bar{\mP}_v$.
Gradients to $\mP$ and $\mP_{vt}$ are blocked, as both are already supervised, so this term adjusts only $\mW$.
This objective learns weights that make the recovered transformation agree with the ground-truth camera motion without rigidity annotations.

\paragraph{Rigidity entropy.}
To keep $\mW$ from collapsing onto a few pixels, we maximize its entropy $\mathcal{H}(\mW)$ over the source pixels by minimizing
\begin{equation}
    \mathcal{L}_{\mathcal{H}} = -\frac{\mathcal{H}(\mW)}{\log(HW)} , \quad
    \mathcal{H}(\mW) = -\sum_i \mW^{(i)} \log \mW^{(i)} ,
    \label{eq:loss_entropy}
\end{equation}
where dividing by $\log(HW)$, the entropy of uniform weights, makes the term independent of the image resolution.
This term is applied to every pair without masking, and the loss weights and pair sampling follow \cref{sec:method_training}.

\section{Additional Quantitative Results}
\label{app:results}

\subsection{Video Depth and Camera Pose}
\label{app:video}

We additionally evaluate video depth and camera pose under the MonST3R protocol~\cite{monst3r}.
Video depth is evaluated on Sintel~\cite{sintel}, Bonn~\cite{bonn}, and KITTI~\cite{kitti}, and camera pose on ScanNet~\cite{scannet} and TUM-Dynamics~\cite{tumrgbd}.
Depth is aligned to the ground truth by one scale-and-shift fit per sequence over the stacked clip, with maximum depth capped at 70\,m on Sintel and Bonn and uncapped on KITTI.
Pose is measured with \texttt{evo} under a $\mathrm{Sim}(3)$ alignment.
On ScanNet, all methods run the same 94 of 100 scenes, excluding six scenes with malformed ground-truth pose files.
The input long side is 518 pixels for \method, OmniX, Point4D, V-DPM, 4RC, SM4RT, and UniQuery4R, and 512 pixels for Track4World.
Track4World's correlation pyramid needs a short side of at least 256 pixels, which the $155\times512$ aspect-preserving resize of KITTI does not meet.
For its KITTI entry, we therefore downsample each frame to $155\times512$ and upsample it to a canvas the model accepts, so the input carries no more image information than at a 512-pixel long side.
\method ranks among the top three across Sintel, Bonn, and KITTI depth without domain collapse (whereas UniQuery4R leads Sintel but degrades sharply on outdoor KITTI), and achieves the lowest errors on all three ScanNet pose metrics---outperforming 4RC's pretrained pose head via closed-form alignment (\cref{tab:video}).
On TUM, \method ranks second in translational relative pose error and remains competitive in ATE and rotational error.

\enlargethispage{\baselineskip}  %
\subsection{Consistency of Joint Inference}
\label{app:consistency}

\paragraph{Bidirectional consistency.}
The forward transformation of pair $(n, a)$ and the backward transformation of pair $(a, n)$ are solved independently by \cref{eq:solve} in double precision, and nothing in training ties them together.
Even so, their composition remains tightly consistent with the identity (\cref{tab:app-cycle}).
While this consistency stems from the shared formulation and is present already with two views, joint inference lowers both cycle errors and reduces relative pose rotation and translation errors by 20\% and 35\%.

\paragraph{Scale consistency.}
Joint inference also keeps all outputs at a single consistent scale across the clip.
Re-fitting the scale per frame instead of once per clip changes the average tracking APD by less than $0.1$ for joint inference, compared with $+1.1$ for pairwise inference.

\begin{table}[t]
\centering
\caption{\textbf{Bidirectional consistency and relative pose.}
Independent forward $(n, a)$ and backward $(a, n)$ alignments are tightly cycle-consistent without cycle supervision, and joint inference cuts relative pose errors by 20--35\%.
Rotation in degrees, translation in normalized units, averaged over ADT, DR, PO, and TUM.}
\label{tab:app-cycle}
\footnotesize
\begin{tabular}{@{}lcccc@{}}
\toprule
 & \multicolumn{2}{c}{Cycle\,$\downarrow$} & \multicolumn{2}{c}{Pose\,$\downarrow$} \\
\cmidrule(lr){2-3}\cmidrule(lr){4-5}
Inference & Rot. & Trans. & Rot. & Trans. \\
\midrule
Pairs & 0.34 & 0.0168 & 0.41 & 0.0238 \\
Joint & \textbf{0.33} & \textbf{0.0144} & \textbf{0.33} & \textbf{0.0154} \\
\bottomrule
\end{tabular}
\end{table}

\subsection{Runtime Efficiency}
\label{app:efficiency}

\begin{figure}[t]
\begin{minipage}{\linewidth}  %
\centering
\captionsetup{type=table,position=top}
\caption{\textbf{Runtime efficiency.}
\method ranks second in throughput and third in peak VRAM for producing both 64-view world-space point maps and tracks on one H100 GPU (518-pixel long side).}
\label{tab:efficiency}
\footnotesize
\begin{tabular}{@{}l@{\hspace{6pt}}c@{\hspace{6pt}}c@{\hspace{6pt}}c@{}}
\toprule
Method & \# par.\ (B) & views/s\,$\uparrow$ & VRAM (GB)\,$\downarrow$ \\
\midrule
V-DPM & 1.66 & 0.6 & 32.4 \\
4RC & 1.52 & \cellcolor{rank3}23.1 & \cellcolor{rank2}17.5 \\
Track4World & 1.72 & 15.1 & \cellcolor{rank1}15.9 \\
OmniX & 1.33 & 6.6 & 32.8 \\
SM4RT & 1.38 & \cellcolor{rank1}30.6 & 21.5 \\
Point4D & 1.45 & 1.9 & 21.3 \\
Any4D & 0.61 & 14.4 & 30.7 \\
UniQuery4R & 1.65 & 12.7 & 29.4 \\
\midrule
\textbf{\method} & 1.73 & \cellcolor{rank2}28.3 & \cellcolor{rank3}20.6 \\
\bottomrule
\end{tabular}
\end{minipage}

\par\vspace{\floatsep}
\centering
\includegraphics[width=\linewidth]{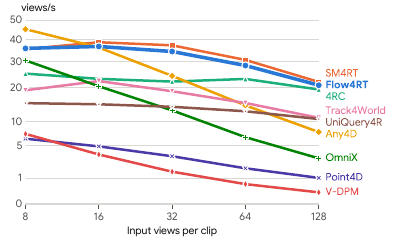}
\caption{\textbf{Throughput against clip length.} Views per second against the number of input views on a log-scaled view axis and a square-root rate axis, under the protocol of \cref{tab:efficiency}.
Cost linear in the clip length gives a horizontal line, whereas downward slopes indicate super-linear cost.}
\label{fig:efficiency-scaling}
\end{figure}

\paragraph{Cost per clip.}
\cref{tab:efficiency} reports the cost of producing both world-space point maps and world-space tracks for a 64-view clip on one H100 GPU.
We time GPU computation from each method's preprocessed inputs to its outputs, including separate passes when required for reconstruction and tracking.
\method ranks second in throughput behind SM4RT and ahead of 4RC, while all remaining baselines are substantially slower.
In peak memory, \method ranks third behind Track4World and 4RC, using markedly less VRAM than UniQuery4R, Any4D, V-DPM, and OmniX.

\paragraph{Scaling with clip length.}
\cref{fig:efficiency-scaling} reports throughput for clips of 8--128 views.
All methods encode the full clip jointly, while Track4World performs its subsequent flow refinement in overlapping 16-frame windows matching its accuracy evaluation.
In \method, cross-view attention grows with clip length, whereas each motion block attends only within one source--target pair and thus scales linearly with the number of requested pairs ($2N$ for the bidirectional anchor configuration, each followed by a closed-form alignment).
By contrast, V-DPM reruns its time-conditioned decoder over the full clip for each query time to produce tracks, taking 110\,s for 64 views versus 2.3\,s for \method.
Reusing geometry features across requested pairs avoids this repeated full-clip decoding.

\section{Additional Qualitative Results}
\label{app:qualitative}

\begin{figure*}[p]
\centering
\ifdefined\qualbox\else\newsavebox\qualbox\fi
\setlength{\tabcolsep}{1pt}
\begin{tabular}{@{}cc@{}}
\footnotesize Input & \begin{tabular}{@{}cccc@{}}\makebox[0.16\linewidth]{\footnotesize 4RC} & \makebox[0.16\linewidth]{\footnotesize OmniX} & \makebox[0.16\linewidth]{\footnotesize UniQuery4R} & \makebox[0.16\linewidth]{\footnotesize \textbf{\method}}\end{tabular} \\
\global\setbox\qualbox\hbox{\begin{tabular}[b]{@{}cccc@{}}\includegraphics[width=0.16\linewidth]{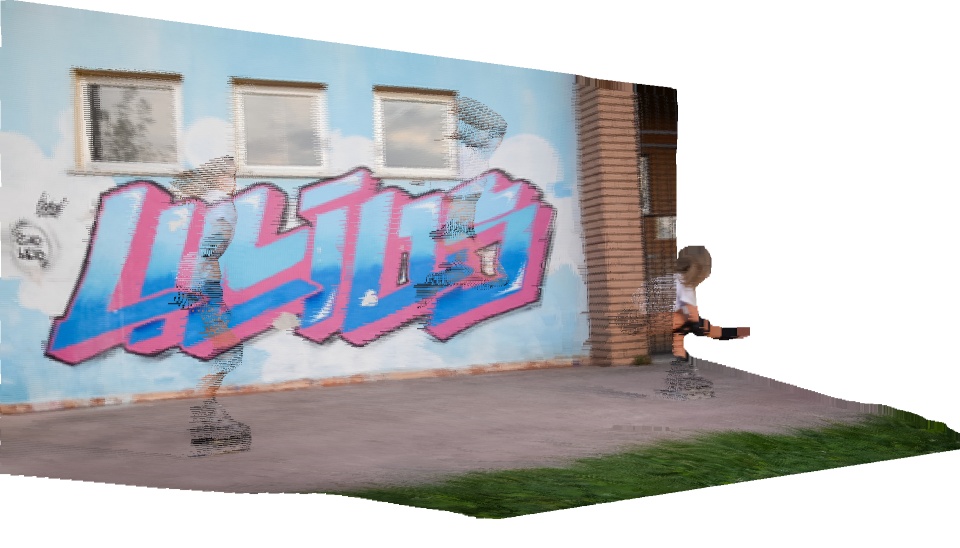} & \includegraphics[width=0.16\linewidth]{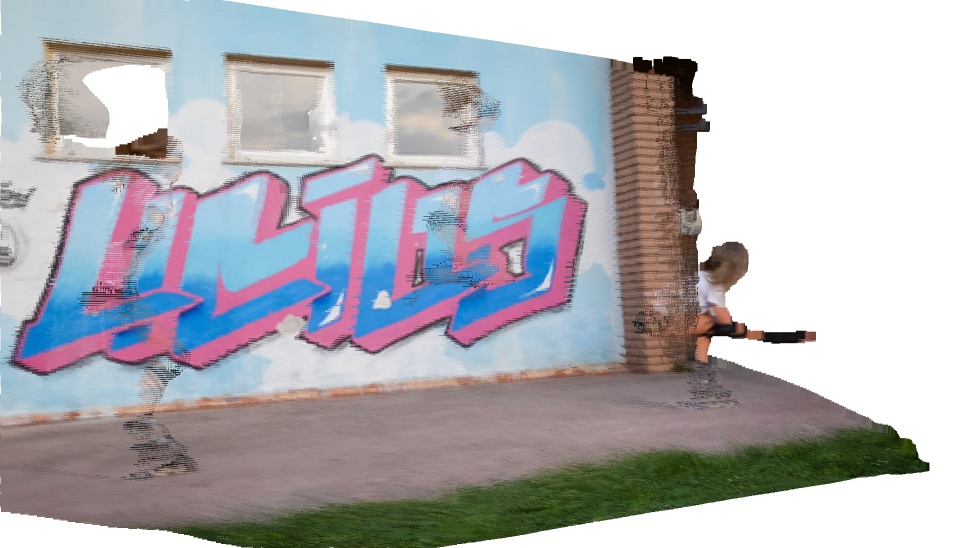} & \includegraphics[width=0.16\linewidth]{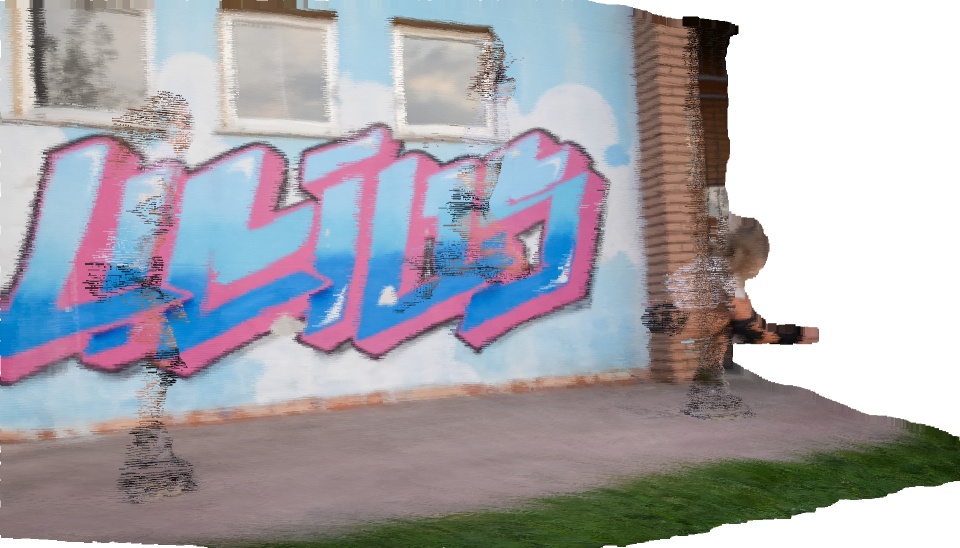} & \includegraphics[width=0.16\linewidth]{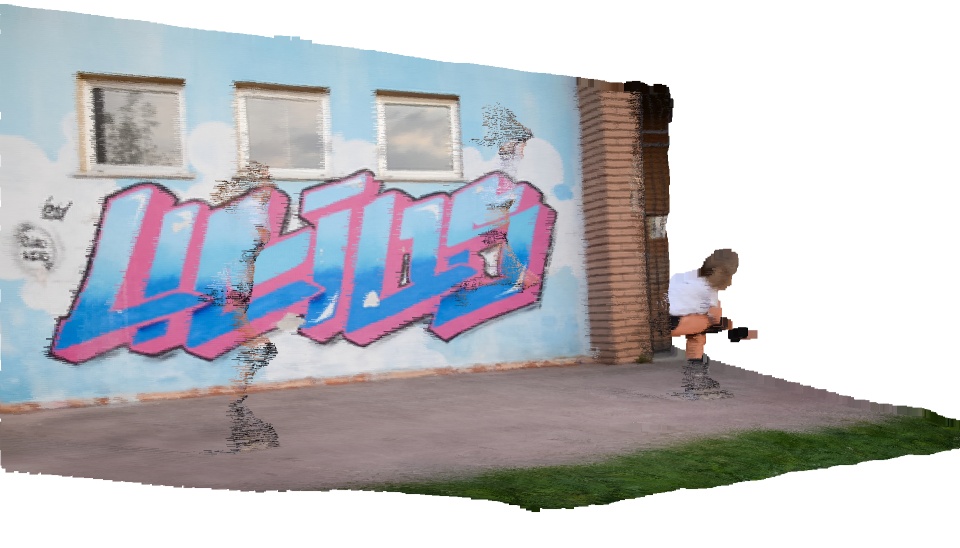} \\[-2pt] \includegraphics[width=0.16\linewidth]{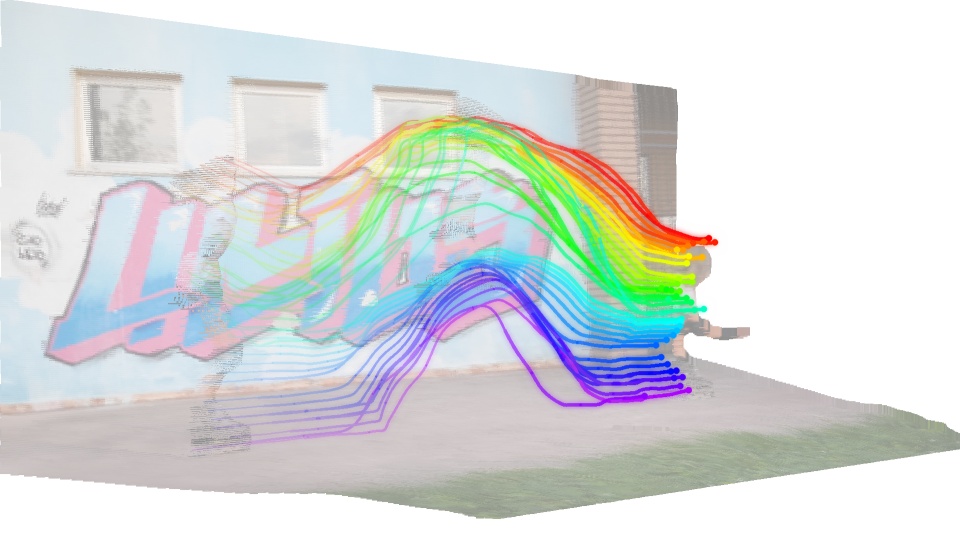} & \includegraphics[width=0.16\linewidth]{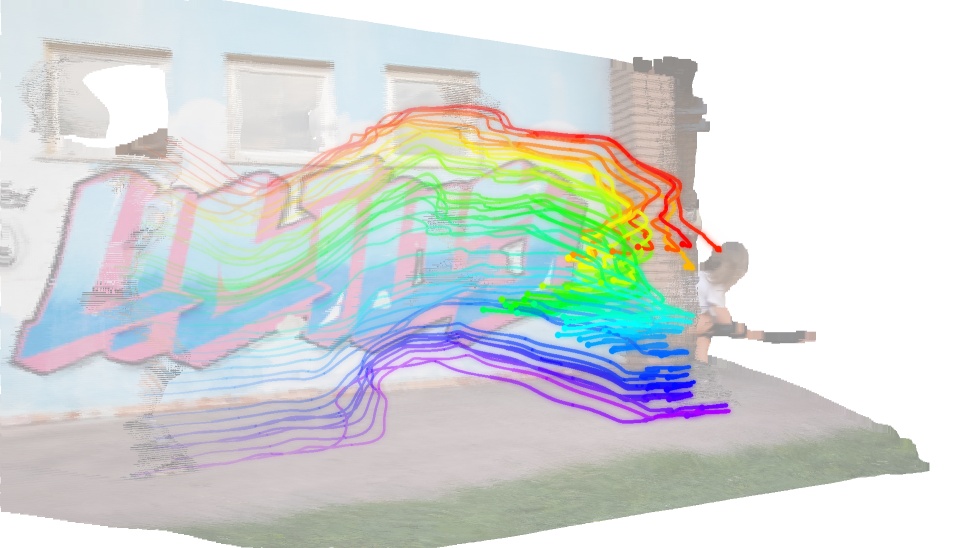} & \includegraphics[width=0.16\linewidth]{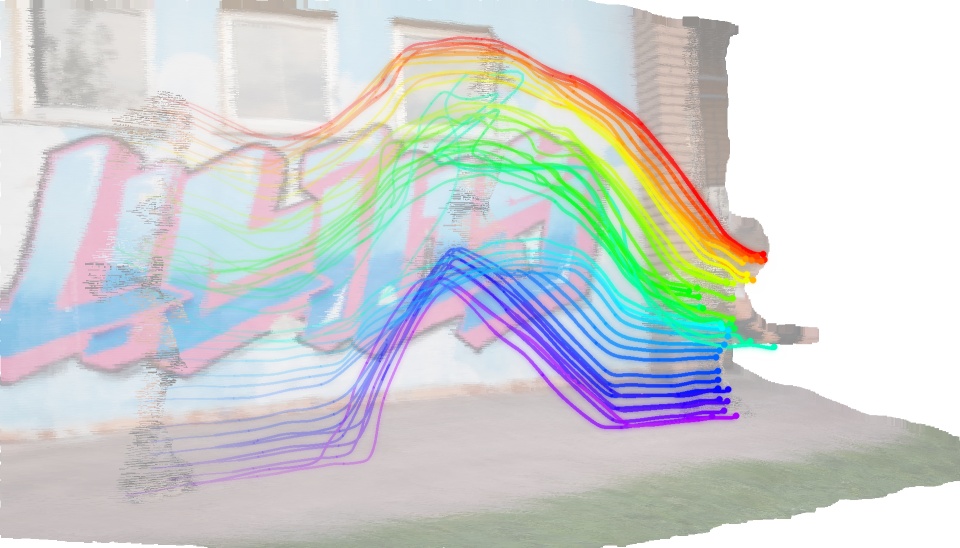} & \includegraphics[width=0.16\linewidth]{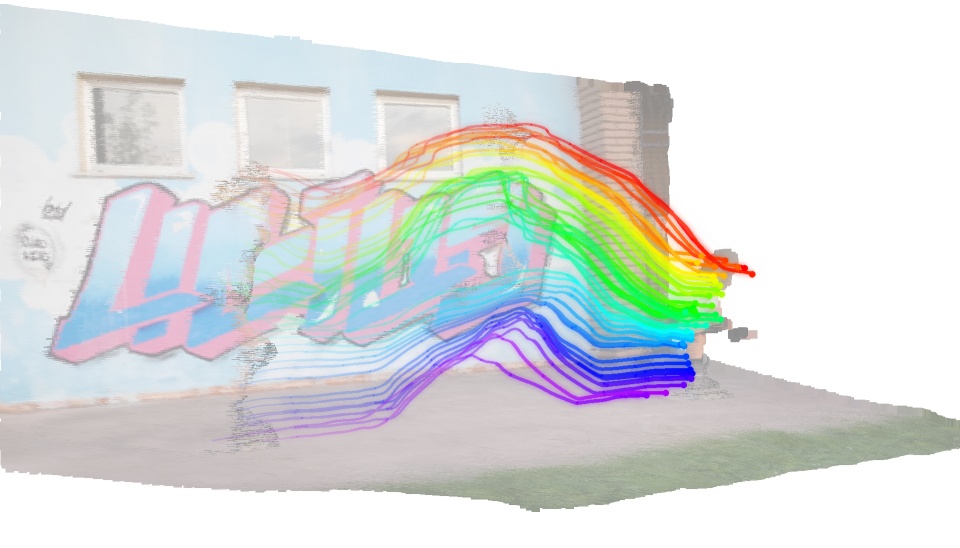} \end{tabular}}%
\raisebox{-\dp\qualbox}{\includegraphics[height=\dimexpr\ht\qualbox+\dp\qualbox\relax]{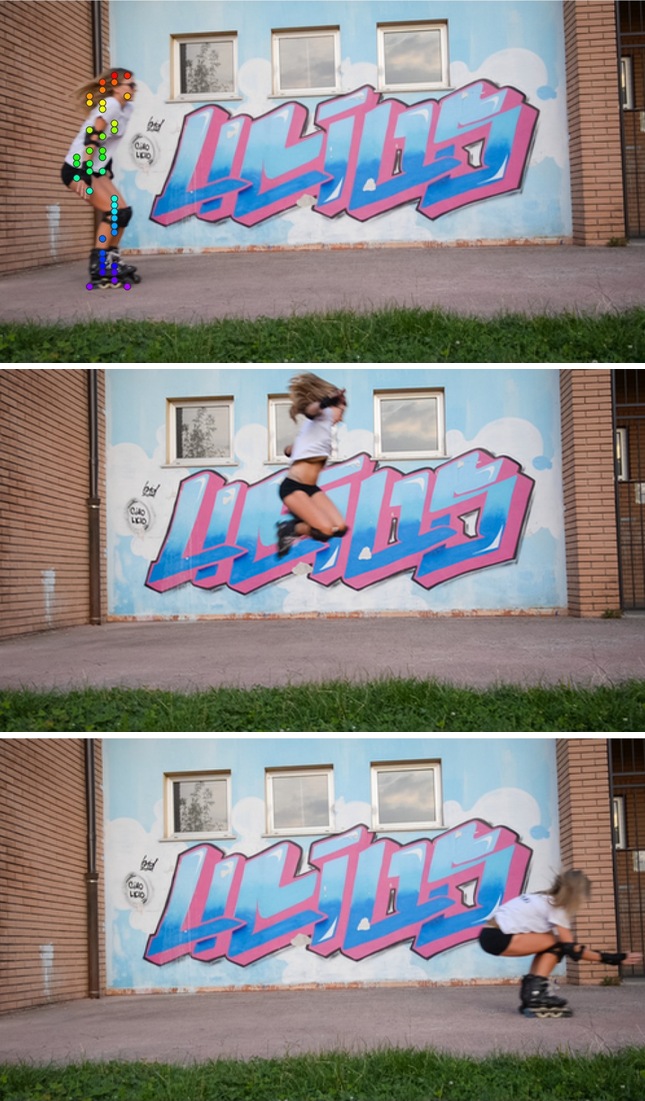}} & \usebox\qualbox \\[3pt]
\global\setbox\qualbox\hbox{\begin{tabular}[b]{@{}cccc@{}}\includegraphics[width=0.16\linewidth]{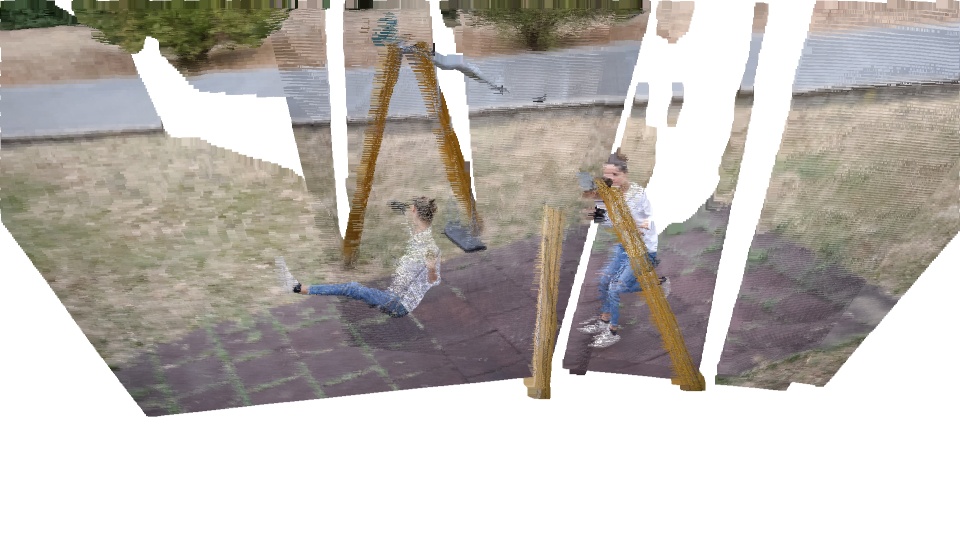} & \includegraphics[width=0.16\linewidth]{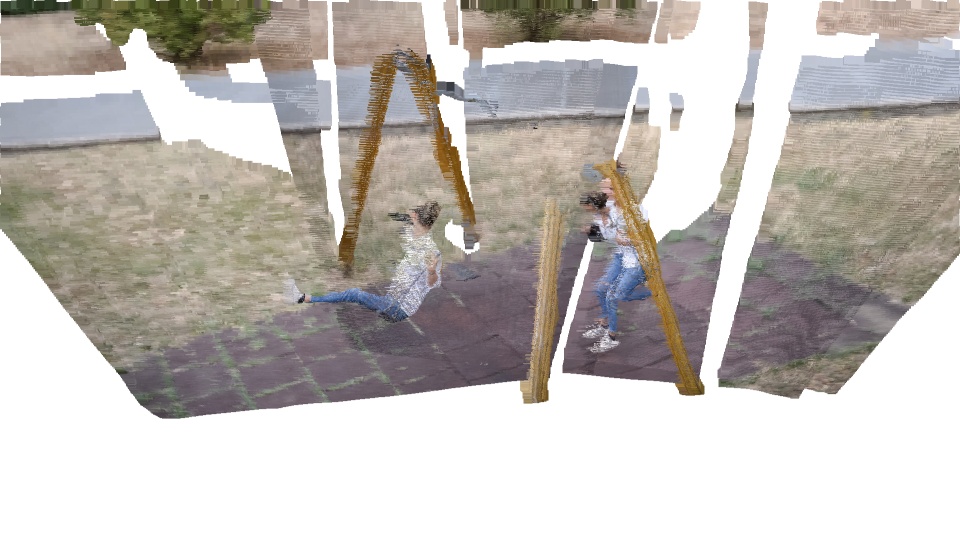} & \includegraphics[width=0.16\linewidth]{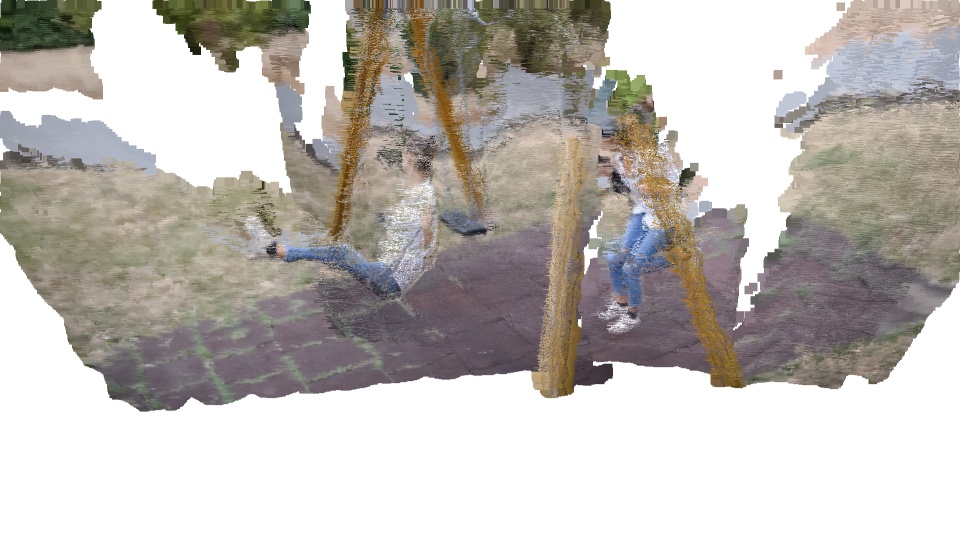} & \includegraphics[width=0.16\linewidth]{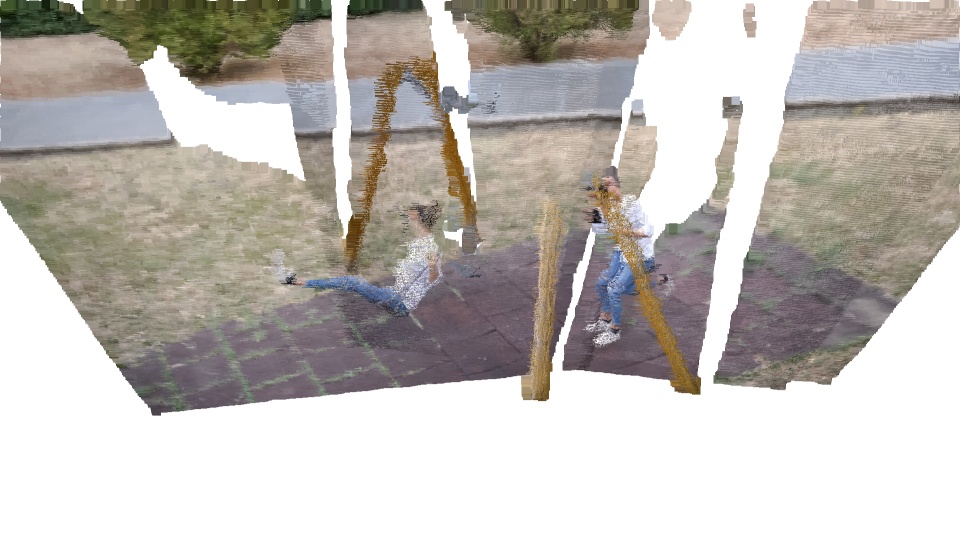} \\[-2pt] \includegraphics[width=0.16\linewidth]{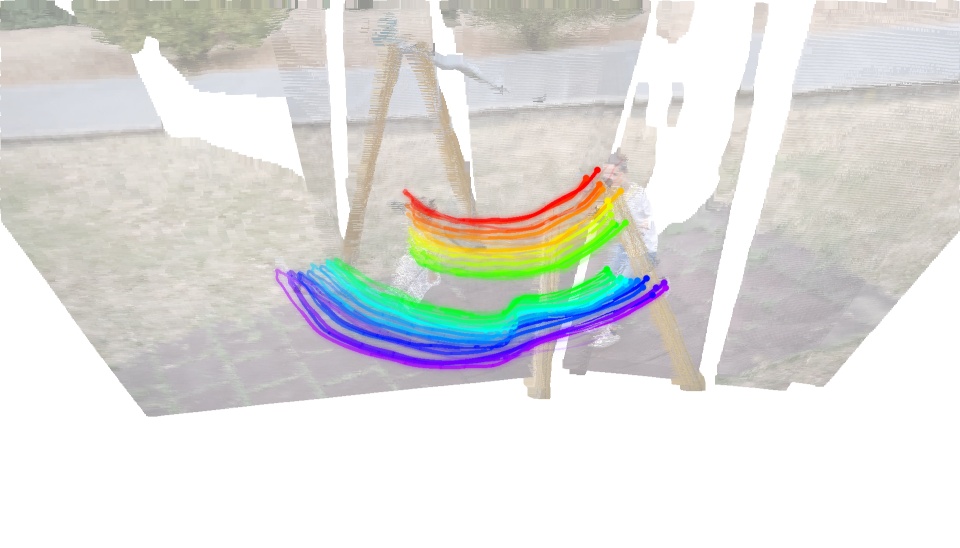} & \includegraphics[width=0.16\linewidth]{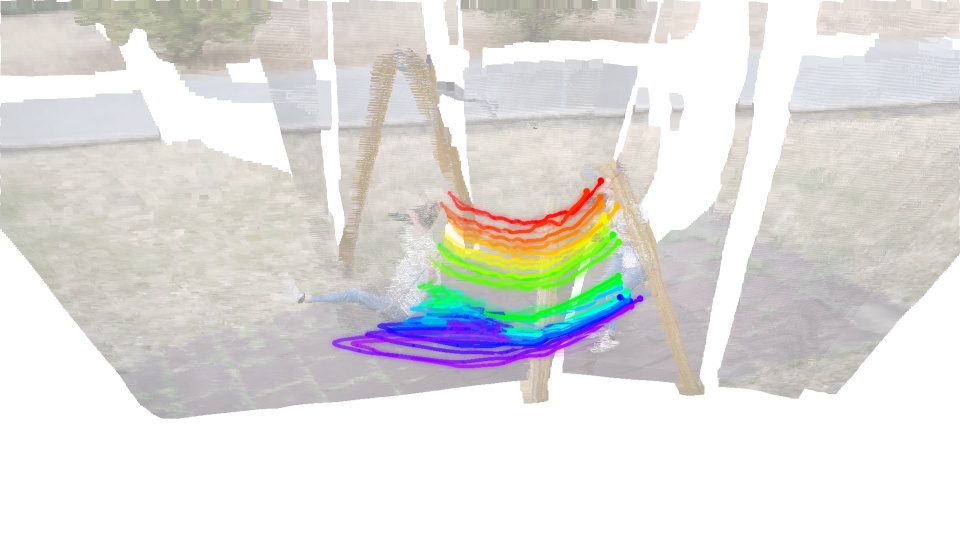} & \includegraphics[width=0.16\linewidth]{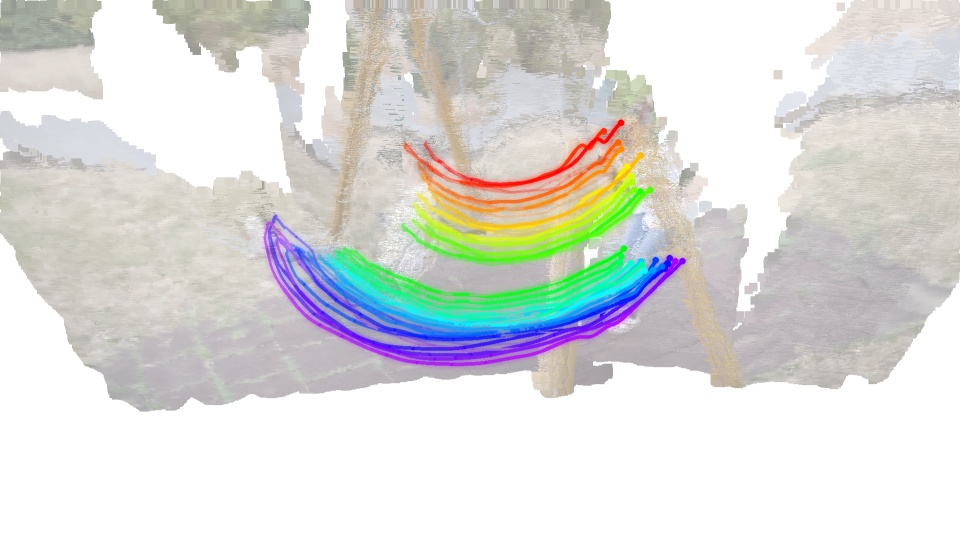} & \includegraphics[width=0.16\linewidth]{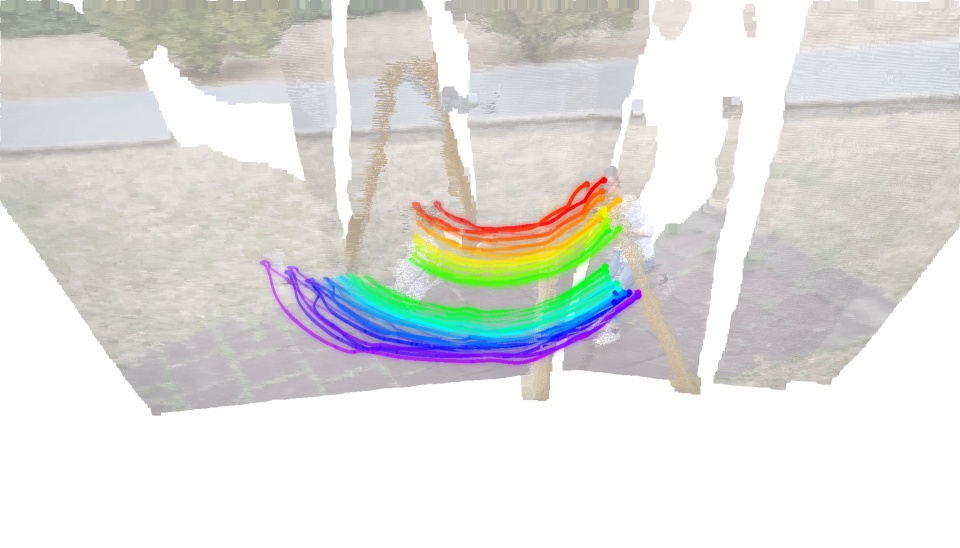} \end{tabular}}%
\raisebox{-\dp\qualbox}{\includegraphics[height=\dimexpr\ht\qualbox+\dp\qualbox\relax]{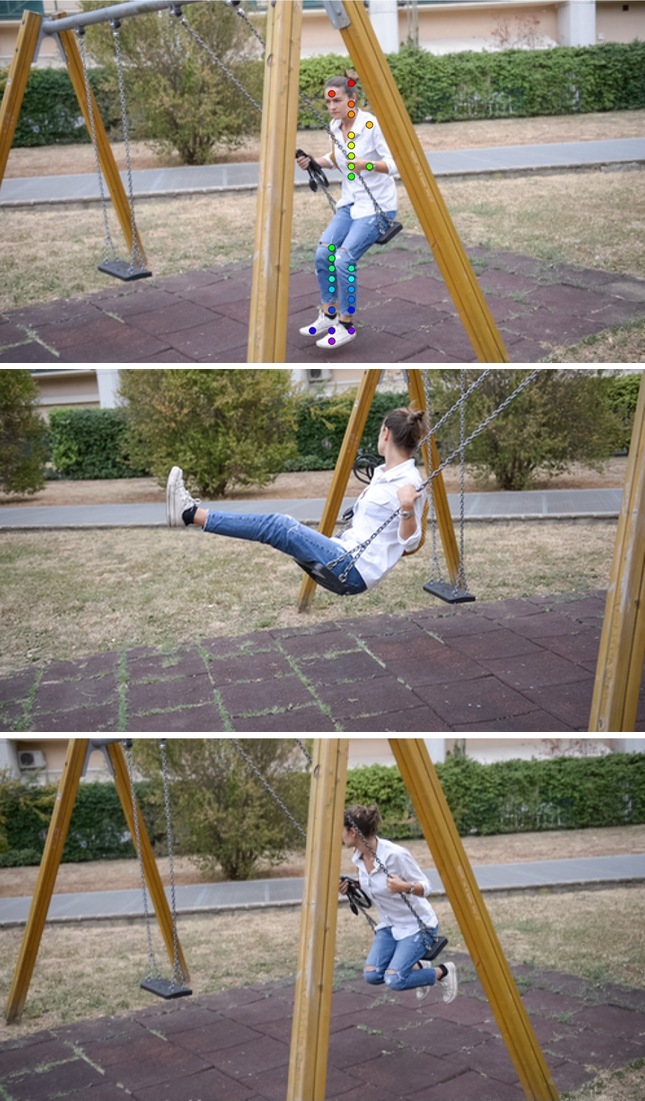}} & \usebox\qualbox \\[3pt]
\global\setbox\qualbox\hbox{\begin{tabular}[b]{@{}cccc@{}}\includegraphics[width=0.16\linewidth]{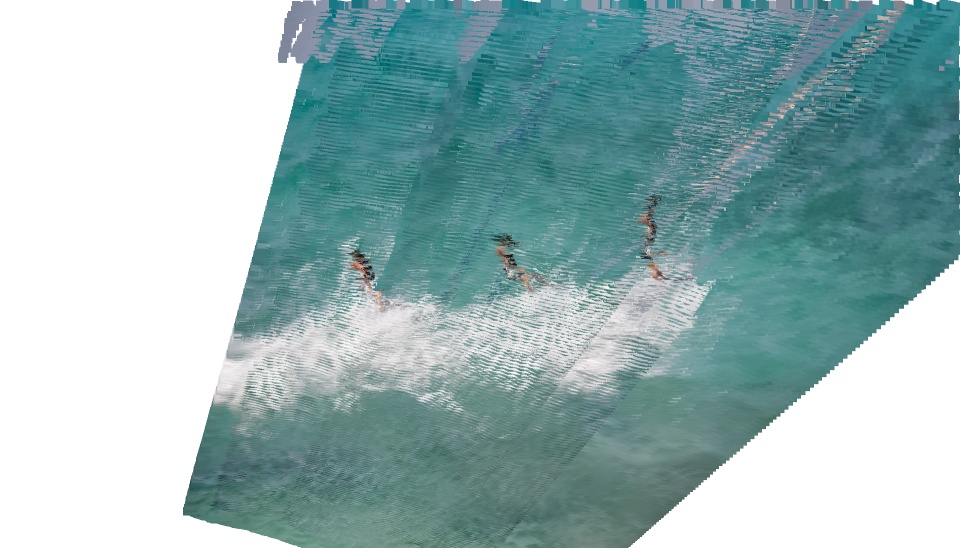} & \includegraphics[width=0.16\linewidth]{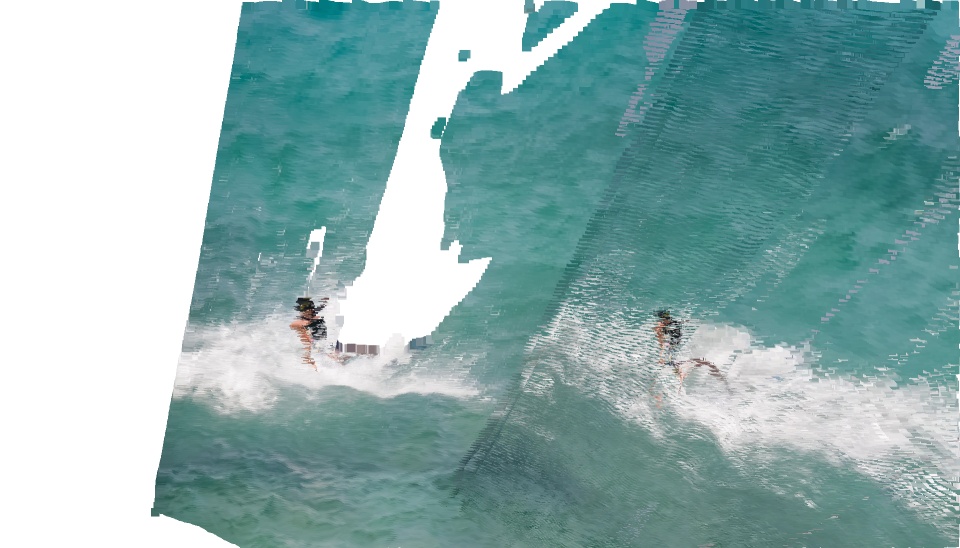} & \includegraphics[width=0.16\linewidth]{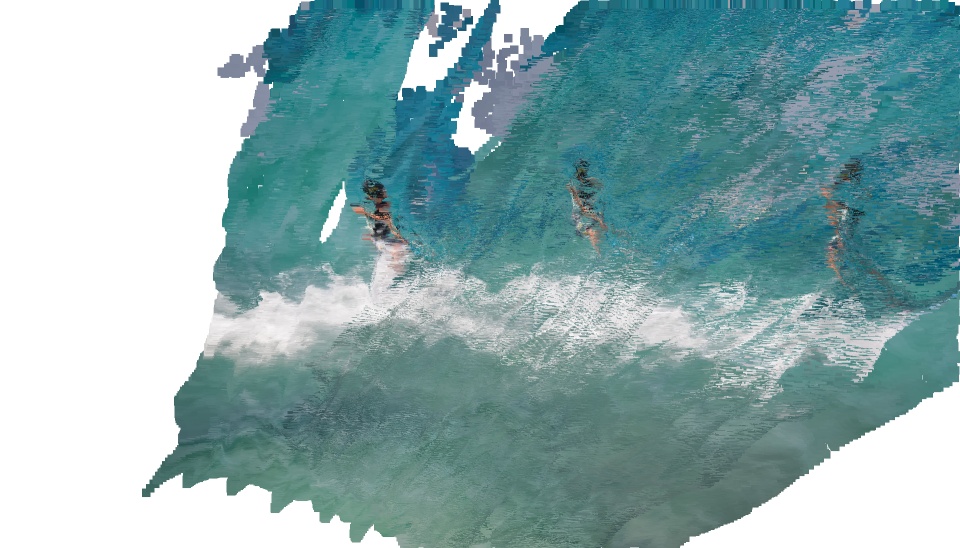} & \includegraphics[width=0.16\linewidth]{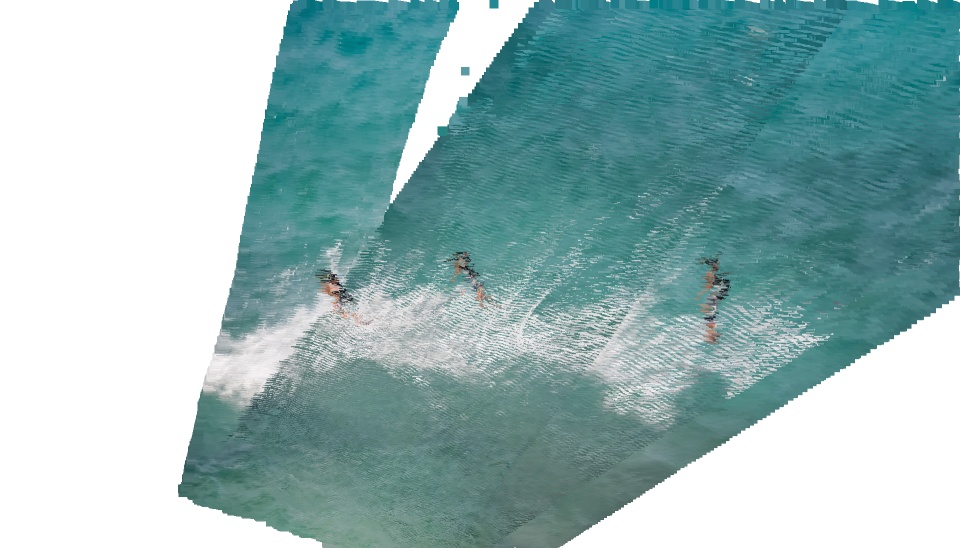} \\[-2pt] \includegraphics[width=0.16\linewidth]{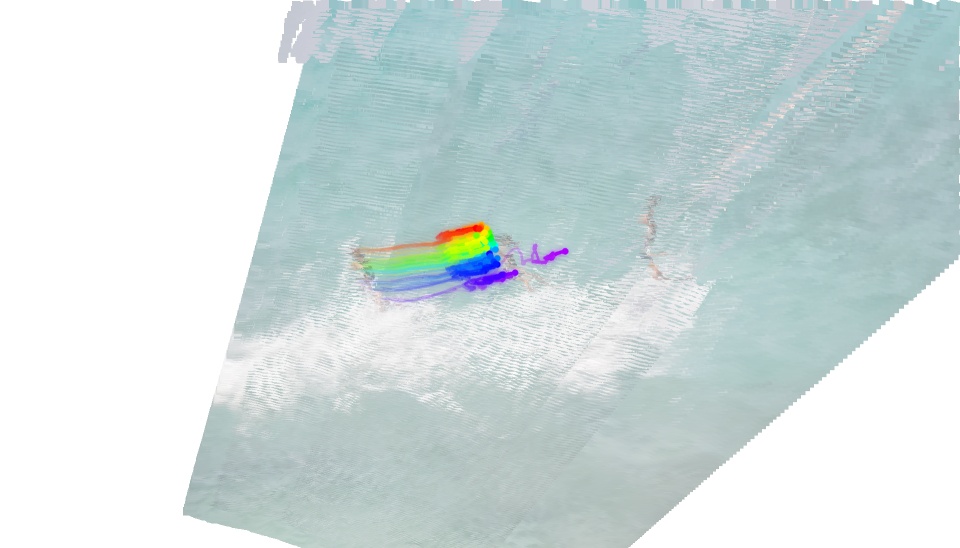} & \includegraphics[width=0.16\linewidth]{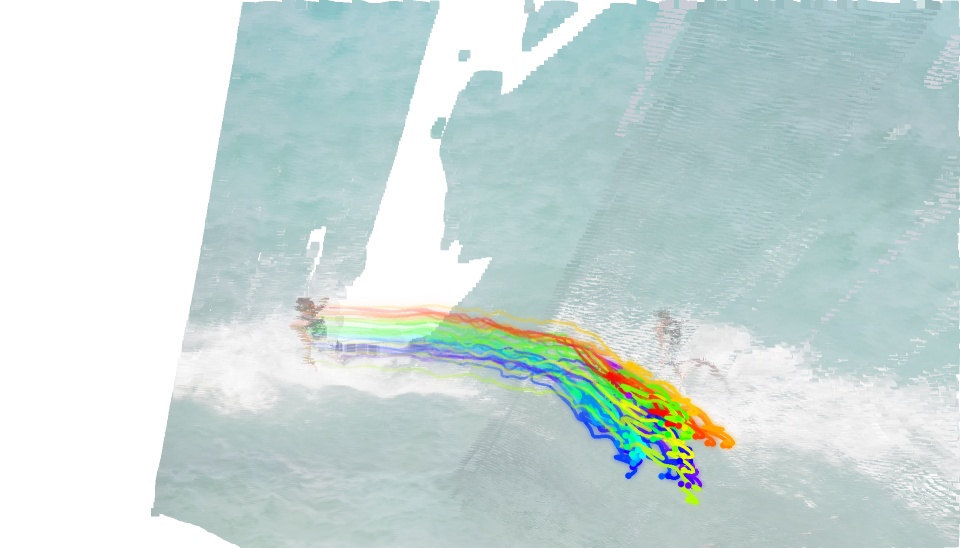} & \includegraphics[width=0.16\linewidth]{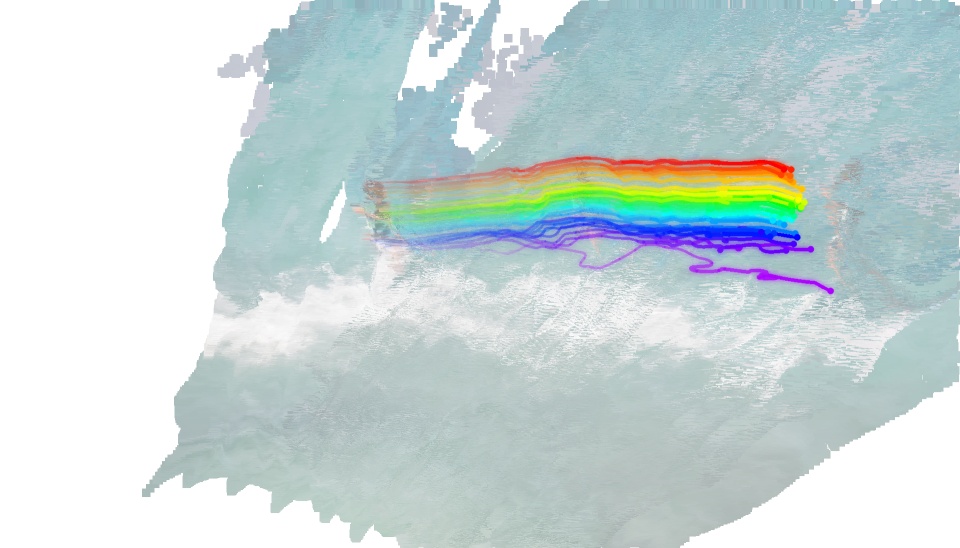} & \includegraphics[width=0.16\linewidth]{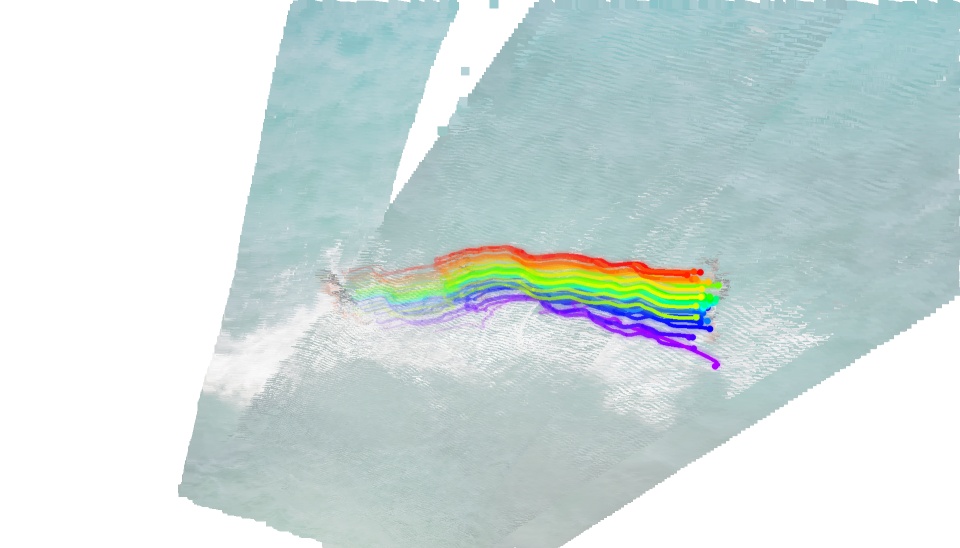} \end{tabular}}%
\raisebox{-\dp\qualbox}{\includegraphics[height=\dimexpr\ht\qualbox+\dp\qualbox\relax]{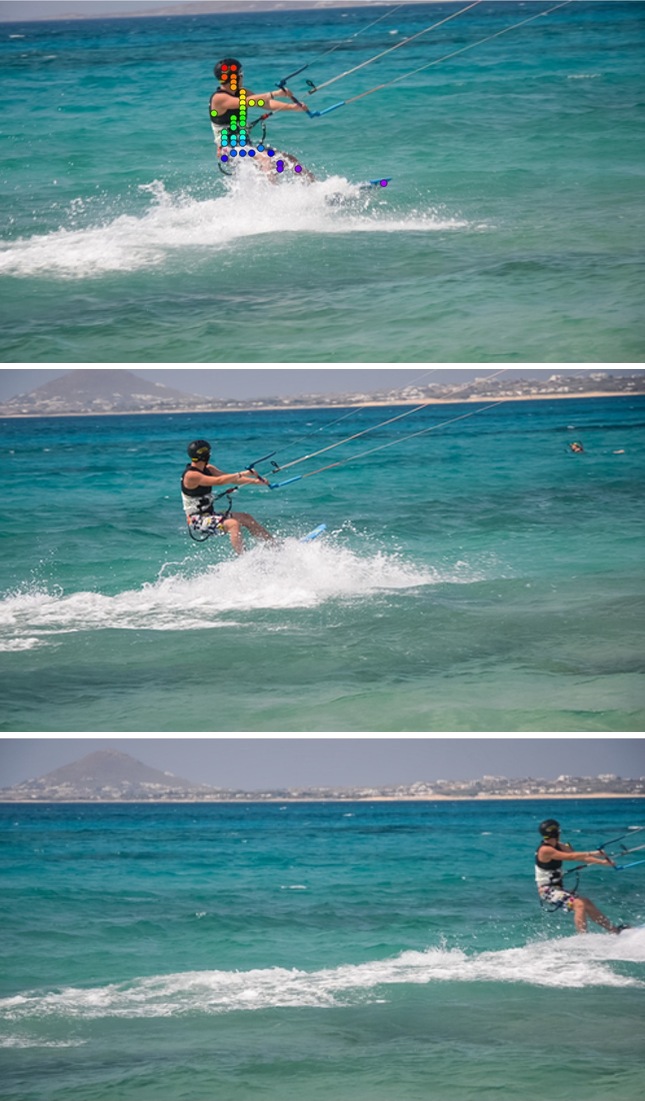}} & \usebox\qualbox \\[3pt]
\global\setbox\qualbox\hbox{\begin{tabular}[b]{@{}cccc@{}}\includegraphics[width=0.16\linewidth]{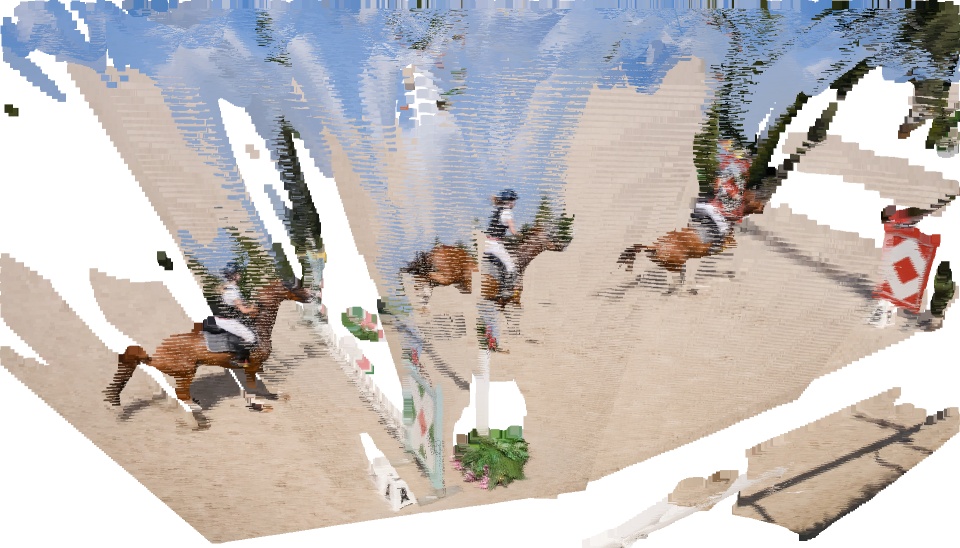} & \includegraphics[width=0.16\linewidth]{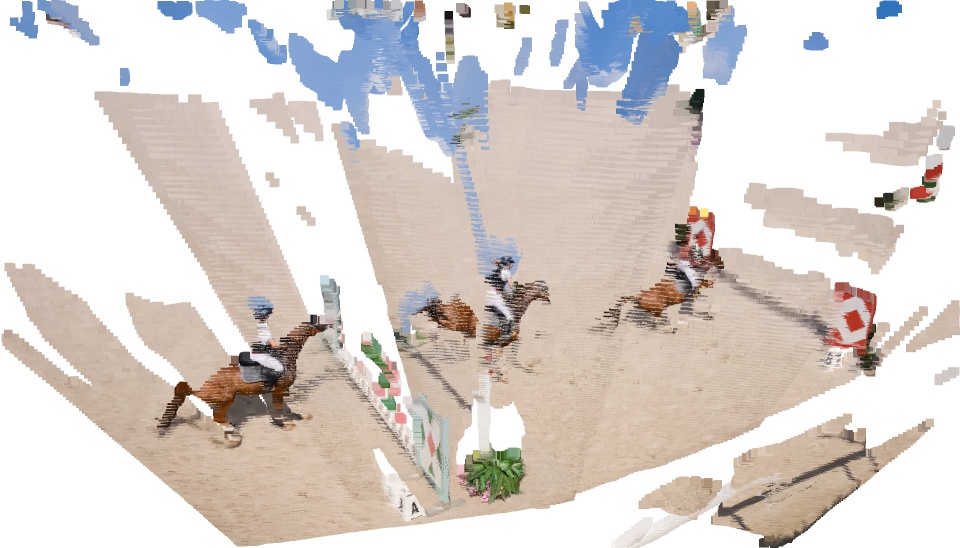} & \includegraphics[width=0.16\linewidth]{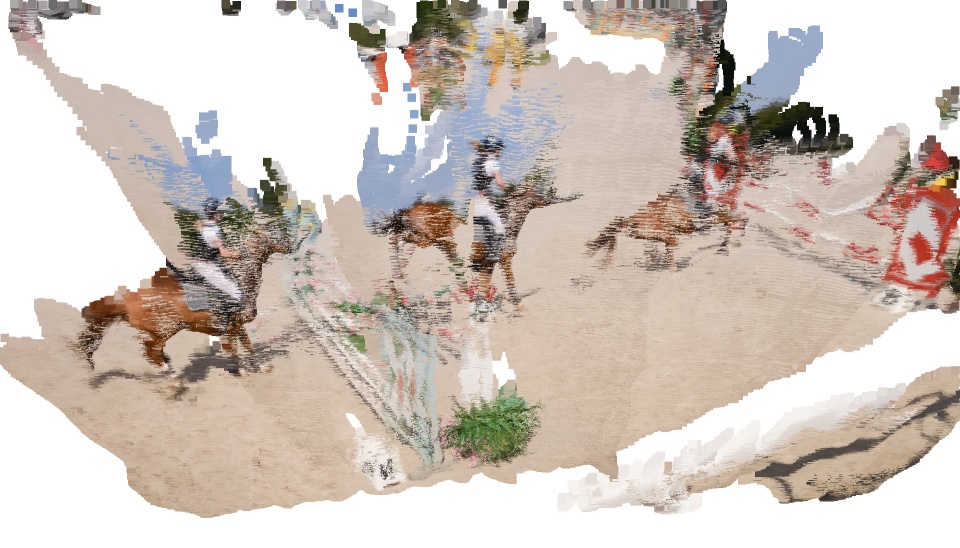} & \includegraphics[width=0.16\linewidth]{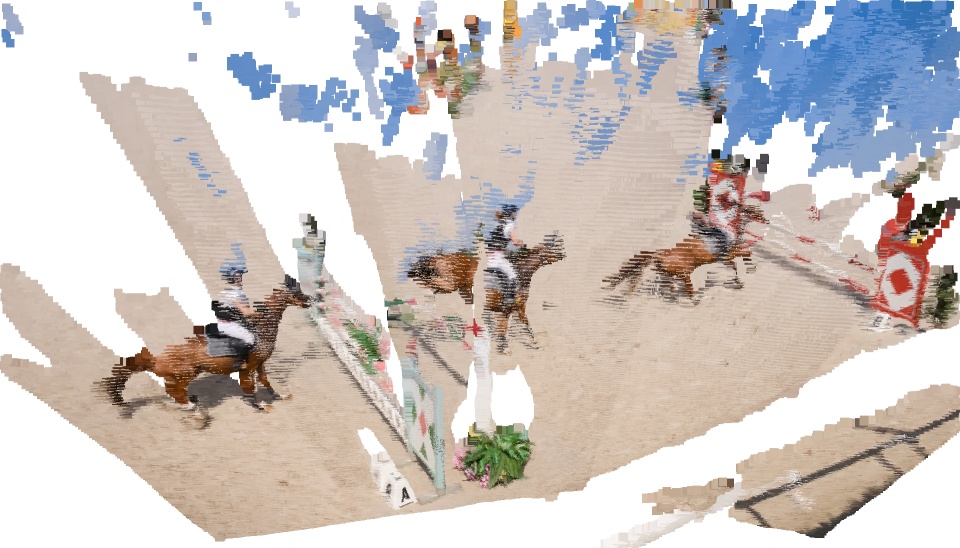} \\[-2pt] \includegraphics[width=0.16\linewidth]{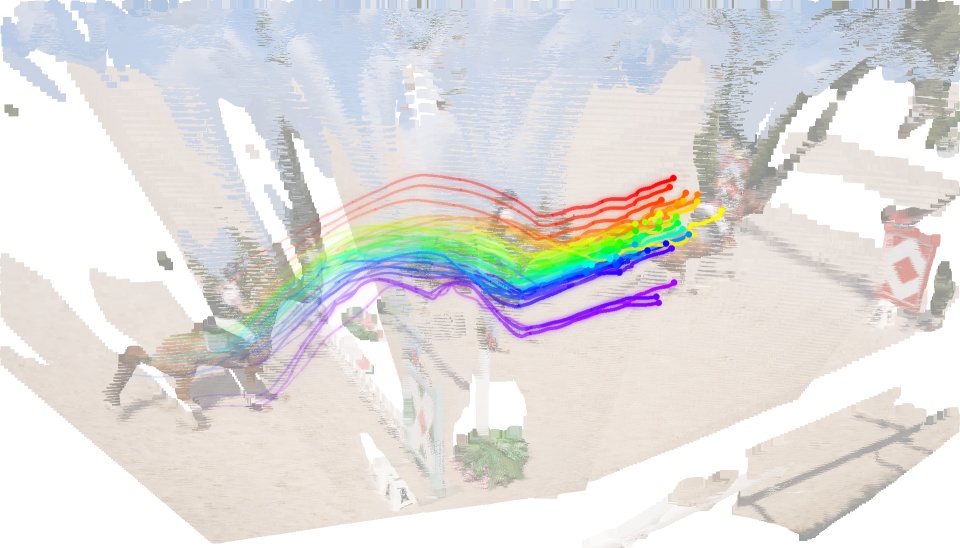} & \includegraphics[width=0.16\linewidth]{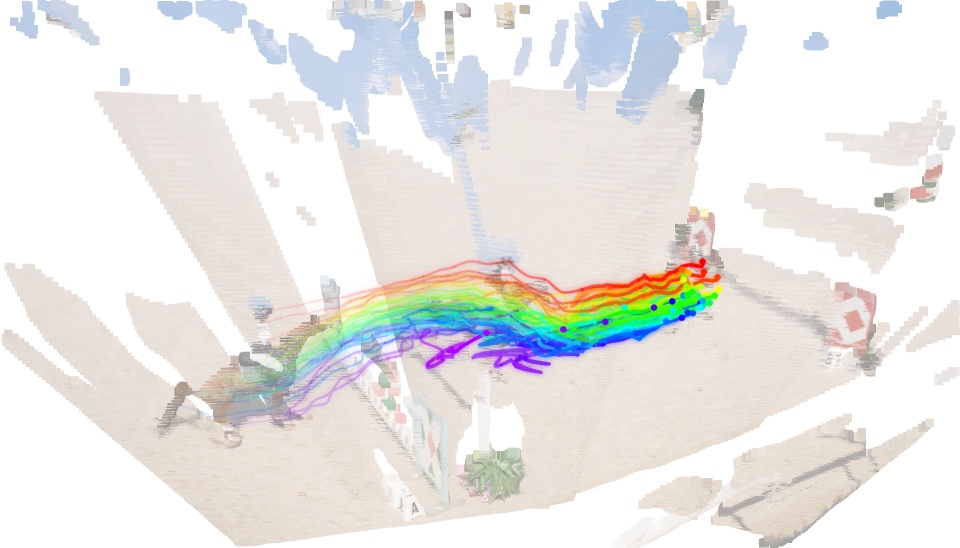} & \includegraphics[width=0.16\linewidth]{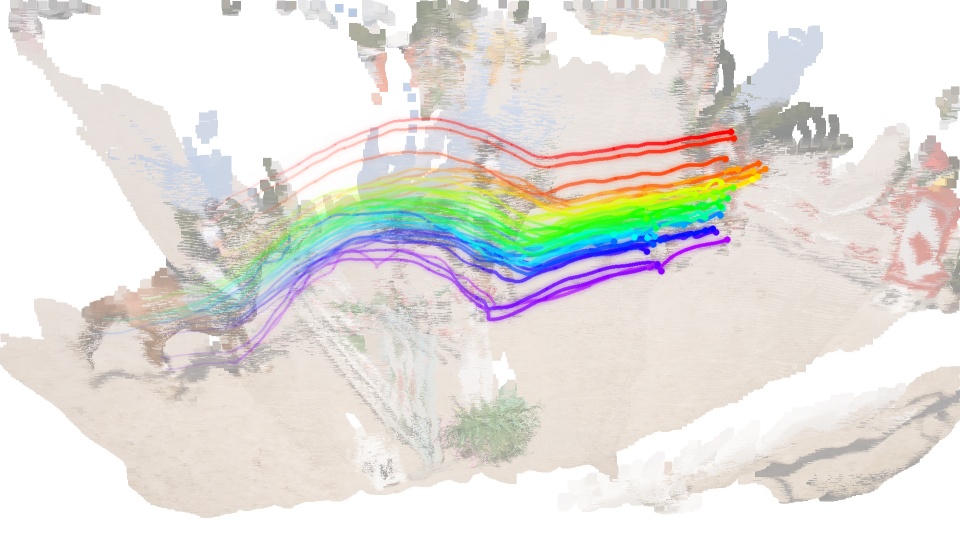} & \includegraphics[width=0.16\linewidth]{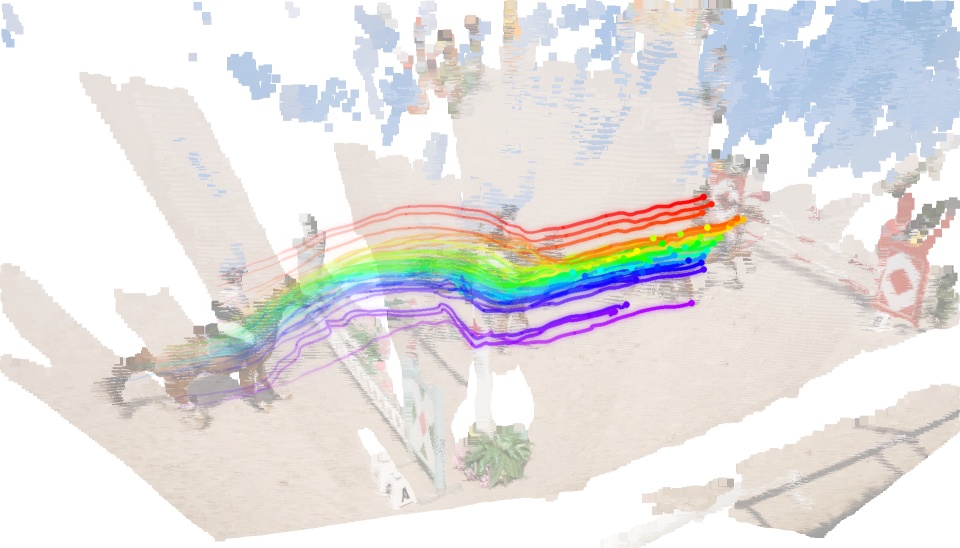} \end{tabular}}%
\raisebox{-\dp\qualbox}{\includegraphics[height=\dimexpr\ht\qualbox+\dp\qualbox\relax]{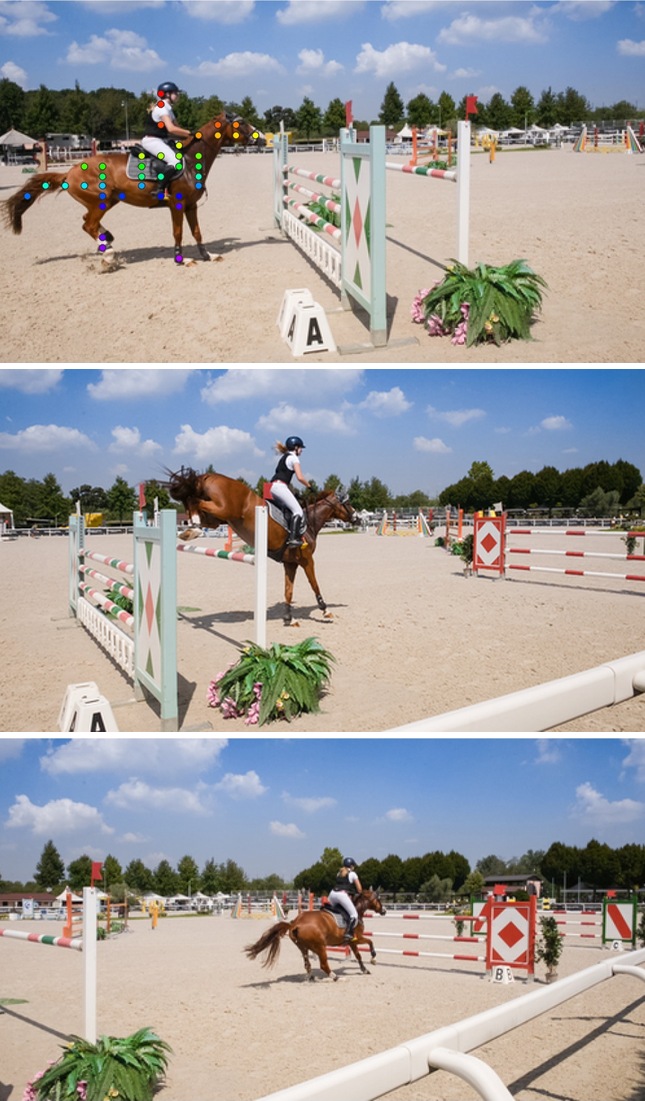}} & \usebox\qualbox \\[3pt]
\global\setbox\qualbox\hbox{\begin{tabular}[b]{@{}cccc@{}}\includegraphics[width=0.16\linewidth]{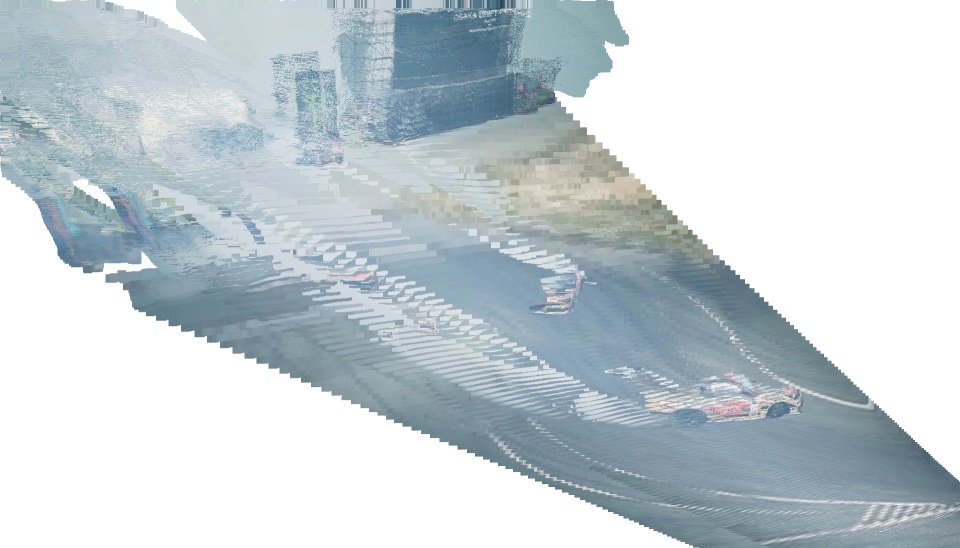} & \includegraphics[width=0.16\linewidth]{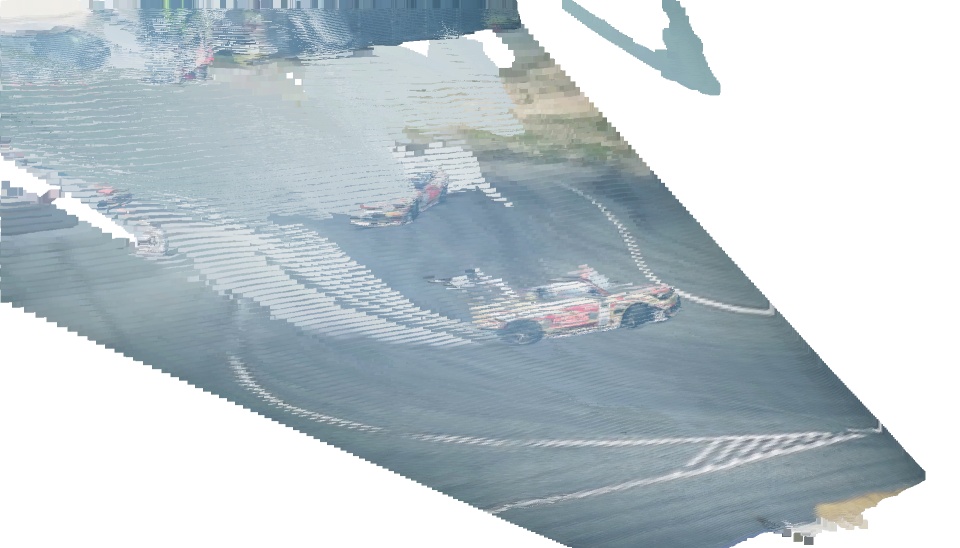} & \includegraphics[width=0.16\linewidth]{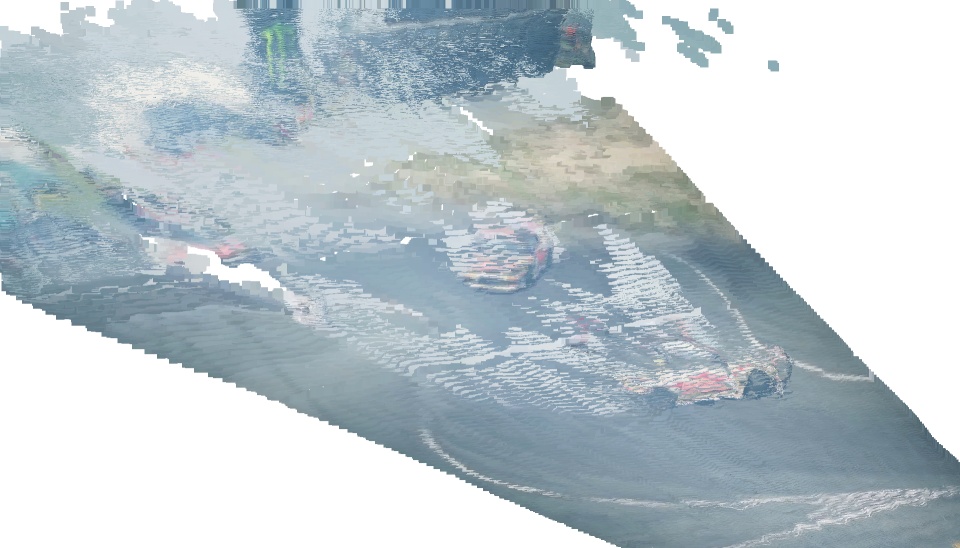} & \includegraphics[width=0.16\linewidth]{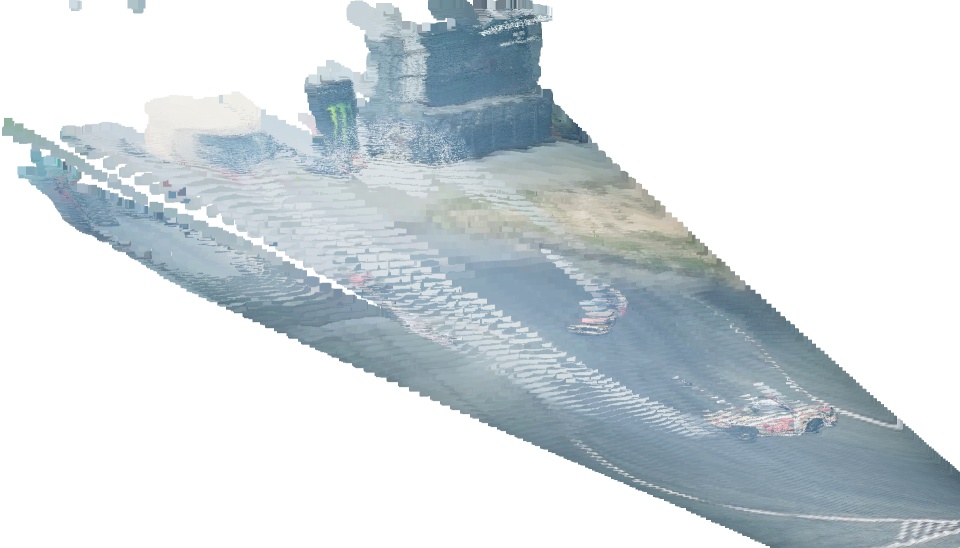} \\[-2pt] \includegraphics[width=0.16\linewidth]{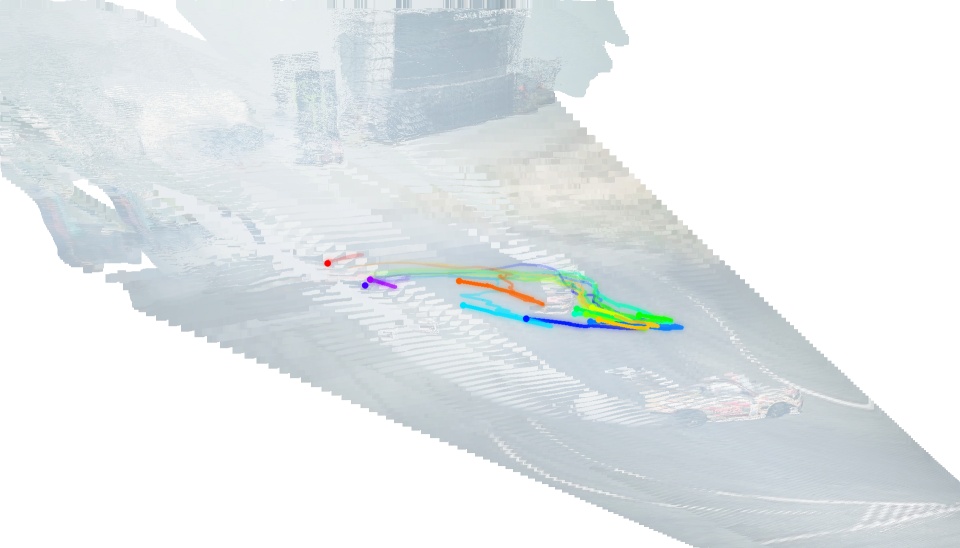} & \includegraphics[width=0.16\linewidth]{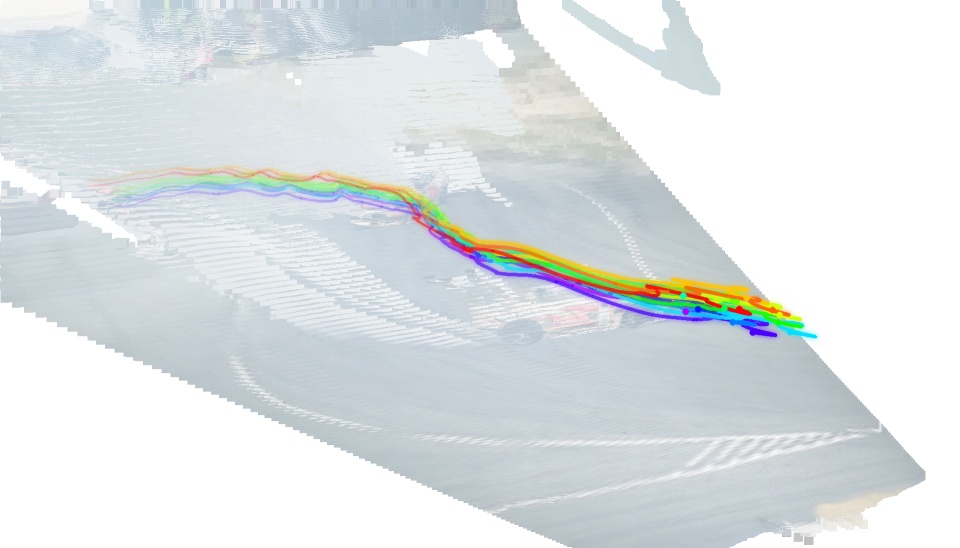} & \includegraphics[width=0.16\linewidth]{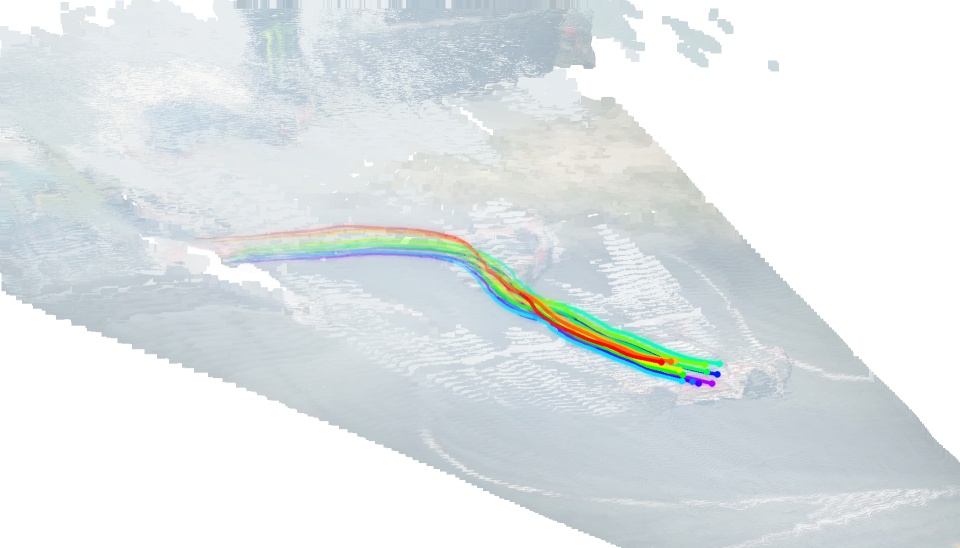} & \includegraphics[width=0.16\linewidth]{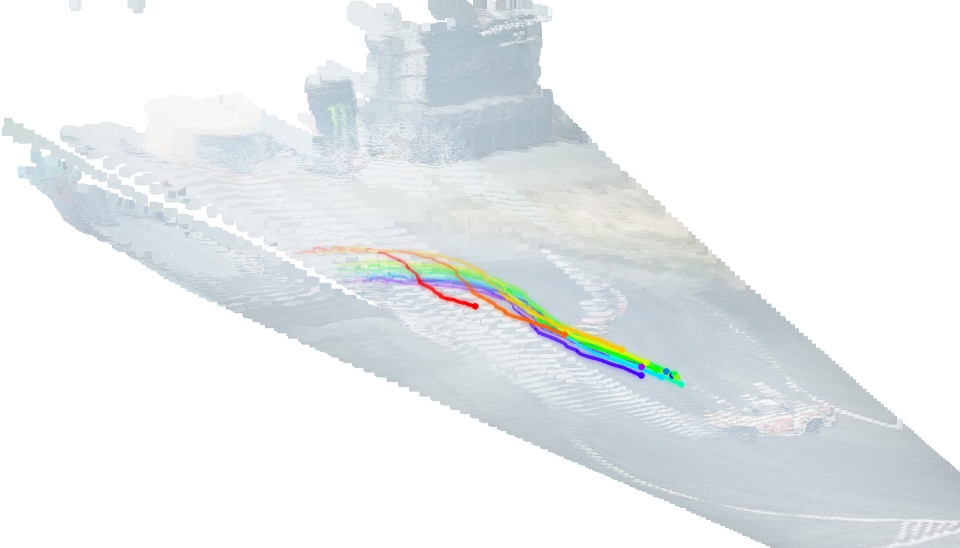} \end{tabular}}%
\raisebox{-\dp\qualbox}{\includegraphics[height=\dimexpr\ht\qualbox+\dp\qualbox\relax]{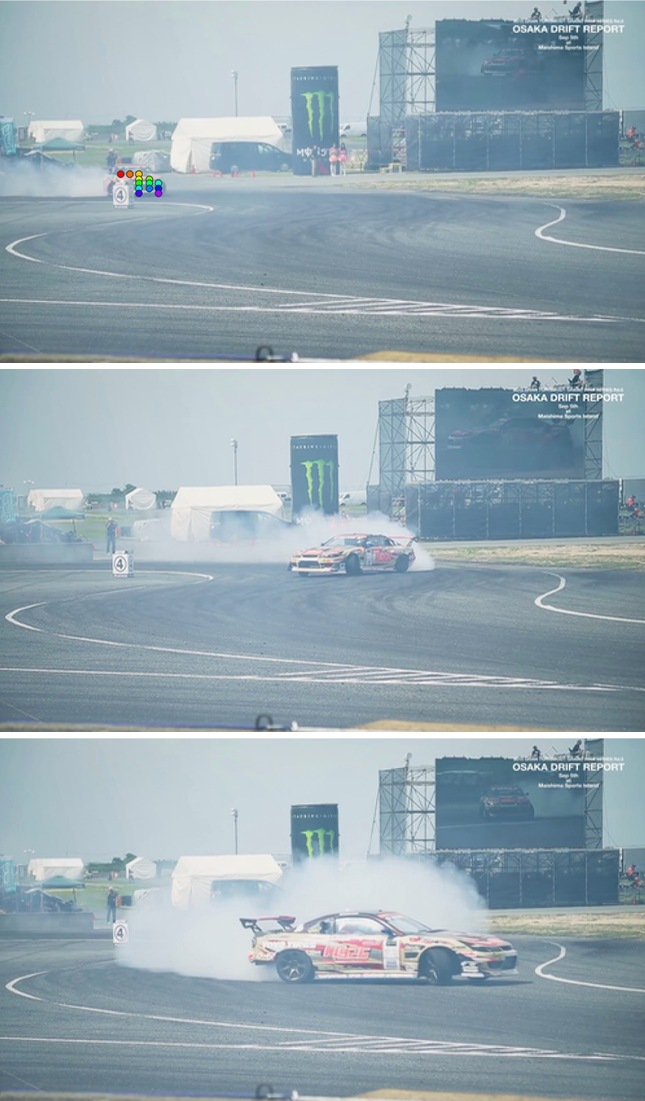}} & \usebox\qualbox \\[3pt]
\global\setbox\qualbox\hbox{\begin{tabular}[b]{@{}cccc@{}}\includegraphics[width=0.16\linewidth]{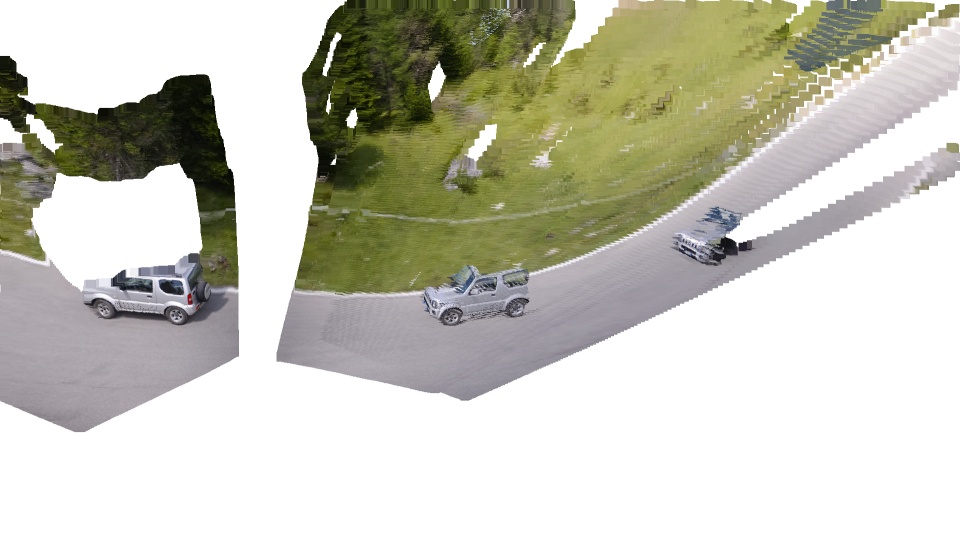} & \includegraphics[width=0.16\linewidth]{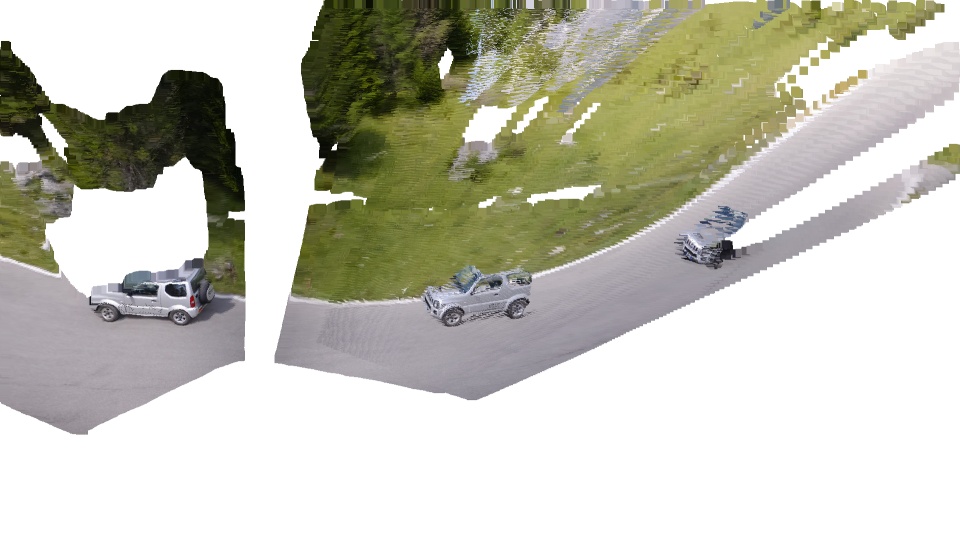} & \includegraphics[width=0.16\linewidth]{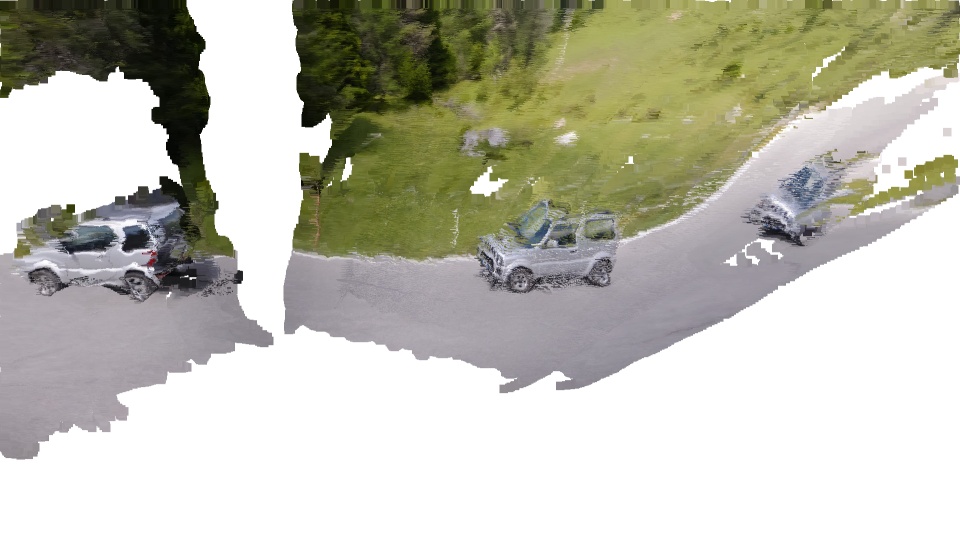} & \includegraphics[width=0.16\linewidth]{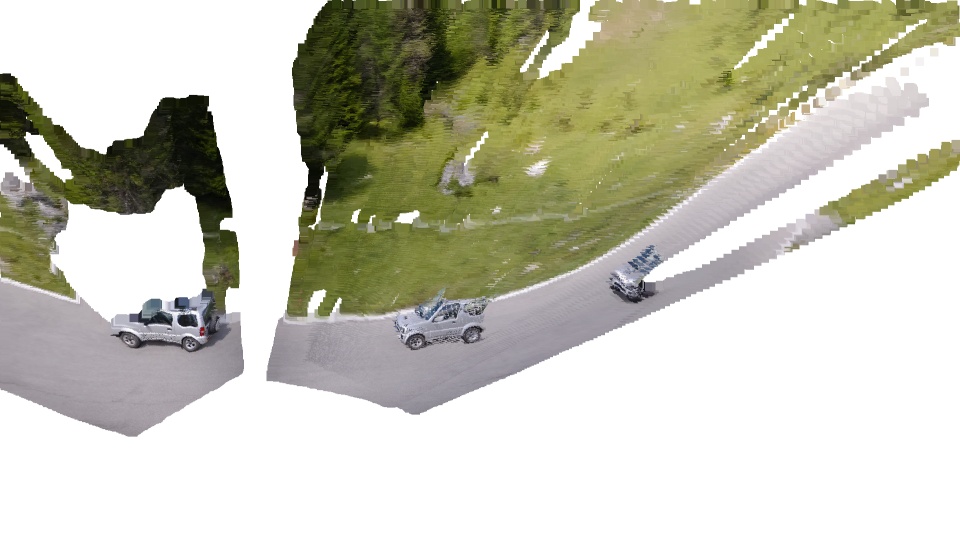} \\[-2pt] \includegraphics[width=0.16\linewidth]{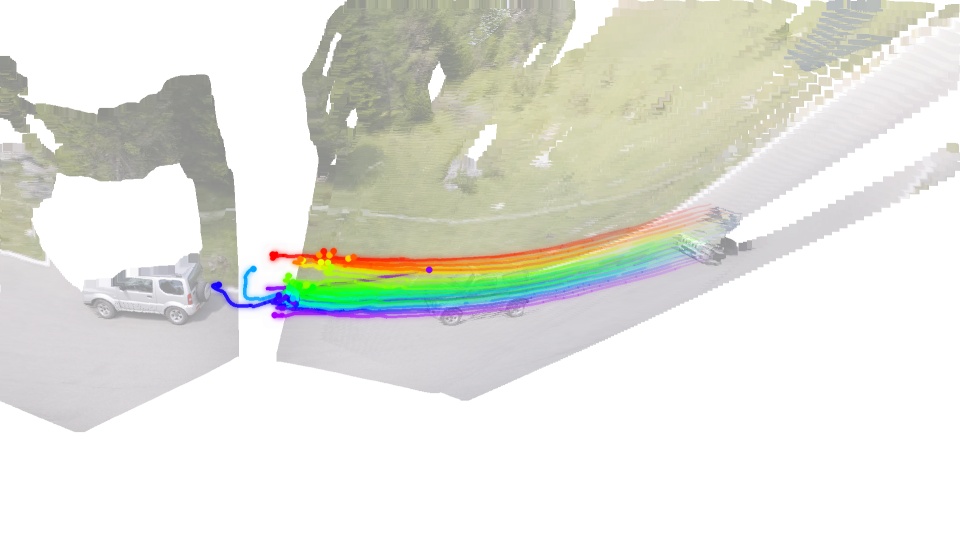} & \includegraphics[width=0.16\linewidth]{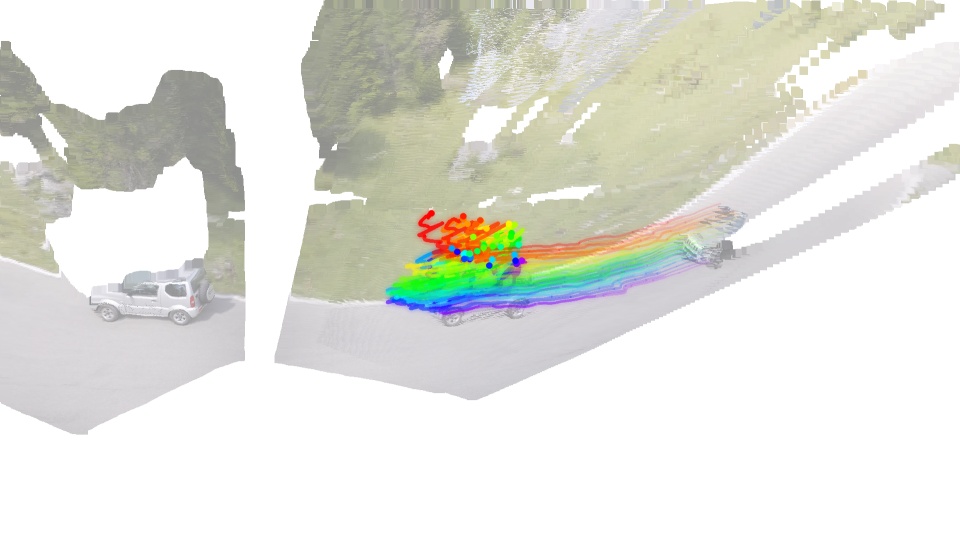} & \includegraphics[width=0.16\linewidth]{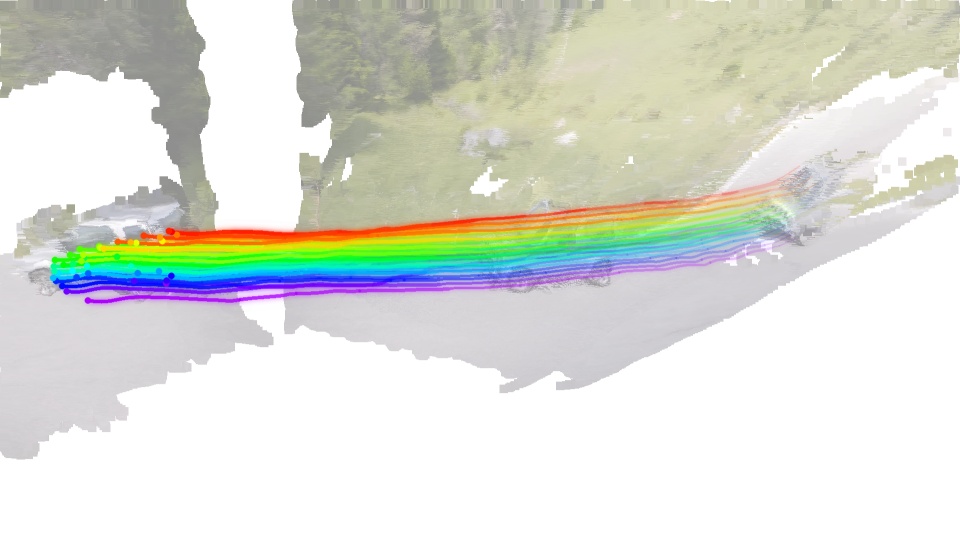} & \includegraphics[width=0.16\linewidth]{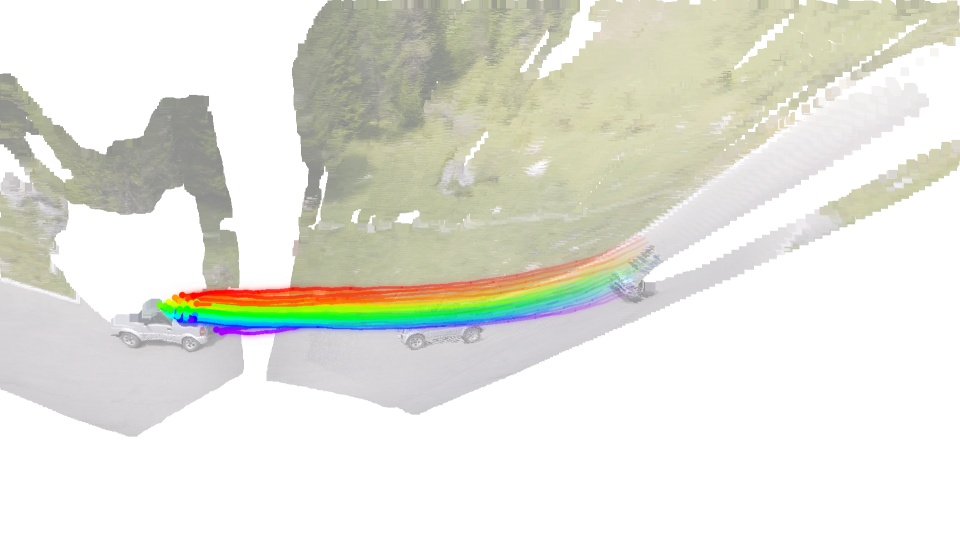} \end{tabular}}%
\raisebox{-\dp\qualbox}{\includegraphics[height=\dimexpr\ht\qualbox+\dp\qualbox\relax]{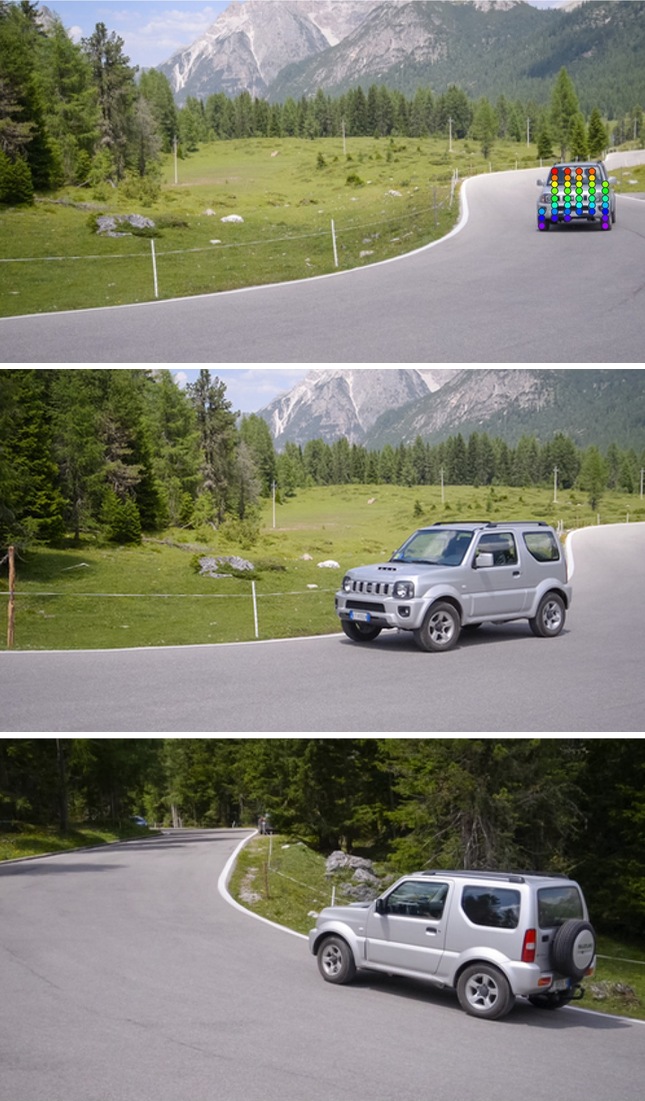}} & \usebox\qualbox
\end{tabular}
\vspace{-0.6em}
\caption{\textbf{Additional 4D reconstruction and world-space tracking results} on DAVIS~\cite{davis}: \emph{rollerblade}, \emph{swing}, \emph{kite-surf}, \emph{horsejump-high}, \emph{drift-chicane}, and \emph{car-turn}.
Layout and protocol follow \cref{fig:qual}.}
\label{fig:qualapp}
\end{figure*}

\cref{fig:qualapp} shows six further DAVIS clips.
In \emph{rollerblade}, \method keeps the skater's jump as a coherent bundle of arcs, while OmniX's upper trajectories drift.
In \emph{swing}, all methods recover the pendulum arcs, and OmniX flattens the lowest ones.
In \emph{kite-surf}, \method's trajectories reach the rider's position in the last frame, whereas 4RC's stop near the middle of the path and OmniX's bend downward.
In \emph{horsejump-high}, \method and 4RC follow the horse over the fence, while UniQuery4R's reconstruction blurs the horse and splits the static fence into offset copies.
In \emph{car-turn}, OmniX's trajectories tangle and UniQuery4R's overshoot past the car, whereas \method's end on the reconstructed car.
\emph{drift-chicane} is harder: the car is small and partly hidden by tire smoke, so all methods produce noisy trajectories, \method's shorter than UniQuery4R's.

\end{document}